%% file: main.tex
\documentclass[twoside,11pt]{article}

\usepackage{blindtext}

\usepackage[abbrvbib,preprint]{jmlr2e}

\usepackage{mathtools}
\usepackage{amsmath}
\usepackage{float}
\usepackage{placeins}

\usepackage{lastpage}
\jmlrheading{23}{2026}{1-\pageref{LastPage}}{5/28; Revised 5/22}{9/22}{21-0000}{Charles Bokor, Mark Cary,  Denise Morrey, Fabrizio Bonatesta}

\setcitestyle{numbers,square,comma,sort&compress}

\ShortHeadings{fSRD: Fuzzy Spectral Region Decomposition}{Bokor et al.}
\firstpageno{1}

\begin{document}

\title{fSRD: Fuzzy Spectral Region Decomposition -- Automated 
Multi Operator Koopman Representations via an Adaptive Spectral
Learning Architecture}

\author{\name Charles Bokor* \email charlesbokor@outlook.com \\
       \addr School of Engineering, Computing and Mathematics \\
       Oxford Brookes University \\
       Oxford OX33 1HX, UK
       \AND
       \name Mark Cary \email m.cary@lboro.ac.uk \\
       \addr Department of Aeronautical and Automotive Engineering \\
       Loughborough University \\
       Loughborough LE11 3TU, UK
       \AND
       \name Denise Morrey \email dmorrey@brookes.ac.uk \\
       \name Fabrizio Bonatesta \email fbonatesta@brookes.ac.uk \\
       \addr School of Engineering, Computing and Mathematics \\
       Oxford Brookes University \\
       Oxford OX33 1HX, UK}

\editor{???}

\maketitle

\begin{abstract}
Highly nonlinear chaotic dynamical systems remain difficult to model
due to fundamental trade-offs between complexity, expressivity, and data efficiency.
Modern machine learning methods achieve strong predictive performance
but often rely on a-priori system knowledge or curated data with limited 
interpretability. Koopman operator theory offers a promising direction 
via linear representation in an infinite-dimensional observable space. 
However, many data-driven Koopman methods seek globally valid operators 
for which useful finite-dimensional spectral embeddings remain difficult 
to identify under these constraints. To overcome associated limitations, 
we introduce Fuzzy Spectral Region Decomposition (fSRD), a fully automated 
learning framework for estimating finite Koopman representation via 
multiple operators. The proposed method realizes a data-adaptive 
framework for assembling locally invariant embeddings, termed Invariant 
Decomposition. fSRD achieves highly accurate linear reconstructions of nonlinear
systems while learning finite-dimensional representations of their induced evolution
operators, bridging interpretable operator-theoretic models with expressive data-driven
sequence learning. These embeddings are adaptively constructed via a global fuzzy tree model, 
drawing inspiration from fuzzy neural architectures to learn the induced dynamics while 
prioritizing parsimonious solutions. Empirical results across canonical chaotic systems 
(e.g., Lorenz and Duffing) and high-dimensional real-world data demonstrate strong predictive
accuracy, interpretability, and robust expressivity across data-rich and data-limited regimes, 
highlighting the method's generality.
\end{abstract}


\noindent
*Corresponding author.

\input{sections/introduction.tex}
\input{sections/methodology.tex}
\input{sections/results_duffing.tex}
\input{sections/results_lorenz.tex}
\input{sections/results_solar.tex}
\input{sections/descussion.tex}


\acks{The authors thank Hilal Sengul for valuable advice and
discussion regarding manuscript preparation, publication strategy, and
academic presentation of the work.}

\section*{Author Contributions}\label{sec:author-contributions}

Charles Bokor conceived the core research direction and developed the
primary machine learning architecture. Charles Bokor led the methodology
development, implementation, experimentation, validation, analysis, and
preparation of the manuscript throughout the project.

Mark Cary contributed to technical discussion,
statistical and machine learning methodology,
manuscript review, and made contributions to the regularization algorithm derived from
prior related work.

Denise Morrey and Fabrizio Bonatesta provided academic guidance and feedback during the early stages of the research. 
Denise Morrey additionally provided academic support and manuscript review in the 
final stages of the project.

\newpage
\vskip 0.2in
\bibliography{refs}

\end{document}

%% file: sections/introduction.tex
\section{Introduction}\label{sec:introduction}

Real-world systems frequently exhibit complex and evolving dynamics that
challenge conventional modelling approaches \cite{Guckenheimer1983}. Established fields
such as system identification and control theory have developed a wide
range of techniques for modelling such behaviour using combinations of
first-principles reasoning and statistical inference, typically tailored
to specific application goals \cite{Ljung1999,Eykhoff1971,Tangirala2015,Colonius2000}. As
system complexity increases, however, these approaches are increasingly
supplemented or replaced by data-driven machine learning (ML),
artificial intelligence (AI), and optimization techniques \cite{BruntonBook2019}.

Two persistent challenges motivate this shift. First, many real systems
exhibit nonlinear or chaotic behaviour such as strange attractors,
multi-stability, and non-periodic trajectories \cite{Gros2012,Valentin2014}. Modern
ML/AI methods can often capture such high-dimensional structure from
data, identifying patterns or representation beyond direct analytical
intuition \cite{BruntonBook2019,Champion2019,Brunton2021,Sarker2021,Korda2018}. However, these
tools often introduce new difficulties including limited
interpretability, increasing model complexity, non-parsimonious
representations, and reliance on substantial prior knowledge or curated
datasets \cite{Alzubaidi2021}. Consequently, there remains no widely adopted
mathematical framework that simultaneously combines expressive modelling
capability, interpretability, and systematic construction without
reliance on strong prior system knowledge.

Koopman operator theory provides one promising direction, offering a
principled route to universal linear representations of nonlinear
dynamical systems through spectral embeddings in function space
\cite{Koopman1931}. In practice, however, constructing useful Koopman embeddings
remains challenging beyond relatively simple settings. Data driven
Koopman estimates such as dynamic mode decomposition (DMD) \cite{SCHMID2010} have
achieved broad practical success, yet a framework that demonstrates
universal expressivity combined with robustness and automation has not
fully emerged.

Motivated by these gaps, this work introduces Fuzzy Spectral Region
Decomposition (fSRD), a systematic and fully automated ML architecture
for estimating a Koopman representation. The following subsections
outline only the theoretical foundations necessary to construct and
motivate this approach, before returning with the associated gaps and
resulting contribution. Further technical details are provided in
\cite{BokorThesisfSRD} and the references therein. Supplementary Material
for this manuscript is available as an ancillary PDF with the arXiv submission.

\subsection{Dynamical System}\label{subsec:dynamical-system}

To characterize system dynamics, one can seek a mapping
\(F:\mathbf{x}_{t} \rightarrow \mathbf{x}_{t + 1}\). Consider the
discrete formulation:

\begin{equation}
\mathbf{x}_{t + 1} = F(\mathbf{x}_{t})
\label{eq:dynamics}
\end{equation}

Where t is the evolution variable,
\(\mathbf{x}_{\mathbf{t}}\mathcal{\in M \subseteq}\mathbb{R}^{n}\) is a
vector snapshot of the system state at \(t\) (phase/state space,
vectorized video frame pixels, parameters, etc), and
\(F\mathcal{:M \rightarrow M}\). For an arbitrary step \(t\), the flow
map \(F^{\circ t}\) denoted via superscript \(t\) is a repeating
\(t\)-fold function composition \cite{Ghosh2022} s.t.:

\begin{equation}
\mathbf{x}(t) = F^{\circ t}(\mathbf{x}(0))
\label{eq:composition1}
\end{equation}

and:

\begin{equation}
F^{\circ t}( \mathbf{x} ) = F(F(F(\ldots(F(\mathbf{x})))))
\label{eq:composition2}
\end{equation}

These dynamics can be placed into a linear framework:

\begin{equation}
\mathbf{\ }\mathbf{x}_{t + 1}\mathbf{=}F( \mathbf{x}_{t} )\mathbf{= A}\mathbf{x}_{t}
\label{eq:linear_dynamics}
\end{equation}

where \(\mathbf{A}\) is a best fit linear matrix approximation for the
linear operator \(A(t)\), advancing snapshot measurements forward
\cite{Guckenheimer1983}. \(A(t)\) is often assumed to be normal and compact, acting on
an infinite dimensional complex Hilbert space \(\mathcal{H}\), s.t.
\(A(t)\mathcal{:H \rightarrow H}\) \cite{Hall2013,Hoffman1971,Lang1987}. Under
this assumption, a finite-dimensional approximation \(\mathbf{A}\) may
be obtained by restricting \(\mathcal{H}\) to a subspace \(\mathcal{V}\)
with dimensionality no greater than that of \(\mathcal{M}\) s.t.,
\(\mathcal{V \subseteq}\mathbb{C}^{n}\mathcal{\subset H}\), and
\(\mathbf{A}\mathcal{:V \rightarrow V\ }\)\cite{Hall2013}.

\subsection{Koopman Operator Theory}\label{subsec:koopman-operator-theory}

As defined so far, \(A(t)\) would struggle to capture
strongly nonlinear characteristics within \(F\) if applied to
\(\mathbf{x}_{t}\). To circumvent this, one can search for alternate
vectors of co-ordinates \(\mathbf{y}_{t}\), s.t. the dynamics
are ideally linearized but more realistically simplified
\cite{Colonius2000,Bedrossian1991,Meiss2007,Nelles2001,Farrell2006}.

Koopman Operator Theory is a popular form of this, evolving a family of
functions in a space of possible observables
\(\mathbf{y}_{t} = g(\mathbf{x}_{t})\) that ideally lift the dynamics
into a higher dimensional space s.t. the nonlinearities are `smoothed'
or `unfolded' \cite{Nandanoori2021}. This results in:

\begin{equation}
\mathcal{K}^{\circ t}g( \mathbf{x}_{0} ) = g( F^{\circ t}( \mathbf{x}_{0} ) ) = g( \mathbf{x}_{t} )
\label{eq:koopman}
\end{equation}

where \(G( \mathcal{M}) \coloneqq H\) is the linear vector
space containing the family of measurement functions or observables
\(g\mathcal{:M} \rightarrow \mathbb{C}\), and
\(\mathcal{K}^{\circ t}:G( \mathcal{M}) \rightarrow G\mathcal{(M)}\)
is the infinite-dimensional linear Koopman operator evolving the system
along the trajectory \(t \rightarrow g_{t}\) \cite{Koopman1931}. Broadly, any
nicely behaved function of the system states can be a measurement
function. If invertible and measure preserving, a \(\mathcal{K}^{\circ t}\)
defined on a Hilbert space is unitary \cite{Hall2013}, again being a repeating
t-fold composition that acts as a transfer operator for
\(F^{\circ t}\) \cite{Klus2015}.

\subsection{Invariance}\label{subsec:invariance}

A key characteristic for this work is that \(G\mathcal{(M)}\) is forward
invariant under the action of \(\mathcal{K}^{\circ t}\), meaning a finite
set of functions \(g_{j}\) in the space \(G\mathcal{(M)}\), where
\(j = 1 \rightarrow n\), remain in that space after \(\mathcal{K}\) is
applied \cite{Chai2015}:

\begin{equation}
g = a_{1}g_{1} + a_{2}g_{2} + \ldots + a_{n}g_{n},\ \ s.t.,\ \ \mathcal{K}g = b_{1}g_{1} + b_{2}g_{2} + \ldots + b_{n}g_{n}
\label{eq:invariance}
\end{equation}

where \(\mathcal{K}\) is the generator for the family of transforms
\(\mathcal{K}^{\circ t}\). One method of enforcing this is through the
application of spectral theory to the Koopman operator. Under suitable
assumptions, this allows the evolution of the vector-valued \(\boldsymbol{g}=[g_{1},g_{2},\dots,g_{n}]\) to be expressed
via a Koopman mode decomposition \cite{Brunton2021,Mezi2005}:

\begin{equation}
\boldsymbol{g}( \mathbf{x}_{t} )
=
\mathcal{K}^{\circ t}
\boldsymbol{g}( \mathbf{x}_{0} )
=
\mathcal{K}^{\circ t}
\sum_{j}^{\infty}
\boldsymbol{\varphi}_{j}(\mathbf{x})\mathbf{v}_{j}
=
\sum_{j}^{\infty}
\lambda_{j}^{t}
\boldsymbol{\varphi}_{j}(\mathbf{x})\mathbf{v}_{j}
\label{eq:koopman_decomp}
\end{equation}

where \(\mathbf{v}_{j}\) is the \(j^{th}\) Koopman mode with
its respective eigenvector \(\boldsymbol{\varphi}_{j}\) and eigenvalue
\(\lambda_{j}^{t}\). Formulating an observable as a linear combination
of eigenfunctions infers that the subspace of \(G\mathcal{(M)}\) which
the decomposition spans is invariant under the action of \(\mathcal{K}\)
\cite{SCHMID2010,Mezi2005}.

\subsection{Dynamic Mode Decomposition
(DMD)}\label{subsec:dynamic-mode-decomposition-dmd}

DMD is a popular non-parametric algorithm for the data-driven
characterization of a best fit linear operator that approximates the
Koopman operator \cite{SCHMID2010,Rowley2009}. Following Tu et al \cite{Tu2014}, DMD is
presented here in its generalisable formulation, referred to as the
`exact DMD'. Consider the collection of system snapshots
\(\mathbf{X}^{n \times m}=[ \mathbf{x}_{t},\mathbf{x}_{t + 1},\ldots,\mathbf{x}_{t = m} ]\).
This can be organised into pairs
\(\left\{ ( \mathbf{x}_{t_{k}},\mathbf{x}_{t_{k} + \mathrm{\Delta}t} ) \right\}_{k = 1}^{m}\)
to produce the matrices:

\begin{equation}
\mathbf{X}_{1} = \left\lbrack \begin{matrix}
| \\
\mathbf{x}_{t_{1}} \\
|
\end{matrix},\begin{matrix}
| \\
\mathbf{x}_{t_{2}} \\
|
\end{matrix},\ldots,\begin{matrix}
| \\
\mathbf{x}_{t_{m - 1}} \\
|
\end{matrix} \right\rbrack,\ \ \mathbf{X}_{2} = \left\lbrack \begin{matrix}
| \\
\mathbf{x}_{t_{1} + \mathrm{\Delta}t} \\
|
\end{matrix},\begin{matrix}
| \\
\mathbf{x}_{t_{2} + \mathrm{\Delta}t} \\
|
\end{matrix},\ldots,\begin{matrix}
| \\
\mathbf{x}_{t_{m - 1} + \mathrm{\Delta}t} \\
|
\end{matrix} \right\rbrack
\label{eq:shift_matrices}
\end{equation}

Notably the snapshot vectors \(\mathbf{x}_{t_{k}}\)need not be
sequential or evenly spaced. From \eqref{eq:shift_matrices} the dynamical system
\(\mathbf{x}_{t + 1}=F( \mathbf{x}_{t} )=\mathbf{A}\mathbf{x}_{t}\)
can be re-written as:

\begin{equation}
\mathbf{X}_{2} \approx \mathbf{A}\mathbf{X}_{1}
\label{eq:shift_matrices_eq} 
\end{equation}

which is a potentially over or under constrained problem where
\(\mathbf{A}\) acts as a best fit linear matrix that estimates
\(\mathcal{K}\) in the discrete form. It has been shown that \eqref{eq:shift_matrices_eq} can
be solved using a pseudo-inverse \(\mathbf{X}_{1}^{+}\) via
\(\mathbf{A} \approx \mathbf{X}_{2}\mathbf{X}_{1}^{+}\) within the least
squares optimization problem \cite{Tu2014}:

\begin{equation}
\underset{A}{argmin}\left\| \mathbf{X}_{2} - \mathbf{A}\mathbf{X}_{1} \right\|_{F}
\label{eq:DMD_cost} 
\end{equation}

\eqref{eq:DMD_cost} is often solved via a decomposition of \(\mathbf{X}_{1}^{+}\),
such as QR or SVD, to form a truncated projection basis of dimension
\(r\) \cite{Tu2014}. This allows the formulation of a DMD spectral expansion:

\begin{equation}
\boldsymbol{g}( \mathbf{x}_{t} ) 
=
\mathbf{x}_{t} 
=
\sum_{i = 1}^{r}{\phi_{i}\lambda_{i,DMD}^{t - 1}b_{i}} 
=
\mathbf{\Phi}\mathbf{\Lambda}^{t - 1}\mathbf{b}
\label{eq:DMD_decomp} 
\end{equation}

where \(\mathbf{\Lambda}\) is a diagonal matrix of the DMD eigenvalues
\(\mathbf{\lambda}_{DMD}^{t - 1}\), \(\mathbf{b}\) is the DMD mode
amplitudes vector, and \(\mathbf{\Phi}^{n\mathbf{\times r}}\) are DMD
modes. \eqref{eq:DMD_decomp} then allows an approximation of the full data matrix:

\begin{equation}
\mathbf{X} = \mathbf{\Phi}diag( \mathbf{b} )\mathbf{T}( \boldsymbol{\omega}_{DMD} )
\label{eq:DMD_decomp_matrix} 
\end{equation}

where \(\mathbf{T}( \boldsymbol{\omega}_{DMD} )\) is a
Vandermonde matrix raising the continuous DMD eigenvalue vector
\(\boldsymbol{\omega}_{DMD}\) to the powers \(0,...,t-1\). The full derviation for
\eqref{eq:DMD_decomp} and \eqref{eq:DMD_decomp_matrix} is provided in supplimentry material 1, i.e., SM-1.

The DMD spectral expansion, \eqref{eq:DMD_decomp}, provides a finite spectrum estimate
of the Koopman operator, confining \(\mathcal{K}^{\circ t}\) to an invariant
subspace with co-ordinates defined by a finite set of restricted
observables
\(\boldsymbol{g}( \mathbf{x}_{\mathbf{t}} ) = \mathbf{x}_{\mathbf{t}}\)
s.t., \(\mathcal{V}\mathbf{\subseteq}G( \mathcal{M} )\). Any
finite set of eigenfunctions for \(\mathcal{K}^{\circ t}\) will correspond
to an invariant subspace, or atleast where each observable is well
estimated an approximately invariant subspace \cite{Brunton2021}. Consequently,
identifying such a set is highly desirable, providing intrinsic
co-ordinates in which the dynamics behave linearly, while also allowing
system reconstruction via \eqref{eq:DMD_decomp_matrix}.

\subsection{Extended Dynamic Mode Decomposition
(EDMD)}\label{subsec:extended-dynamic-mode-decomposition-edmd}

While attractive, exact DMD presented has several pitfalls. As this is
not intended to be a comprehensive review of DMD history or development,
the interested reader is directed to \cite{BruntonBook2019,Brunton2021,Tu2014,Colbrook2023}. Here we highlight only those relevant to fSRD's
construction. The first concerns the use of
\(\boldsymbol{g}( \mathbf{x}_{t} ) = \mathbf{x}_{t}\),
which in isolation has been shown to be insufficiently rich to
universally characterize many nonlinear system dynamics \cite{Tu2014,Williams2015}.
Several methods thus attempt to determine or allocate
nonlinear observable embeddings that sufficiently enrich the model
within strongly nonlinear systems.

One of the most prominent examples of this is EDMD \cite{Williams2015}, which
follows the procedure of exact DMD but augments the initial data matrix
to contain nonlinear observables of the system. Consider an augmented
state snapshot \(\mathbf{y}_{t}\):

\begin{equation}
\mathbf{y}_{t}=\boldsymbol{g}( \mathbf{x}_{t} )=\begin{bmatrix}
g_{1}( \mathbf{x}_{t} ) \\
g_{2}( \mathbf{x}_{t} ) \\
\begin{matrix}
\mathbf{\vdots} \\
g_{q}( \mathbf{x}_{t} )
\end{matrix}
\end{bmatrix}
\label{eq:observables}
\end{equation}

Two data matrices are then formed:

\begin{equation}
\mathbf{Y}_{1} = \left\lbrack \begin{matrix}
| \\
\mathbf{y}_{t_{1}} \\
|
\end{matrix},\begin{matrix}
| \\
\mathbf{y}_{t_{2}} \\
|
\end{matrix},\ldots,\begin{matrix}
| \\
\mathbf{x}_{t_{m - 1}} \\
|
\end{matrix} \right\rbrack,\ \ \mathbf{Y}_{2} = \left\lbrack \begin{matrix}
| \\
\mathbf{y}_{t_{1} + \mathrm{\Delta}t} \\
|
\end{matrix},\begin{matrix}
| \\
\mathbf{y}_{t_{2} + \mathrm{\Delta}t} \\
|
\end{matrix},\ldots,\begin{matrix}
| \\
\mathbf{y}_{t_{m - 1} + \mathrm{\Delta}t} \\
|
\end{matrix} \right\rbrack
\label{eq:shift_matrices_Y}
\end{equation}

A best fit linear matrix \(\mathbf{A}_{\mathbf{Y}}\) for the operator
\(\mathbf{A}_{y}(t)\) is then estimated as in the DMD procedure but on
this new augmented state:

\begin{equation}
\mathbf{A}_{\mathbf{Y}} = \mathbf{Y}_{2}\mathbf{Y}_{1}^{+}
\label{eq:EDMD_eq}
\end{equation}

\begin{equation}
\text{s.t.,}\ \ \underset{A_{y}}{argmin}\left\| \mathbf{Y}_{2} - \mathbf{A}_{\mathbf{Y}}\mathbf{Y}_{1} \right\|_{F}
\label{eq:EDMD_cost}
\end{equation}

It has been demonstrated that taking the limit for infinite data, eDMD
converges to the Koopman operator in the subspace spanned by a `correct'
observable embedding \cite{KordaEDMD2018}. Without knowing the ideal embedding
\emph{a-priori} however, the assurance of a closed invariant subspace
with a candidate augmented measurement is forfeit \cite{Kamb2020}, generating
spurious eigenvalues that at best are non-generalisable \cite{Brunton2015}.

In some cases, application specific knowledge can be used to guide the
choice of observables, or to inform the assembly of a candidate library
of observables \cite{Lan2013,OWilliams2015,Mezi2013}. While effective, such
approaches partially undermine the data-driven objectives of applied
Koopman and EDMD methods, motivating substantial effort toward
mitigating reliance on system-specific priors. A range of
machine-learning-based approaches have sought to automate observable or
dictionary selection, either adaptively or directly from data through
optimization or learning frameworks such as neural embeddings or neural
operators \cite{Champion2019,Jin2024,Lusch2018,Folkestad2020}. Despite substantial
progress, such approaches continue to rely (explicitly or implicitly) on
assumptions about the structure of the underlying dynamics.

\subsection{\texorpdfstring{A Lack of Eigenfunctions
}{A Lack of Eigenfunctions }}\label{subsec:a-lack-of-eigenfunctions}

A structural concern for implementing any form of Koopman mode
decomposition is the possible absence of eigenfunctions. When an
operator is normal and compact, the spectral theorem yields a
decomposition in terms of eigenfunctions with an orthonormal basis
\cite{Hall2013,Searle2017}, though the associated eigenvalues need not be real
unless the operator is Hermitian. Nevertheless, even for seemingly
simple cases \cite{Mezic2017}, these assumptions are not satisfied, and no
Koopman eigenfunctions exist that may generate invariant transformations
of the global dynamics regardless of one's choice of observables.

Consider a generic operator \(\mathbf{\Gamma}:G \rightarrow G\)
acting upon some \(G \in \mathbb{R}^{n}\). One can formalise the above
issue utilizing the term spectrum as defined to operators on Banach
spaces \cite{Hall2013}. Application of the spectral theorem \cite{Hall2013} yields:

\begin{equation}
( \mathbf{\Gamma} - \lambda_{i}I )\upsilon_{i}= 0
\label{eq:lack_of_eigenfunctions}
\end{equation}

where \(\upsilon_{i}\) is an eigenvector of \(\mathbf{\Gamma}\) and
\(\lambda_{i}\) is its eigenvalue. \eqref{eq:lack_of_eigenfunctions} can be interpreted as a rule
to outline the scalars \(\lambda\) to which the operator
\((\mathbf{\Gamma} - \lambda_{i})\) does not have a bounded
inverse \cite{Brunton2021}. One can then decompose \(\sigma(\mathbf{\Gamma})\), 
where \(\sigma(\cdot)\) denotes the spectrum of a given matrix/operator, 
into two subsets based on the reason why this bounded inverse is
non-achievable \cite{Mezi2005}. Utilizing \eqref{eq:koopman_decomp} and generalising \eqref{eq:lack_of_eigenfunctions} to
the infinite dimensional \(\mathcal{K}^{\circ t}\boldsymbol{g}\) allows representation
via spectral subsets for some systems \cite{Mezi2005,Mezi2004}:

\begin{equation}
\mathcal{K}^{\circ t}\boldsymbol{g} = \int_{- \pi}^{\pi}{\exp(it\omega)d[ \mathbb{E}(\omega)\boldsymbol{g} ]} 
=
\sum_{k}^{}{\exp(it\omega_{k})\mathbb{P}_{k}(\omega)\boldsymbol{g}}
+ \int_{- \pi}^{\pi}{\exp(it\omega)d[ \mathbb{E}_{c}(\omega)\boldsymbol{g} ]}
\label{eq:projection_value_measure}
\end{equation}

where \(k\) denotes the counter for a finite set of eigenvalues \(\omega_{k}\) ,
\(\omega\) are continuous Koopman eigenvalues, and
\(\mathbb{E}( \cdot )\) is a projection-valued measure that is
partitioned into `atomic' projections \(\mathbb{P}_{k}\) and
`continuous' projections, \(\mathbb{E}_{c}\) \cite{Mezi2005,Mezi2004}. The
atomic component aligns with a finite spectrum termed point spectra.
These can be expressed in terms of \eqref{eq:koopman_decomp} eigenfunction based
orthonormal projections:

\begin{equation}
\mathbb{P}_{k}(\omega)\boldsymbol{g} = \varphi_{k}\left\langle \varphi_{k},\boldsymbol{g} \right\rangle
\label{eq:orthonormal_proj}
\end{equation}

Eigenvalues of \(\mathcal{K}^{\circ t}\) generated from \(\mathbb{P}_{k}\)
are often associated with `regular' or non-chaotic dynamics. The
residual \(\mathbb{E}_{c}\), however, coincides to the continuous
spectrum, which cannot be fully represented as a finite set of
eigenvalues \cite{Mezi2005}. Cases where the Koopman operator, on the chosen
observable space, has an empty point spectrum and admits only continuous
spectrum, such that for \(\lambda \in \sigma(\mathcal{K)}\) the operator
\(( \mathcal{K}-\lambda I )\) does not possess a bounded
inverse, corresponds to a lack of globally defined Koopman
eigenfunctions that admits a global invariant set.

\subsection{\texorpdfstring{Limits of Linearization
}{Limits of Linearization }}\label{subsec:limits-of-linearization}

Capuring \(\mathbb{E}_{c}\) via finite representations remains an area
of active research. Whenever \(\mathcal{K}^{\circ t}\) exhibits in some part
a continues spectra, no finite, globally defined set of eigenfunctions
can capture the dynamics \cite{Mezi2005,Mezi2013}. Beyond just a lack of
eigenfunctions, this can occur due to several other structural features
of the system, such as multiple orbits or chaotic dynamics. Consider for
example the former via Hartman-Grobman Theory, stating that a nonlinear
system can only be linearized within the neighbourhood of a hyperbolic
fixed point, i.e., the nonlinear system is locally topologically
conjugate to its linearization about said fixed point \cite{Hartman2002,Hartman1960,Grobman1959}.
This implies the existence of a local homeomorphism
\(h_{l}\mathcal{:M}\rightarrow\mathcal{M}_{y}\) i.e., a topological
isomorphism, between the two mappings \(F\) and \(A_{y}(t)\) s.t.:

\begin{equation}
h_{l} \circ F = A_{y}(t) \circ h_{l}
\label{eq:homeomorphism}
\end{equation}

Ideally, such mappings are diffeomorphic, allowing transition between
co-ordinates without complications. In practise, this stronger condition
is not always achievable \cite{Farrell2006}. While Koopman theory can extend
Hartman--Grobman linearization along trajectories or within attraction
basins \cite{Lan2013,Mezi2013}, systems with multiple attractors admit no
global conjugacy across invariant basins, often manifesting as
continuous spectra within \eqref{eq:projection_value_measure}. As a result, finite-dimensional
Koopman approximations based on a single global observable space (i.e.,
DMD, EDMD or learned neural embeddings) can only recover local or
partial linearizations, rather than a globally invariant representation.

\subsection{Ergodic Partitioning}\label{subsec:ergodic-partitioning}

A phase portrait perspective of the dynamics provides an alternative
approach for obtaining global Koopman representation. The ergodic
average of any \(g(x) \in G\mathcal{(M)}\) projects onto invariant
eigenspaces of the Koopman operator \cite{Brunton2021}. Ergodicity implies the
average orbital behaviour can be characterized by trajectories of single
points that encompass the average statistical properties of all points
within that orbit \cite{Mezi2004}. While useful to identify orbits of origin
for individual trajectories \cite{Budii2012}, this was developed into the
concept of ergodic partitioning \cite{Mezic1999}. Consider the complete dynamics
\(x_{t} = \mathbf{F}^{\circ t}(x_{0})\) where \(x\mathcal{\in M}\)
and \(\mathbf{F}\mathcal{:M \rightarrow M}\). As described in \cite{Nandanoori2021,Nandanoori2019},
if \(\mathbf{F}\) is not itself ergodic, then inherently
\(\mathcal{M}\) must be partitionable into multiple sets
\(\mathcal{M}_{i \rightarrow p}\mathcal{\in M}\) s.t. each respective
\(\mathbf{F}_{i}\) is ergodic, i.e., separable into ergodic partitions.

Assuming the ability to represent a given system via a global Koopman
operator \(\mathbf{K}_{s}\), Nandanoori et al \cite{Nandanoori2019} outline three
means of implementing ergodic partitioning:

\begin{enumerate}
\def\labelenumi{\arabic{enumi}.}
\item
  A top-down global phase space partitioning/classification of a given
  \(\mathcal{M}\) into valid Koopman Invariant subspaces
  \(\mathcal{M}_{i}\) with resultant operator
  \(\mathcal{K}_{i}^{\circ t}\)assuming access to sufficient trajectories.
\item
  Convergence to \(\mathbf{K}_{s}\) by expanding a local subset
  \(\mathcal{K}_{i}^{\circ t}\) spectra with iterative training on new
  trajectories.
\item
  Coupling several local \(\mathcal{K}_{i}^{\circ t}\) to form a candidate
  \(\mathbf{K}_{s}\) for covered trajectories, assuming only local
  information is available.
\end{enumerate}

Perspectives 1 and 3 both end with some form of disjointed sets being
combined to form a global \(\mathbf{K}_{s}\), which can be expressed as:

\begin{equation}
\begin{split}
\boldsymbol{g}_{s}( x_{t} ) = \mathbf{K}_{s}\boldsymbol{g}_{s}( x_{0} ) = \begin{bmatrix}
\mathcal{K}_{1}\mathcal{X}_{1}\boldsymbol{g}_{1}( x_{0} ) \\
 \vdots \\
\mathcal{K}_{p}\mathcal{X}_{p}\boldsymbol{g}_{p}( x_{0} )
\end{bmatrix} 
\\ \propto \bigsqcup_{}^{}{\mathcal{K}_{i}\mathcal{X}_{i}\boldsymbol{g}_{i}( x_{0} )} 
=
\mathcal{K}_{1}\mathcal{X}_{1}\boldsymbol{g}_{1}( x_{0} ) \sqcup \ldots \sqcup \mathcal{K}_{p}\mathcal{X}_{p\ }\boldsymbol{g}_{p}(x_{0})
\label{eq:global_K}
\end{split}
\end{equation}

where the corresponding global observable vector \(\boldsymbol{g}_{s}\)
contains the individual observable vectors
\(\boldsymbol{g}_{i} = \lbrack g_{1}^{i},g_{2}^{i},\ldots,g_{n}^{i}\rbrack\)
for each sub-region \(\mathcal{M}_{i}\mathcal{\in M}\) for
\(i = 1,2,\ldots,p\) s.t.:

\begin{equation}
\boldsymbol{g}_{s}(x) = \begin{bmatrix}
\begin{matrix}
\mathcal{X}_{1}(x)\boldsymbol{g}_{1}(x) \\
{\mathcal{X}_{2}(x)\boldsymbol{g}}_{2}(x)
\end{matrix} \\
 \vdots \\
\mathcal{X}_{p}(x)\boldsymbol{g}_{p}(x)
\end{bmatrix}
\label{eq:global_g}
\end{equation}

and \(\mathcal{X}_{i}\) are characteristic functions that act as
indicator functions s.t.:

\begin{equation}
\mathcal{X}_{i}(x) = \mathbf{1}_{\mathcal{M}_{i}}(x) = \left\{ \begin{matrix}
1,\ \ \text{if}\ x \in \mathcal{M}_{i} \\
0,\ \ \text{if}\ x \notin \mathcal{M}_{i}
\end{matrix} \right.\
\label{eg:indicator_f}
\end{equation}

It is not necessary in application to demonstrate that the disjointed
subspaces are truly dynamically separate to be combined. Instead, they
require only the ability to form estimations of the associated operators
s.t. they can be distinguished from each other \cite{Sinha2020}. Consequently,
each characteristic function \(\mathcal{X}_{i}\) can be viewed as an
activation function that parametrizes the geometric location of the
respective subspace, distinguishing each relative to all other. In
\cite{Nandanoori2021,Nandanoori2019,Sinha2020} however this concept is not automated,
defining \emph{a-priori} via system knowledge each \(\mathcal{X}_{i}\)
and the observables required to approximate local finite ergodic
partitions via EDMD.

\subsection{Problem Statement}\label{subsec:problem-statement}

As outlined initially, existing approaches to nonlinear dynamics
struggle to balance expressive modelling capability, interpretability,
and systematic construction without reliance on strong prior system
knowledge. In principle, a valid global Koopman operator would provide a
universal linear representation of sequentially dependent processes,
though this generally requires an infinite-dimensional space of
observables \cite{Koopman1931,Mezi2005,Rowley2009,BudiiKoopman2012}. While recent
learning-based Koopman methods reduce reliance on hand-crafted
observables, the assumption of a single, globally defined observable
space, together with inherent structural limits of the system, continues
to limit the scope of Koopman-based approaches as universal linear
frameworks. These hard structural limits included:

\begin{itemize}
\item
  For many nonlinear systems, no finite-dimensional observable space
  admits nontrivial globally valid eigenfunctions (e.g., the Koopman
  operator exists but may possess predominantly continuous spectrum,
  which is poorly approximated by finite expansions).
\item
  A lack of global topological conjugacy across invariant sets
  preventing a globally valid Koopman-based eigenspectral representation
  (e.g., multi operator system).
\end{itemize}

While many such methods attempt to approximate or regularize continuous
spectral components, a systematic approach that addresses these
challenges while robustly handling continues spectra without the use of
a-priori system knowledge (e.g., avoiding spurious eigenvalues) remains
an open challenge.

While the assumption of globally valid observables has currently not
produced a universally accepted solution, a multi-operator approach has
potential to circumvent many of these challenges if the highlighted
region placement and boundary integration was resolved (i.e., the
automatic search, allocation, and union of locally valid Koopman
invariant partitions). A bespoke automated ML architecture could be
developed to this end.

\subsection{Summary of Contribution}\label{subsec:summary-of-contribution}

The principal contribution of this work is the ML algorithm fSRD,
realizing a novel framework for assembling invariant restricted spectral
integrals called invariant decomposition. This systematically automates
the generation of a multi-operator finite Koopman representation. In the
context of Koopman spectral embeddings, to the authors knowledge no
literature for such an automated methodology currently exists at time of
writing. fSRD removes Koopmans historic dependence on restrictive system
priors by accepting and building upon the structural limitations in
given data (noise, incomplete orbits, continues spectra, limiting
co-ordinates, lack of global eigenvalues, etc), providing a robust
methodology.

This Koopman inspired architecture suggests universal approximator
capabilities for sequentially dependant processes, serving as a general
alternative to a broad class of non-parametric estimators. Unlike other
methods however, the use of local Koopman embeddings inherently
preserves structure, with the provided architecture assembly
prioritizing a parsimonious set of locally invariant regions each with
interpretable orthogonal projections. This collection of linear
operators does not require the underlying system to admit a single
global linear operator, and so can provide global representation to high
dimensional, nonlinear, nonstationary, or chaotic dynamics when
structure exists locally.

Several novel solutions were required to realise this scheme, including:
a framework for assembling locally invariant Koopman regions (geometric
perspective of representation space and invariant decomposition); a core
ML/AI architecture to unite said disjoint regions (non-parametric fuzzy
hierarchal tree network); a methodology for selecting and optimizing
candidate region ensembles (forward pass and bespoke invariance
metrics); a local region definition in light of architectural
limitations (local framework and topological transform); and methods for
enforcing local and global parsimony (automated SVD regularization and
restricted integral pruning respectively). What follows is a methodical
breakdown of these elements throughout Section~\ref{sec:methodology}.
Due to the scope of the work, much of the individual
algorithms and proofs will be relegated to supporting material and
\cite{BokorThesisfSRD}, with the overarching operation summarised at the sections
end. Several case studies to demonstrate fSRD's performance in
nonlinear, chaotic and high dimensional problems will then be presented
in Section~\ref{sec:results}, including the Duffing oscillator, Lorenz system, and high
dimensional spatio-temporal solar data. The paper concludes with a
discussion and final remarks in Section~\ref{discussion-and-conclusions}.

%% file: sections/methodology.tex
\section{Methodology}\label{sec:methodology}

To develop an architecture for the purpose of constructing
multi-operator Koopman representation, much inspiration can be taken
from a subset of neuro-fuzzy networks called Local Model Networks
(LMN's) \cite{Babuka2003}. These are fuzzy ML architectures which smoothly
interpolate among systematically generated disjoint local models. The
local models are usually linear regressions, aiding interpretability,
but lacking descriptive power \cite{Nelles2001}. Consequently, as system
complexity increases, capturing strongly nonlinear behaviour typically
necessitates finer asymptotic discretization, which can reduce model
interpretability and increase structural complexity \cite{Nelles2001}, or a
selection of nonlinear basis functions based on a-priori system
knowledge \cite{Nelles2001}. What follows is a novel approach to assembling
invariant spectral embeddings, expanding upon current Koopman and LMN
theory.

\subsection{Assembling Spectral
Regions}\label{subsec:assembling-spectral-regions}

In exact DMD, \(\mathbf{X}\in\mathbb{R}^{m \times t}\)
s.t. \(m\) rows are state-space parameters assembled as system snapshot
vectors \(\mathbf{x}_{t_{k}}\), and \(t\ \)columns their evolution with
a given variable from \(k = 1,\ldots,t\). This notation will be used
throughout the rest of this work s.t., \(\mathbf{X} = \begin{bmatrix}
\mathbf{x}_{t_{1}} & \mathbf{x}_{t_{2}} & \begin{matrix}
\cdots & \mathbf{x}_{t_{t}}
\end{matrix}
\end{bmatrix}\). Although historically popularized in fluid mechanics,
DMD is not restricted to spatially discretized physical fields. In
principle, it applies to any sequentially observed system whose
measurements can be embedded in a common finite-dimensional vector
space. \(\mathbf{X}\) thus encodes the full trajectory of the system in
a finite-dimensional state or embedded representation space, capturing
both instantaneous states (columns) and their evolution along a given
variable (rows). Consider a temporal system
\(\mathbf{X}\mathcal{\in M}\) in which two invariant dynamic operators,
\(\mathcal{K}_{1}:G( \mathcal{M}_{1} ) \rightarrow G( \mathcal{M}_{1} )\)
and
\(\mathcal{K}_{2}:G( \mathcal{M}_{2} ) \rightarrow G( \mathcal{M}_{2} )\),
dictate system evolution, i.e., \(p = 2\). Via \eqref{eq:koopman_decomp} and \eqref{eq:global_K}, this
can be represented via disjointed union:

\begin{equation}
\boldsymbol{g}_{s}( x_{t} ) 
=
\mathbf{K}_{s}\boldsymbol{g}_{s}( x_{0} ) 
\propto
\bigsqcup_{}^{}{\mathcal{K}_{i}\mathcal{X}_{i}\boldsymbol{g}_{i}( x_{0} )} 
\approx 
\bigsqcup_{}^{}{\mathcal{X}_{i}\mathbf{K}_{i}\boldsymbol{g}_{i}( x_{0} )}
=
\mathcal{X}_{1}\mathbf{\Phi}_{1}\mathbf{\Lambda}_{1}^{t-1}\mathbf{b}_{1}
\sqcup
\mathcal{X}_{2}\mathbf{\Phi}_{2}\mathbf{\Lambda}_{2}^{t-1}\mathbf{b}_{2}
\label{eq:assemble_regions}
\end{equation}

with matrix form:

\begin{equation}
g_{s}(X) 
=
( \mathcal{X}_{1}\mathbf{\Phi}_{1}diag( \mathbf{b}_{1} )\mathbf{T}( \boldsymbol{\omega}_{DMD,1} ) )
\sqcup
( \mathcal{X}_{2}\mathbf{\Phi}_{2}diag( \mathbf{b}_{2} )\mathbf{T}( \boldsymbol{\omega}_{DMD,2} ) )
\label{eq:assemble_DMD}
\end{equation}

where each \(\mathbf{K}_{i}\boldsymbol{g}_{i}(x_{0})\) is an ergodic
partition approximated by
\(\mathbf{\Phi}_{i}\mathbf{\Lambda}_{i}^{t-1}\mathbf{b}_{i}\)
\cite{Nandanoori2019}. The missing element is the activation functions
\(\mathcal{X}_{i}\).

This fuzzy rule premises \(\varnothing_{i}(\mathbf{u})\) of LMN's can be
adapted s.t.
\({\varnothing_{i}( \mathbf{u} ) \propto \mathcal{X}}_{i}\).
\(\varnothing_{i}(\mathbf{u})\) smoothly interpolate among local models,
providing a weight of evidence in favour of a particular local model
\cite{Adeniran2017}. They are commonly parameterized by the set of input parameters
\(\mathbf{u = \lbrack 1,}\mathbf{u}_{1}\mathbf{,}\mathbf{u}_{2}\mathbf{,..,}\mathbf{u}_{H}\rbrack\)
representing individual axes \cite{Nelles2001}. These provide vector based
geometric locations for the respective activation weights, segmenting
the input space into activated regions divided by fuzzy interpolation
boundaries known as splits. \(\varnothing_{i}(\mathbf{u})\) is
parameterized here via a novel geometric perspective of the
representation space \(\mathbf{X}\in\mathbb{R}^{m \times t}\),
utilizing the column indexes as one input axis
\(\mathbf{u}_{t} = \lbrack 1,2,\ldots,t\rbrack\) (evolution), and the
row indexes as the other
\(\mathbf{u}_{m} = \lbrack 1,2,\ldots,m\rbrack^{\top}\) (spatial) s.t.
\(\mathbf{u} = [ \mathbf{u}_{m},\mathbf{u}_{t} ]\in\mathbb{R}^{H=2}\).
Utilizing sigmoidal activation functions \cite{Nelles2001}, axis-oblique splits
are generated. The full derivation is provided in SM-2, but essentially
each \(j^{th}\) split creates two sigmoid matrices,
\(\mathbf{\Omega}_{j}( \mathbf{U} )\in\mathbb{R}^{m \times t}\)
and its complement
\({\widetilde{\mathbf{\Omega}}}_{j}( \mathbf{U} )\in\mathbb{R}^{m \times t}\).
These correspond to two global validity/activation matrices
\(\varnothing_{i}( \mathbf{U} )\) and
\(\varnothing_{i + 1}( \mathbf{U} )\):

\begin{equation}
\varnothing_{i}( \mathbf{U} )
=
\mathbf{\Omega}_{j}( \mathbf{U} ), \ \
\text{and,}
\ \ {\varnothing_{i + 1}( \mathbf{U} )
=
{\widetilde{\mathbf{\Omega}}}_{j}( \mathbf{U} )
=
\mathbf{J}_{m,t}-\mathbf{\Omega}}_{j}( \mathbf{U} )
\label{eq:global_val}
\end{equation}

satisfying the conditions:

\begin{equation}
\mathbf{\Omega}_{j}( \mathbf{U} )
+
{\widetilde{\mathbf{\Omega}}}_{j}( \mathbf{U} )
=
\mathbf{J}_{m,t},\ \ 
\text{and,}
\ \ \sum_{i = 1}^{p}{\varnothing_{i}( \mathbf{U} ) = \mathbf{J}_{m,t}}
\label{eq:val_conditions}
\end{equation}

All \(\varnothing_{i}( \mathbf{U} )\) elements are thus a
subset of the total set \(\mathbf{J}_{m,t}\), which is a \(m \times t\) matrix of all ones. Via the Hadamard
product \cite{Searle2017}, \(\varnothing_{i}(\mathbf{U})\) assigns each element
of the corresponding local model a fuzzy activation weight between \(0 \leftrightarrow 1\)
relative to the global co-ordinates of the input \(\mathbf{X}\),
indicating which belong to one region or another s.t.:

\begin{equation}
\varnothing_{i}( \mathbf{U} )=\left\{ \begin{matrix}
1 & \ \text{if}\ x \in G( \mathcal{M}_{i} ) \\
0 < \varnothing_{i}( \mathbf{U} )<1 & \text{if}\ x \in G( \mathcal{M}_{i} ),G( \mathcal{M}_{g} ) \\
0 & \text{otherwise}
\end{matrix} \right.\ 
\label{eq:val_definition}
\end{equation}

where
\(x \in G( \mathcal{M}_{i} ),G( \mathcal{M}_{g} )\)
depicts the fuzzy overlap between two regions. Taking the DMD assumption
of \(\boldsymbol{g}_{i}( \mathbf{x} ) = \mathbf{x}\)
\cite{Tu2014} s.t. \(G( \mathcal{M}_{i} ) = \mathcal{M}_{i}\),
this produces:

\begin{equation}
\begin{split}
g_{s}(X) 
\approx
X 
=
[ \varnothing_{1}( \mathbf{U} )
\odot
( \mathbf{\Phi}_{1}diag( \mathbf{b}_{1} )\mathbf{T}^{t}( \boldsymbol{\omega}_{DMD,1} ) ) ]
\\ +
[ \varnothing_{2}( \mathbf{U} )
\odot
( \mathbf{\Phi}_{2}diag( \mathbf{b}_{2} )\mathbf{T}^{t}( \boldsymbol{\omega}_{DMD,2} ) ) ]
\label{eq:fSRD_eg}  
\end{split}
\end{equation}

where for the \(j = 1\) split,
\(\varnothing_{1}( \mathbf{U} )=\mathbf{\Omega}_{1}( \mathbf{U} )\)
and
\(\varnothing_{2}( \mathbf{U} )={\widetilde{\mathbf{\Omega}}}_{1}( \mathbf{U} )\).
Figure-S1 in SM-2 visually demonstrates the link between the original
matrix data, sigmoidal activation functions, activated regions, and 2-D
top-down view with normalized input scales between \(0 \rightarrow 1\).

\subsection{Invariant Decomposition}\label{subsec:invariant-decomposition}

The position and direction of a given sigmoidal split is manipulated by
altering the values within the coefficient vector \(\mathbf{v}_{j}\)
given
\(\mathbf{\Omega}_{j}( \mathbf{U} )=f( \mathbf{v}_{j} )\).
In traditional LMN applications (e.g. nonlinear multivariate response
surfaces), axis-oblique splits are often justified through improvements
in model parsimony by matching the direction of nonlinearity, i.e., less
models required if splits are orthogonal to the vector of nonlinearity
\cite{Shahzad2020}. In fSRD oblique partitions are re-purposed for the
segmentation of regions where different invariant dynamics apply. In
this way, splits aim to identify region boundaries within the underlying
system dynamics that correspond to limits of invariance, termed here as
invariant interfaces. As demonstrated in SM-3, these can be orthogonal
to each axis or oblique \cite{Nelles2001}. Orthogonal interfaces correspond to
purely temporal or spatial splits, aligning to phenomena such as
external forcing or bifurcated systems respectively. Such violations to
invariance may occur to isolated groups of spatial measurements at
differing time points however, generating oblique interfaces.

Such an interpretation assumes that spatial parameters (neighbouring
rows of \(X\)) are somewhat correlated per input region
\(G( \mathcal{M}_{i} )\), or ideally hold a continuous
relationship. If violated, a `straight' invariant interface as so far
discussed wouldn't represent `real' limits to invariance present in
\(\mathbf{X}\), perhaps requiring more complex geometries (e.g.,
curves). Consider the \(p = 4\) system in figure-1.

\begin{figure}
\centering
\includegraphics[width=\linewidth]{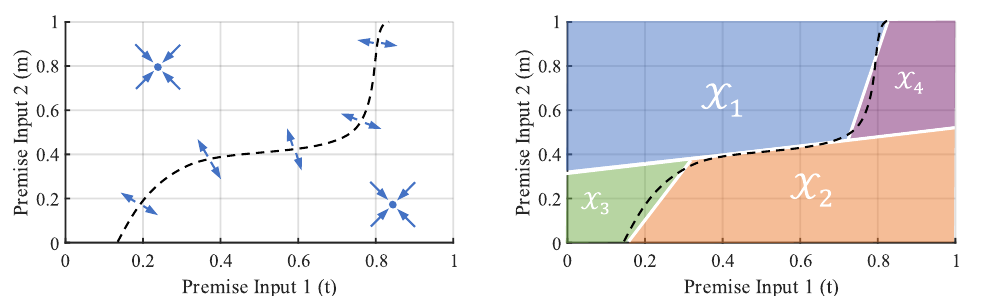}
\caption{: Left - Non-straight invariant interface example (black dashed
line = interface, blue arrows = dynamics vectors). Right - example
partitioning strategy.}
\end{figure}

If a lack of correlation occurs between subsets of the spatial
parameters, then an invariant interface may occur at separate time
points along the evolving dynamics. As in figure-1, further partitions
could be taken to segment the non-straight interfaces until a more
suitable estimate is reached. This is where the asymptotic decomposition
of the input space via traditional LMN architectures can be used,
implemented to Koopman via novel use of \eqref{eq:projection_value_measure}.

If multiple partitions of the observable space
\(G( \mathcal{M}_{i} )\) are demonstrably dynamically
invariant, this would conceptually align to localising the spectral
integral of \eqref{eq:projection_value_measure} within several sub-sets, e.g.,
\(\lbrack a,b\rbrack \subset \lbrack - \pi,\pi\rbrack\) \cite{Mezi2005,Mezi2004}.
Notably, one may take further restricted integrals even after
having already projected to a restricted observable. Consider a
candidate region \(i = 1,\ldots,p\) s.t. the interfaces resultant from a
candidate finite invariant region
\(\mathbf{K}_{i}\boldsymbol{g}_{i}( x_{0} )\) cannot be
characterized with a single oblique partition. Via \eqref{eq:projection_value_measure}, this
subspace/region of the full Hilbert space
\(\mathcal{M}_{i} = \mathcal{H}_{i}\mathcal{\subset H}\) where
\(\mathcal{K}_{i}^{\circ t}g_{i} \in \mathcal{H}_{i}\) may be presented as:

\begin{equation}
\begin{split}
\boldsymbol{g}_{i}(\mathbf{x}_t)
=
(\mathcal{K}_{i}^{\circ t}\boldsymbol{g}_{i})(\mathbf{x}_0)
\approx
\mathbf{K}_{i}\boldsymbol{g}_{i}( x_{0} ) 
=
\mathbf{\Phi}_{i}\mathbf{\Lambda}_{i}^{t-1}\mathbf{b}_{i}
\approx
\left(\int_{- \pi}^{\pi}{\exp(it\omega)1_{[ a.b]}d[ \mathbb{E}(\omega)\boldsymbol{g}_{i} ]}\right)(\mathbf{x}_{0})
\\ = \left(
\sum_{k}^{}{\exp(it\omega_{k})\mathbb{P}_{k,i}( \omega_{k} )\boldsymbol{g}_{i}} 
+ 
\int_{[ a,b]\backslash\omega_{k}}^{}{\exp(it\omega)d[ \mathbb{E}_{c,i}(\omega)\boldsymbol{g}_{i} ]}
\label{eq:K_to_restricted_integral}
\right)(\mathbf{x}_{0})
\end{split}
\end{equation}

where 
\(1_{\lbrack a.b\rbrack}d[ \mathbb{E}(\omega)\boldsymbol{g}_{i} ] = d[ \mathbb{E}_{i}(\omega)\boldsymbol{g}_{i} ]\)
is the restriction of the global projection value measure to the
subspace \(\mathcal{H}_{i}\), \((\mathcal{K}_{i}^{\circ t}\boldsymbol{g}_{i})(\mathbf{x}_0)\) 
denotes the Koopman-evolved observable evaluated at state \(\mathbf{x}_0\),
\(\mathbb{P}_{k,i}( \omega_{k} )\boldsymbol{g}_{i}\) is
the projection onto the eigenspace associated with the finite atomic
frequencies \(\omega_{k} \in \lbrack a,b\rbrack\), and
\(d[ \mathbb{E}_{c,i}(\omega)\boldsymbol{g}_{i} ]\)
for \(\omega \in \lbrack a,b\rbrack\backslash\ \omega_{k}\) is the
remaining part of the spectra not captured by any atomic projection
(i.e., continuous). Note \(1_{\lbrack a,b\rbrack}\) is the indicator
function that acts as a multiplicative filter to select only the portion
of the spectral measure lying in \(\lbrack a,b\rbrack\) s.t.:

\begin{equation}
1_{\lbrack a,b\rbrack}(\omega) = \left\{ \begin{matrix}
1,\ \ \text{if}\ \omega \in \lbrack a,b\rbrack \\
0,\ \ \text{if}\ \omega \notin \lbrack a,b\rbrack
\end{matrix} \right.\
\label{eq:indicator_f}
\end{equation}

Assuming \(j = 1,\ldots,f\) characterizable invariant subspaces of the
region \(\mathcal{M}_{i}\) exist, and taking the disjoint union \eqref{eq:assemble_regions},
the above proposal with \eqref{eq:K_to_restricted_integral} would suggest that:

\begin{equation}
\begin{split}
(\mathcal{K}_{i}^{\circ t}\boldsymbol{g}_{i})(\mathbf{x}_0) 
=
\left( \int_{- \pi}^{\pi}{\exp(it\omega)1_{\lbrack a.b\rbrack}d[ \mathbb{E}(\omega)\boldsymbol{g}_{i} ]} \right)(\mathbf{x}_0) 
\approx
\mathbf{K}_{i}\boldsymbol{g}_{i}( x_{0} )
\\ \approx
\lim_{j \rightarrow \infty}{( \bigsqcup_{j}^{\infty}{\mathcal{X}_{j}\mathbf{K}_{j}\boldsymbol{g}_{j}( x_{0} )} )\ }
=
\bigsqcup_{j}^{f}{\mathcal{X}_{j}[ \mathbf{\Phi}_{j}\exp( diag( \boldsymbol{\omega}_{DMD,j} )t )\mathbf{b}_{j} ]}
\label{eq:fSRD_projValue_connection}
\end{split}
\end{equation}

To be more precise, say the original sub-region
\(\mathcal{M}_{i}\ \)dictated by \(\mathcal{K}_{i}^{\circ t}\boldsymbol{g}_{i}\)
involved an orthogonal projection \(P_{\lbrack a,b\rbrack}\) on the
restricted spectral integral
\(\lbrack a,b\rbrack \subset \lbrack - \pi,\pi\rbrack\) from the global
\(\mathbf{K}_{s}\boldsymbol{g}_{s}\) as to satisfy the
following equality \cite{Quefflec2010}:

\begin{equation}
\begin{split}
\int_{-\pi}^{\pi} \exp(it\omega)\, 1_{\lbrack a,b\rbrack}\, d\left\langle \mathbb{E}(\omega)\boldsymbol{g}_{s},\boldsymbol{g}_{s} \right\rangle
&=
\int_{a}^{b} \exp(it\omega)\, d\left\langle \mathbb{E}_{i}(\omega)\boldsymbol{g}_{s},\boldsymbol{g}_{s} \right\rangle \\
&=
\left\langle \mathbf{K}_{s}P_{\lbrack a,b\rbrack}\boldsymbol{g}_{s},\, P_{\lbrack a,b\rbrack}\boldsymbol{g}_{s} \right\rangle
\label{eq:restricted_integral}
\end{split}
\end{equation}

where
\(d\left\langle \mathbb{E}(\omega)\boldsymbol{g}_{s},\boldsymbol{g}_{s} \right\rangle\)
is the induced scalar spectral measure corresponding to the Fourier
spectrum of observable \(\boldsymbol{g}_{s}\) \cite{Quefflec2010}, obtained by taking the inner
product of \eqref{eq:projection_value_measure} with \(\boldsymbol{g}_{s}\) s.t.:

\begin{equation}
\left\langle \mathbf{K}_{s}^{\circ t}\boldsymbol{g}_{s},\boldsymbol{g}_{s} \right\rangle 
=
\left\langle \int_{- \pi}^{\pi}{\exp(it\omega)d[ \mathbb{E}(\omega)\boldsymbol{g}_{i} ],\ \boldsymbol{g}_{s}} \right\rangle 
=
\int_{- \pi}^{\pi}{\exp(it\omega)d\left\langle \mathbb{E}(\omega)\boldsymbol{g}_{s},\boldsymbol{g}_{s} \right\rangle}
\label{eq:inner_product}
\end{equation}

Further arbitrary nested restrictions, say
\(\lbrack c,d\rbrack \subset \lbrack a,b\rbrack\), may then be taken for
an invariant subregion \(\mathcal{M}_{j} \subset \mathcal{M}_{i}\),
involving the orthogonal projection \(P_{\lbrack c,d\rbrack}\) s.t.
\(P_{\lbrack c,d\rbrack}P_{\lbrack a,b\rbrack} = P_{\lbrack a,b\rbrack \cap \lbrack c,d\rbrack}\)
satisfies:

\begin{equation}
\begin{split}
\int_{- \pi}^{\pi}{\exp(it\omega)1_{\lbrack c,d\rbrack}1_{\lbrack a,b\rbrack}d\left\langle \mathbb{E}(\omega)\boldsymbol{g}_{i},\boldsymbol{g}_{i} \right\rangle}
=
\int_{c}^{d}{\exp(it\omega)d\left\langle \mathbb{E}_{j}(\omega)P_{\lbrack a,b\rbrack}\boldsymbol{g}_{i},P_{\lbrack a,b\rbrack}\boldsymbol{g}_{i} \right\rangle}
\\ =
\left\langle \mathcal{K}_{i}^{\circ t}P_{\lbrack a,b\rbrack \cap \lbrack c,d\rbrack}\boldsymbol{g}_{i},\ P_{\lbrack a,b\rbrack \cap \lbrack c,d\rbrack}\boldsymbol{g}_{i} \right\rangle
\label{eq:further_restricted_integral}
\end{split}
\end{equation}

where
\(1_{\lbrack c,d\rbrack}1_{\lbrack a,b\rbrack}d\left\langle \mathbb{E}(\omega)\boldsymbol{g}_{i},\boldsymbol{g}_{i} \right\rangle = d\left\langle \mathbb{E}_{j}(\omega)P_{\lbrack a,b\rbrack}\boldsymbol{g}_{i},P_{\lbrack a,b\rbrack}\boldsymbol{g}_{i} \right\rangle\).
Eq's-(30) to (35) infer that both the global
\(\mathbf{K}_{s}\boldsymbol{g}_{s}( x_{0} )\) and any \(i^{th}\)
arbitrary region \(\mathcal{K}_{i}^{\circ t}\boldsymbol{g}_{i}\) can approximate there
atomic
\(\mathbb{P}_{k,i}( \omega_{k} )\boldsymbol{g}_{i}\) and
continuous spectra
\(d[ \mathbb{E}_{c,i}(\omega)\boldsymbol{g}_{i} ]\)
with further asymptotic partitioning via finite invariant regions.

This suggests that a global strategy of locating decomposed finite
invariant-regions with the sigmoid membership functions in \eqref{eq:global_val} could
estimate full orbit invariant regions \((I)\), continuous spectra
\((cS)\), and non-straight geometry interfaces \((G)\) all via a single
scheme, but without a designated means of decerning between them:

\begin{equation}
\begin{split}
\mathbf{K}_{s}\boldsymbol{g}_{s}( x_{0} )
=
( \bigsqcup_{}^{}{\mathcal{X}_{i}\mathbf{K}_{i}\boldsymbol{g}_{i}( x_{0} )} )_{I} 
+
( \bigsqcup_{}^{}{\mathcal{X}_{i}\mathbf{K}_{i}\boldsymbol{g}_{i}( x_{0} )} )_{cS} 
+
( \bigsqcup_{}^{}{\mathcal{X}_{i}\mathbf{K}_{i}\boldsymbol{g}_{i}( x_{0} )} )_{G} 
\\ \approx
\sum_{i = 1}^{p}{\varnothing_{i}( \mathbf{U} )\odot( \mathbf{\Phi}_{i}diag( \mathbf{b}_{i} )\mathbf{T}^{t}( \boldsymbol{\omega}_{DMD,i} ) )}
\label{eq:invariant_decomposition}
\end{split}
\end{equation}

This novel perspective is termed here as invariant decomposition,
requiring three main assumptions:

\begin{enumerate}
\def\labelenumi{\arabic{enumi})}
\item
  Locatable invariant sub-set estimates exist regardless of number of
  partitions.
\item
  The DMD assumption
  \(\boldsymbol{g}_{i}( \mathbf{x} ) = \mathbf{x}\) can
  characterize highly nonlinear dynamics when using asymptotic
  decomposition.
\item
  The membership functions \(\varnothing_{i}( \mathbf{U} )\)
  or model
  \(\mathbf{\Phi}_{i}diag( \mathbf{b}_{i} )\mathbf{T}^{t}( \boldsymbol{\omega}_{DMD,i} )\)
  in \eqref{eq:invariant_decomposition} can be amended s.t. DMD is applicable to non-square
  matrices in the presence of an axis-oblique split.
\end{enumerate}

For parsimonious results, leveraging assumption 1 requires partitioning
only continues until a given finite approximation proves sufficient.
This implies that further partitions can characterize invariant
interfaces while holding sufficient correlation between their own
spatial parameters. This point presents the practical data limits of
asymptotic linearization methods in many applications. One either
assumes sufficient localized correlation is present for the chosen
linearization method, or if not partition the local region further
\cite{Nelles2001,Bishop2006}.

Taken together, assumptions 1 and 2 raise a concern with respect to
interpretability. Namely whether identified
\(\boldsymbol{g}_{i}( \mathbf{x} ) = \mathbf{x}\)
sub-sets maintain system relevance as partitions increase (e.g.,
arbitrary fits like spurious eigenmodes). Ideally, w.r.t \eqref{eq:projection_value_measure},
locally restricted integrals with corresponding \(\boldsymbol{g}_{i}\) that
act as maxima for the Fourier measure should be prioritized \cite{Brunton2021}.
Otherwise, further restricted integrals should only occur to circumvent
non-straight geometry interfaces or decompose continuous spectra, acting
as a best fit estimate. Both this ideal and assumption 3 are tackled via
this works principal output, fSRD.

\subsection{fSRD - Principle of
Operation}\label{subsec:fsrd---principle-of-operation}

fSRD is a novel non-parametric fuzzy hierarchal tree network, automating
the identification of multiple Koopman invariant regions within a given
\(\mathbf{X}\) and combining them via fuzzy boundaries (i.e., invariant
interfaces). The output of the global model \(\widetilde{\mathbf{X}}\) can be written as the
aggregated contribution of several disjointed Koopman regions
\(i = 1,\ldots,p\):

\begin{equation}
\mathbf{X} \approx \widetilde{\mathbf{X}} 
=
\sum_{i = 1}^{p}{\varnothing_{i}( \mathbf{U} )\ 
\odot\ {\widetilde{f}}_{K,i}(\mathbf{X}_{i})}
\label{eq:fSRD}
\end{equation}

where \(\varnothing_{i}(\mathbf{U}) \in \mathbb{R}^{m \times t}\) are
the activation functions, \eqref{eq:global_val}, and
\({\widetilde{f}}_{K,i}(\mathbf{X}_{i}) \in \mathbb{R}^{m \times t}\)
are the globally defined Koopman region models of the corresponding
local representation space dubbed the region of interest
\(\mathbf{X}_{i} \in \mathbb{R}^{f \times q}\subseteq\mathbf{ X}\) s.t.
\({\widetilde{f}}_{K,i}:\mathbb{R}^{f \times q} \rightarrow \mathbb{R}^{m \times t},\ \ f \leq m\),
\(q \leq t\). The corresponding core network architecture is
demonstrated in figure-2-(a), where these disjointed models form each
neuron or local region. Utilizing a DMD spectral decomposition as in
\eqref{eq:DMD_decomp_matrix}, these are defined as:

\begin{equation}
\widetilde{f}_{K,i}( \mathbf{X}_{i}) 
=
\widetilde{f}_{K,i}({\vartheta_{i}^{- 1}\{\mathbf{\Phi}}_{i}diag( \mathbf{b}_{i} )\mathbf{T}( \boldsymbol{\omega}_{i} )\})
\label{eq:global_region}
\end{equation}

where each region \(i\) has its own DMD triplite
\(\{ \mathbf{\Phi,b,T}( \boldsymbol{\omega} ) \}_{i} \in \mathbb{R}^{f \times q}\)
and topological transformation \(\vartheta_{i}\).

\FloatBarrier

\begin{figure}[H]
\centering
\includegraphics[width=\linewidth]{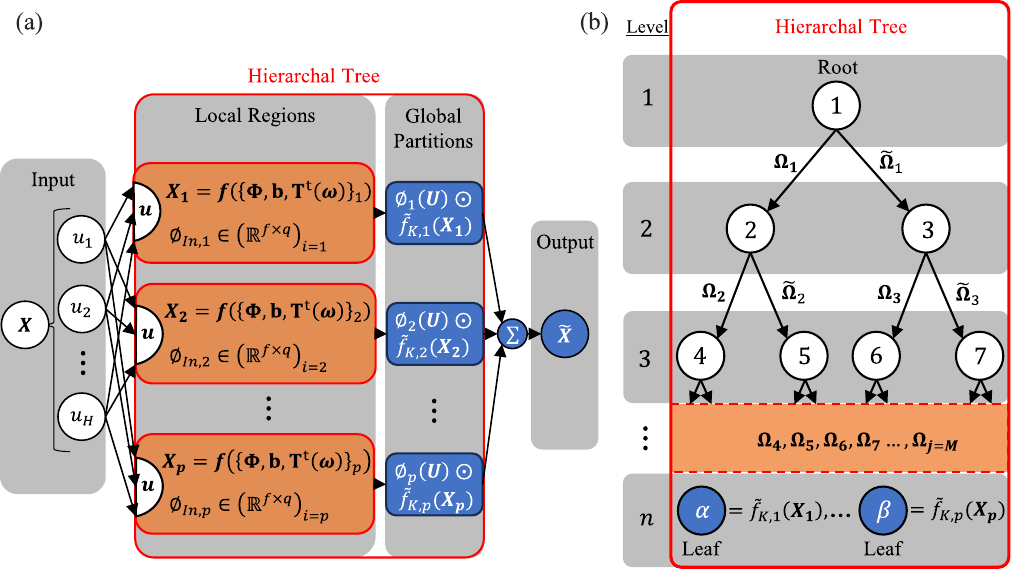}
\caption{: (a) fSRD network architecture. (b) Corresponding forward pass hierarchal tree (no pruning) where
\(\beta = \sum_{k = 1}^{n}2^{k - 1}\),
\(\alpha = ( \sum_{k = 1}^{n - 1}2^{k - 1} ) + 1\), and
\(M = \sum_{k = 1}^{n - 1}2^{k - 1}\).}
\end{figure}

\FloatBarrier

In fSRD a global Koopman operator \(\mathbf{K}_{s}\) is implicitly
defined, with the connection demonstrated in SM-4. fSRD's operation
conceptually aligns to the repeated application of restricted spectral
integrals described previously via invariant decomposition, where each
subsequent restricted set (child) is a subset of the previous
(parent/root), forming a hierarchy of ergodic partitions. This is
depicted in figure-2-(b), where each local model corresponds to a leaf.
Starting with the full set given by \(\mathbf{X}\), a root model is
fitted, and then each subsequent model/leaf is realized in pairs via
splits \(j = 1,\ldots,M\) of the previous levels leaves, starting with
this root. This results in the corresponding activation functions being
assigned as per \cite{Shahzad2020}:

\begin{equation}
\begin{aligned}
\text{Level 1:}\quad
\varnothing_{1}( \mathbf{U} )
&=
\mathbf{J}_{m,t}
\\[0.5em]
\text{Level 2:}\quad
\varnothing_{2}( \mathbf{U} )
&=
\mathbf{\Omega}_{1}( \mathbf{U} ),
\quad \text{and} \quad
\varnothing_{3}( \mathbf{U} )
=
\widetilde{\mathbf{\Omega}}_{1}( \mathbf{U} )
\\[0.5em]
\text{Level 3:}\quad
\varnothing_{4}( \mathbf{U} )
&=
\mathbf{\Omega}_{1}( \mathbf{U} )
\odot
\mathbf{\Omega}_{2}( \mathbf{U} ),
\\
\varnothing_{5}( \mathbf{U} )
&=
\mathbf{\Omega}_{1}( \mathbf{U} )
\odot
\widetilde{\mathbf{\Omega}}_{2}( \mathbf{U} ),
\\
\varnothing_{6}( \mathbf{U} )
&=
\widetilde{\mathbf{\Omega}}_{1}( \mathbf{U} )
\odot
\mathbf{\Omega}_{3}( \mathbf{U} ),
\ldots
\\
\ldots \text{and} \quad
\varnothing_{7}( \mathbf{U} )
&=
\widetilde{\mathbf{\Omega}}_{1}( \mathbf{U} )
\odot
\widetilde{\mathbf{\Omega}}_{3}( \mathbf{U} )
\\[0.5em]
\text{Level 4:}\quad
\text{etc.\ldots}
\end{aligned}
\label{eq:hierarchical_operators}
\end{equation}

Via Hadamard products, both
\({\widetilde{f}}_{K,i}( \mathbf{X}_{i} )\) and
\(\varnothing_{i}( \mathbf{U} )\) can be resolved in
vectorized form for compact coded implementation:

\begin{equation}
\begin{aligned}
vec( {\widetilde{\mathbf{X}}}_{global} )
=
\sum_{i = 1}^{p}{vec( \varnothing_{i}( \mathbf{U} ) )\ 
\odot\ vec( {\widetilde{f}}_{K,i}(\mathbf{X}_{i}) )\ }
\\
\text{E.g.,}\ \ vec( \varnothing_{4}( \mathbf{U} ) )
=
vec( \mathbf{\Omega}_{1}( \mathbf{U} ) )\ 
\odot\ vec( \mathbf{\Omega}_{2}( \mathbf{U} ) )
\label{eq:fSRD_vector_form}
\end{aligned}
\end{equation}

This can be extended to much of the global tree and local model
construction but is omitted here to maintain interpretability. Note the
final output of \eqref{eq:fSRD} only includes partitions that correspond to
chosen active regions (ends of branches), not including any of their
roots/parents.

\subsection{Network Architecture}\label{subsec:network-architecture}

Portions of fSRD's architecture are adaptations or direct inserts from
neuro-fuzzy LMN literature to achieve the described Koopman scheme.
While a comprehensive review is provided in \cite{Nelles2001,Adeniran2017},
fundamentals of LMN's will only be referenced as needed. A key
motivation for adopting Takegi-Sugeno fuzzy logic \cite{Takagi1985} based LMN
architecture was due to its structure optimization. Given the ability to
arbitrarily take restricted spectral integrals in eq's-(30) to (36),
this provided a means of searching for maxima invariant region
combinations without prior system knowledge, evaluating and comparing
several candidate ensembles upon the data \cite{Babuka2003}. This alone is not
sufficient to choose a method, with several elements of structure
assembly and optimization producing highly varied architectures. The
novel reasoning behind many of the fSRD's architectural choices will
only be covered here for function, with further information
provided in \cite{BokorThesisfSRD} and the references within.

\subsection{Structure Optimization}\label{subsec:structure-optimization}

W.r.t LMN literature, fSRD is a locally estimated hierarchal tree
construction (structure), trained via a semi-greedy fully specified
forward pass (assembly) and a subsequent information-criteria based
pruning (selection). The former searches for candidate ergodic
partitions, while the latter selects which of these should be used in
the final output \eqref{eq:fSRD}. This assembly and selection are both adapted
from PRUHINET \cite{Shahzad2020}, aligning fSRD with backwards elimination schemes
in the context of LMN's \cite{Banfer2010}.

\subsubsection{Forward Pass}\label{subsubsec:forward-pass}

First, splits for all regions in each iteration/level are defined via a
local selection criterion, creating a candidate set of cost function
minima \cite{Shahzad2020}. During training, this creates an exponential increase
in global model complexity at each \(n^{th}\) iteration, specifically
\(2^{n - 1}\) \cite{Shahzad2020}. That said, the use of identical local model and
validity function definitions at each level of the tree enables parallel
computing techniques during training \cite{Shahzad2020}. Generating this set
requires the selection of the best fitted split for each region rather
than allowing a worse split in a region from an earlier iteration even
if it would enable a superior fit via a resultant region later \cite{Bnfer2012,BanferLMN2010}.
This path-based dependency for each region corresponds to a
greedy search scheme \cite{Bnfer2012,Banfer2009}. LMN's like HILOMOT \cite{Nelles2006} are
fully greedy by only assigning a best split for a single region per
iteration of the forward pass, emulating a binary tree. By
contrast, PRUHINET fully specifies the tree (i.e., generating a split
for every region per iteration), implementing a spread-out search to
explore a much larger portion of the possible solution space (partition
combinations) \cite{Shahzad2020}. By still selecting only one split for each
region, PRUHINET can thus be framed as semi-greedy. SM-5 (figure-S3)
visually depicts this comparison. The following novel local estimate
heuristic cost function serves to evaluate candidate partitions for
selection:

\begin{equation}
NRMSE_{i} = \frac{RMSE_{i}}{(Mean\ Diff)_{i}} 
=
\frac{\left\| \vartheta(\mathbf{X}_{i}) 
-
\mathbf{\Phi}_{i}diag( \mathbf{b}_{i} )\mathbf{T}( \boldsymbol{\omega}_{DMD,i} ) \right\|_{F}}{\left\|
\vartheta(\mathbf{X}_{i}) 
-
\mu\mathbf{J}_{f,q} \right\|_{F}}
\label{eq:NRMSE}
\end{equation}

The full derivation, symbology, and justification for \eqref{eq:NRMSE} is
provided in SM-6. To optimize any given splits position, an aggregated
measure is required to evaluate the pair of generated partitions. For
all currently active partitions \(i = 1,\ldots,p\), fSRD uses:

\begin{equation}
Net\ Error = WNRMSE = \sum_{i}^{p}{w_{i} \cdot (NRMSE_{i})}
\label{eq:WNRMSE}
\end{equation}

where \(w_{i}\) are scalar region weights calculated as:

\begin{equation}
w_{i} = \frac{( \sum_{k = 1}^{m}{\sum_{j = 1}^{t}\alpha_{k,j}} )_{\varnothing_{i}}}{mt}
\label{eq:WNRMSE_weights}
\end{equation}

s.t. \(\alpha_{k,j} \in \varnothing_{i}(U)\) are the elements of region
\(i\)'s global activation function
\(\varnothing_{i}(U) \in \mathbb{R}^{m \times t}\), satisfying the
conditions:

\begin{equation}
0 \leq w_{i} \leq 1,\ \ \text{and,}\ \ \sum_{i = 1}^{p}w_{i} = 1
\label{eq:weights_conditions}
\end{equation}

Each \(w_{i}\) generates a fraction of unity proportional to the
region's global representation utilizing each \(\varnothing_{i}(U)\)
whose elements \(0 \leq \alpha_{k,j} \leq 1\) are already assigned
across the \(\mathbb{R}^{m \times t}\). In this way fSRD via \eqref{eq:WNRMSE}
searches for increased Koopman invariant estimates, but rewards larger
global representation, penalizing arbitrary restricted subsets
encompassing less input space.

\subsubsection{Split optimization}\label{subsubsec:split-optimization}

The values
\(\mathbf{v}_{j} = [ v_{j,0},\ v_{j,1},\ \ldots,v_{j,H} ]^{\top}\)
that determine split position are defined via optimization, with many
LMN's favouring fast implementation via local routines such as grid
search, heuristics, or gradient based methods in PRUHINET and HILOMOT
\cite{Nelles2001,Shahzad2020,Nelles2006,Mahfuz2022}. By utilising a semi-greedy
hierarchal search scheme, fSRD's interpretability relies on earlier
splits/levels to sufficiently explore the solution space for dominant
invariant regions (regions with the largest coverage, or containing full
orbits, most physically relevant, etc.). Locking to initial splits with
small coverage early irreversibly removes these from the explorable
solution space. Further, given the input sigmoidal function from SM-2:

\begin{equation}
\Omega_j(u)
=
\frac{1}{1+\exp\!\left(-\tau_j \mathbf{v}_{j} u^{\top}\right)}
\\[6pt]
\ \ \text{s.t.}\quad
\mathbf{v}_{j}
=
\left[
v_{(j,0)},
\tilde{v}_{(j,1)},
\tilde{v}_{(j,2)}
\right]
\\[6pt]
,
\tilde{v}_{(j,\ell \neq 0)}
=
\frac{v_{(j,\ell)}}{\operatorname{med}\!\left(v_{(j,\ell)}\right)}
\label{eq:split_cost_function}
\end{equation}

and the cost function, \eqref{eq:WNRMSE}, the split optimization space is likely to be nonlinear
and/or non-convex, changing based on the dynamic system under scrutiny. Note that
constraining the coefficient vector \( \mathbf{v}_{j} \) to be of unit length via the median 
division above, \( \operatorname{med}\!\left(v_{(j,\ell)}\right) \), standardizes the effect of changing \( \mathbf{v}_{j} \) across the input 
space \cite{Shahzad2020}. With these considerations, global optimization methods appear preferable even 
at the cost of longer computational time \cite{Mahfuz2022}. For this proof of concept, 
particle swarm optimization was used to optimize the full vector \(\mathbf{v}_{j}\) with constraints to maintain robust
implementation \cite{Rios2013}. As a metaheuristic, PSO requires few to no
assumptions on the solution space \cite{LeoLiberti2006}, maintaining the minimal
system prior mandate and thus generalisability of fSRD. No calculation
of a gradient is required \cite{Rios2013}, so a lack of differentiability that
can come with discontinuous or chaotic dynamics is also circumvented.
While PSO is generally slower than grid search techniques, it presents a
mid-ground where the global solution space is efficiently explored
without being as slow as methods such as genetic algorithms \cite{Rios2013}.
More details on the exact implementation used here can be found here
\cite{BokorThesisfSRD} and SM-22, but it is expected by the authors that a more
efficient implementation is readily achievable.

\subsubsection{Backwards Elimination}\label{subsubsec: backwards-elimination}

Once generated, the fully specified tree is then pruned to select final
active regions for the output \eqref{eq:fSRD}. Pruning here involves replacing a
given pair of child models resultant from a split with their parent
model. This is evaluated by comparing the global Information Theoretic
Criteria (ITC) contribution of the two options \cite{Tangirala2015,Shahzad2020}. The
child leaves are only retained if they contribute sufficiently to a
chosen cost function given there added complexity. This can be
considered a bottom-up traversal strategy \cite{Shahzad2020}. A visual example is
provided in SM-7 (figure-S4). Any inidivdual prune never removes entire branches,
rather systematically removing individual parent-children groups
throughout one level before proceeding to the next.

W.r.t Koopman, early levels of the tree represent the largest candidate
invariant regions, with the chance of arbitrary restricted integrals
increasing as levels progress. Matching the ideal from Section~\ref{subsubsec:split-optimization},
the scheme described implements full invariant decomposition to explore
possible combinations, then culls any arbitrary partitions, inherently
prioritizing full invariant regions. Without evaluating all possible
arrangements, the true bias/variance trade-off for model assembly is an
unknowable quantity in most instances, but this scheme provides a
balance between computational complexity and exploration. In fSRD, this
pruning is implemented largely as described by PRUHINET \cite{Shahzad2020},
although with one novelty/alteration. PRUHINET uses
AICc \cite{Sugiura1978}, but this is replaced here by BIC adapted for spectral
models \cite{Schwarz1978}:

\begin{equation}
BIC = - 2Ln( \mathcal{L}_{m} ) + kLn(N)\
\label{eq:BIC}
\end{equation}

where \(N = mt\), \(k = f( r_{i} )\) is the spectral
complexity measure, and \(Ln( \mathcal{L}_{m} )\) can be
interpreted as the profile likelihood when using a maximum
log-likelihood having eliminated variance as a nuisance parameter
\cite{DouglasCMontgomery2012}. Full derivation is provided in SM-8. BIC was chosen
predominantly to discourage the overfitting of AIC \cite{Goutte1997}, while also
being simpler than AICc, overall opting for less regions total to again
emphasise larger regions.

\subsubsection{Termination criteria}\label{subsubsec: termination-criteria}

For the forward pass of a fully specified LMN, the primary convergence
criterion is traditionally concerned with identifying the number of
levels at which the assessed data is sufficiently predicted, rather than
the number of splits (i.e., convergence via global predictive accuracy)
\cite{Banfer2009,Nelles2006,Nelles1996,HartmannSUHICLUST2009}. This creates a subtle difference
in goal between tree construction and the local model selection. The
former gauges global information representation, while the latter
identifies and assigns weighting to valid invariant regions. The use of
an optimization routine for the latter allows for separate cost
functions that better suit each intent, hence \eqref{eq:WNRMSE} for split optimization. For tree
construction, \eqref{eq:WNRMSE} is viable but requires the
specification of a residual or iterative difference magnitude given no
guarantee of locating a minimum \cite{IanGoodfellow2016}. ITC's by contrast hold an advantage
here due to the exponential nature of LMN's providing a clear minimum
\cite{Shahzad2020}. Given the unchanging number of samples, the absolute BIC value
is largely irrelevant, but the difference measure
\(\mathrm{\Delta}_{i}\) may be used:

\begin{equation}
\mathrm{\Delta}_{i} = BIC_{i} - BIC_{\min}\
\label{eq:diff_measure}
\end{equation}

where \(BIC_{i}\) is the value of \eqref{eq:BIC} for the \(i^{th}\)
iteration/level of the forward pass, and \(BIC_{\min}\) the lowest value
so far identified. Level iteration utilizing \(\mathrm{\Delta}_{i}\)
continues until the difference measure becomes positive, indicating an
increase in ITC. This behaviour is demonstrated within PRUHINET
\cite{Shahzad2020}, though via AICc.

During forward pass, the final iteration that initiates termination is
unlikely to land squarely on the expected minima, rather overshooting
it. This is due to the several likely unnecessary splits/regions
exponentially generated within each level/iteration. The forward pass
thus acts as a rough convergence measure from a purely information
content perspective, while the backwards pass refines the tree structure
closer to the actual ITC minima. In this way \eqref{eq:diff_measure} is again used but
in reverse, reviewing each child-parent group in a level individually
before evaluating the next, i.e., comparing the two possible tree models
with and without the parents pair of children \cite{Shahzad2020}. Note that
several tree model configurations may be possible dependant on the
backwards pass procedure discussed here \cite{Shahzad2020}, but further
investigation is left for future work.

\subsubsection{Robust Adaptation}\label{subsubsec:robust-adaptation}

The described structure optimization approach allows fSRD to assess each
data set individually and adapt its architecture to prioritize regions
with strong Koopman invariance. In many applications sufficient training
data to identify full invariant regions is not always available, with
full information on some attractors, while only partial orbits of
others. Given the hierarchal forward backward design, areas that do not
admit such representation will inherently be isolated and characterized
via only sufficient further restricted spectral integrals to circumvent
limits to invariance. This unavoidably loses physical interpretability
compared to the non-represented `true' full orbit but provides
robustness in the face of insufficient data. This introduces another
category of non-discernible partition, i.e., partial orbits (pO), into
\eqref{eq:invariant_decomposition}:

\begin{equation}
\begin{split}
\mathbf{K}_{s}\boldsymbol{g}_{s}( x_{0} )
=
( \bigsqcup_{}^{}{\mathcal{X}_{i}\mathbf{K}_{i}\boldsymbol{g}_{i}( x_{0} )} )_{I}
+
( \bigsqcup_{}^{}{\mathcal{X}_{i}\mathbf{K}_{i}\boldsymbol{g}_{i}( x_{0} )} )_{cS}
\\ +
( \bigsqcup_{}^{}{\mathcal{X}_{i}\mathbf{K}_{i}\boldsymbol{g}_{i}( x_{0} )} )_{G}
+
( \bigsqcup_{}^{}{\mathcal{X}_{i}\mathbf{K}_{i}\boldsymbol{g}_{i}( x_{0} )} )_{pO}
\label{eq:invariant_decomposition_pO}
\end{split}
\end{equation}

\subsection{Locally Invariant Spectral
Models}\label{subsec:locally-invariant-spectral-models}

The automated systematic global strategy for invariant decomposition so
far described provides the rule premises with respect to Takegi-Sugeno
fuzzy logic \cite{Takagi1985}. However, following assumption 3 from Section~\ref{subsec:invariant-decomposition},
no automated strategy to fit locally linear spectral embedding estimates
(i.e., rule consequents) exists for such a scheme. What follows in this
section are several contributions to resolve this.

\subsubsection{Local Region
Interpretation}\label{subsubsec:local-region-interpretation}

Two definitions are used in this study for the local regions
\(i = 1,\ldots,p\) dictated by differing finite operators
\(\mathbf{K}_{i}\); the orthogonal input region \(\mathbf{X}_{In,i}\),
and the true region of interest \({\overline{\mathbf{X}}}_{i}\) (notably
a set, not a matrix). This concept is demonstrated in figure-3.

The orthogonal input region \(\mathbf{X}_{In,i}\) is defined by the rows
and columns of \(\mathbf{X}\) that match the extremities in the
corresponding validity function co-ordinates, i.e., the region of
\(\mathbf{X}\) where any rows and columns of
\(\mathbf{\varnothing}_{i}(\mathbf{U})\) have non-zero values are
included (defined against the global co-ordinates). Any rows or columns
with only zero value elements in \(\mathbf{\varnothing}_{i}\) are
excluded. Due to the sigmoid function being continuous, all neighbouring
rows and columns of \(\mathbf{\varnothing}_{i}(\mathbf{U})\) should be
included until a full zero element row/column occurs at the base of the
s-curve. In this way \(\mathbf{X}_{In,i} \in \mathbb{R}^{f \times q}\)
is defined precisely as a sub-matrix of
\(\mathbf{X} \in \mathbb{R}^{m \times t}\) \cite{Horn2012}, such that
\(f \leq m\), \(q \leq t\), and
\(\mathbf{X}_{In,i}\subseteq\mathbf{X}\).

The position of \(\mathbf{X}_{In,i}\) relative to the full data matrix
\(\mathbf{X}\) is dictated by the column index extremities of
\(\mathbf{\varnothing}_{i}\), namely \(m_{min - i}\), \(m_{max - i\ }\),
\(t_{min - i}\) and \(t_{max - i}\), s.t.
\(0 \leq ( m_{min - i},\ m_{max - i} ) \leq m\) and
\(0 \leq ( t_{min - i},\ t_{max - i} ) \leq t\)
(figure-3-(b)). A cut-off value \(\eta\) is used to determine non-zero
elements of \(\mathbf{\varnothing}_{i}\), and by extension their indexes
(i.e., value is non-zero if \(> \eta\), and all elements below are set
to zero). \(\eta\) is implemented for numerical stability given the
continuous nature of the sigmoid (see SM-2), potentially creating
arbitrarily small non-zero values in \(\varnothing_{i}\) at the base of
the s-curve.

\begin{figure}[!ht]
\centering
\includegraphics[width=\linewidth]{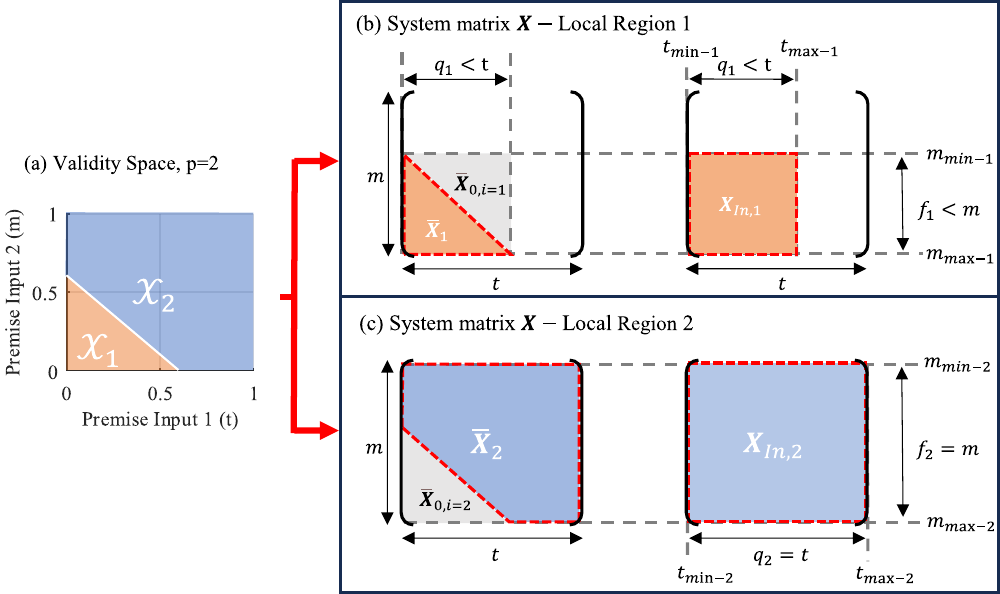}
\caption{: (a) Example split in top-down premise space. (b) \& (c)
local \({\overline{\mathbf{X}}}_{i}\) and \(\mathbf{X}_{In,i}\) co-ordinates
for each region post-split relative to the global co-ordinates.}
\end{figure}

The true region of interest \({\overline{\mathbf{X}}}_{i}\) is defined
as one of the two exact sub-regions (sets) of \(\mathbf{X}\) created by
the discretization process, i.e., all values within \(\mathbf{X}\) that match the
global indexes of all \(\varnothing_{i}(\mathbf{U})\) elements
\(> \eta\). In the intended 2-D input space,
\({\overline{\mathbf{X}}}_{i}\) is non-square when created by an oblique
split and thus cannot be neatly defined as a sub matrix of
\(\mathbf{X}\) as depicted in figure-3. Rather, by the definitions
provided, \(\mathbf{X}_{In,i}\) serves as the smallest possible
corresponding sub matrix to which \({\overline{\mathbf{X}}}_{i}\) is a
sub-set, i.e.,
\({\overline{\mathbf{X}}}_{i} \subseteq \mathbf{X}_{In,i} \subset \mathbf{X}\).
\({\overline{\mathbf{X}}}_{i}\) is thus located globally via the
row/column index extremities of \(\mathbf{X}_{In,i}\), enabling
extraction for local model fitting/processing.~As outlined, this
framework for \(\mathbf{X}\), \({\overline{\mathbf{X}}}_{i}\), and
\(\mathbf{X}_{In,i}\) produces several properties that are relevant to
the operation/behaviour of fSRD. A non-exhaustive list is provided in
SM-9.

\subsubsection{Oblique Matrix
Decomposition}\label{oblique-matrix-decomposition}

Assuming the use of exact DMD \cite{Tu2014} from SM1 to
form local finite Koopman estimates, a matrix decomposition of
\(\widetilde{\mathbf{A}}\) is required. As this forms the orthogonal
basis in which to perform eigen-decomposition, its dimensionality
corresponds to that of the resulting local spectral models. As a linear
method, SVD is favourable here in both speed and simplicity but has
issues regarding its dependence on the data co-ordinates or `alignment'.
As explained in \cite{BruntonBook2019}, SVD can greatly inflate the dimensionality of
the resultant spectra if any misalignment between the recorded
co-ordinates and the data exists. This is due to SVD's preservation of
the inner product, creating a dependency on the data coordinates.
Historically this has been handled by data pre-processing or methodology
extensions covered in \cite{GeneHGolub2013}. Figure-4 demonstrates one ideal solution
via an adaptation of the square example in \cite{BruntonBook2019}, applying a transform
\(\vartheta_{e}\) that retains important traits of the data while
collapsing the SVD spectra. In this case however this as achieved via
prior system information to align the square with the matrix
co-ordinates/axis.

The potential for oblique splits creating non-square
\({\overline{\mathbf{X}}}_{i}\) conflicts with this dependence,
artificially inflating the dimensionality of the resultant SVD when the
regions split and data axis are misaligned. This diminishes the impact
of the parsimonious embeddings in the global model but also creates
issues for the eigen decomposition of candidate oblique regions. The
latter stems from the different means one can interpret SVD's
application in this scheme. Discussion on this is provided in SM-10, but
moving forward the sequence below will be assumed:

\begin{enumerate}
\def\labelenumi{\arabic{enumi}.}
\item
  Extract
  \(\mathbf{X}_{In,i} = {\overline{\mathbf{X}}}_{i} \cup {\overline{\mathbf{X}}}_{0,i}\)
\item
  Create the region of interest matrix \(\mathbf{X}_{i}\) s.t.,
  \(\mathbf{X}_{i} = {\overline{\mathbf{X}}}_{i} \cup {\overline{0}}_{i}\) as per SM-9
\item
  Perform economic SVD s.t.,
  \(\mathbf{X}_{i}\mathbf{=}{\widetilde{\mathbf{U}}}_{i}{\widetilde{\mathbf{\Sigma}}}_{i}{\widetilde{\mathbf{V}}}_{i}^{\mathsf{H}}\)
\item
  Output \({\overline{\mathbf{X}}}_{i} \approx \mathbf{X}_{i}\)
\end{enumerate}

\begin{figure}[!ht]
\centering
\includegraphics[
    width=13cm,
    height=8cm
]{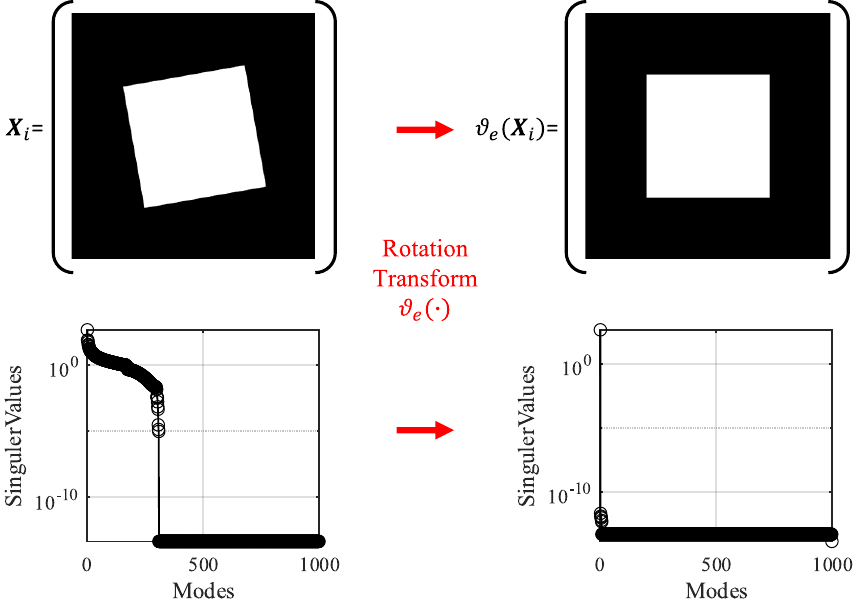}
\caption{: SVD alignment issues corrected via rotational transform. Adapted from \cite{BruntonBook2019}.}
\end{figure}

By this definition, the alignment issue can be interpreted as the
unwanted contribution of \({\overline{0}}_{i}\) within
\(\mathbf{X}_{i}={\widetilde{\mathbf{U}}}_{i}{\widetilde{\mathbf{\Sigma}}}_{i}{\widetilde{\mathbf{V}}}_{i}^{\mathsf{H}}\)\textbf{.}
Much like figure-4, a transform \(\vartheta\) that retains the key
traits of \({\overline{\mathbf{X}}}_{i}\) while circumventing
\({\overline{0}}_{i}\) contribution is thus ideal. Automating the
selection of a transform to achieve this for an arbitrary set of data
has two additional challenges:

\begin{itemize}
\item
  A given transform may alter the original data structure, losing some
  measure of distances between the elements within
  \({\overline{\mathbf{X}}}_{i}\) relative to those not, i.e.,
  \(\mathbf{X}\ \backslash\ {\overline{\mathbf{X}}}_{i} = \{ x \in \mathbf{X}:x \notin {\overline{\mathbf{X}}}_{i}\}\).
\item
  One assumes that \({\overline{\mathbf{X}}}_{i}\) within a given
  \(\mathbf{X}_{i}\) sufficiently describes a region as to capture a
  minimal dimensional embedding.
\end{itemize}

The former can be tackled by selecting a transform that is invertible,
but the latter is primarily an issue of data quality for purpose, with
the local partitions \({\widetilde{f}}_{K,i}(\mathbf{U})\) always being
susceptible due to their more than not shared boundary with
\(\mathbf{X}\). E.g., if only half the square of figure-4 was in the
full data snapshot \(\mathbf{X}\) as a triangle, no rotatable position
would achieve as low-dimensional a decomposition as the original square
in the correct co-ordinates. Assuming prior information is not available
to correct the data, finding the `true' lowest dimensional embedding to
represent the underlying system may thus be un-achievable. However,
minimising required model dimensionality, while retaining physically
meaningful order/structure within a limited dataset of said system is
achievable.

\subsubsection{Local Topological
Transformation}\label{subsubsec:local-topological-transformation}

The solution proposed here involves applying a novel topological
transform \(\vartheta( \cdot )\) that stretches the sub-region
\({\overline{\mathbf{X}}}_{i} \subset \mathbf{X}_{i}\) to match the
dimensionality of the smallest possible matrix formulation
\(\mathbf{X}_{In,i} \in \mathbb{R}^{f \times q}\) s.t.
\(\vartheta(\mathbf{X}_{i}) \in \mathbf{Y}_{i} \subseteq \mathbb{R}^{f \times q}\).
Consequently, one can frame \(\vartheta( \cdot )\) as a mapping that
transitions \(\mathbf{X}_{i}\) to a new space \(\mathbf{Y}_{i}\) s.t.
\(\mathbf{Y}_{i} = \vartheta( \mathbf{X}_{i} )\) and
\(\vartheta:\mathcal{M}_{x} \rightarrow \mathcal{M}_{y}\). The elements
of each row of \(\mathbf{X}_{i}\) that correspond to the set
\({\overline{\mathbf{X}}}_{i}\) are interpolated or `stretched' along
the maximum row length \textquotesingle{}\(q\)\textquotesingle{} as to
replace all elements of \({\overline{0}}_{i}\) within \(\mathbf{X}_{i}\).
Figure-5 demonstrates this process on intuitive examples. In this way
\(\vartheta( \cdot )\) retains the information content of the region
of interest along the evolution dimension (rows) by allowing the
relative distances between points in the spatial dimension (columns) to
be changed though not disconnected, thus preserving in part the datas
order/structure.

\begin{figure}[!ht]
\centering
\includegraphics[width=\linewidth]{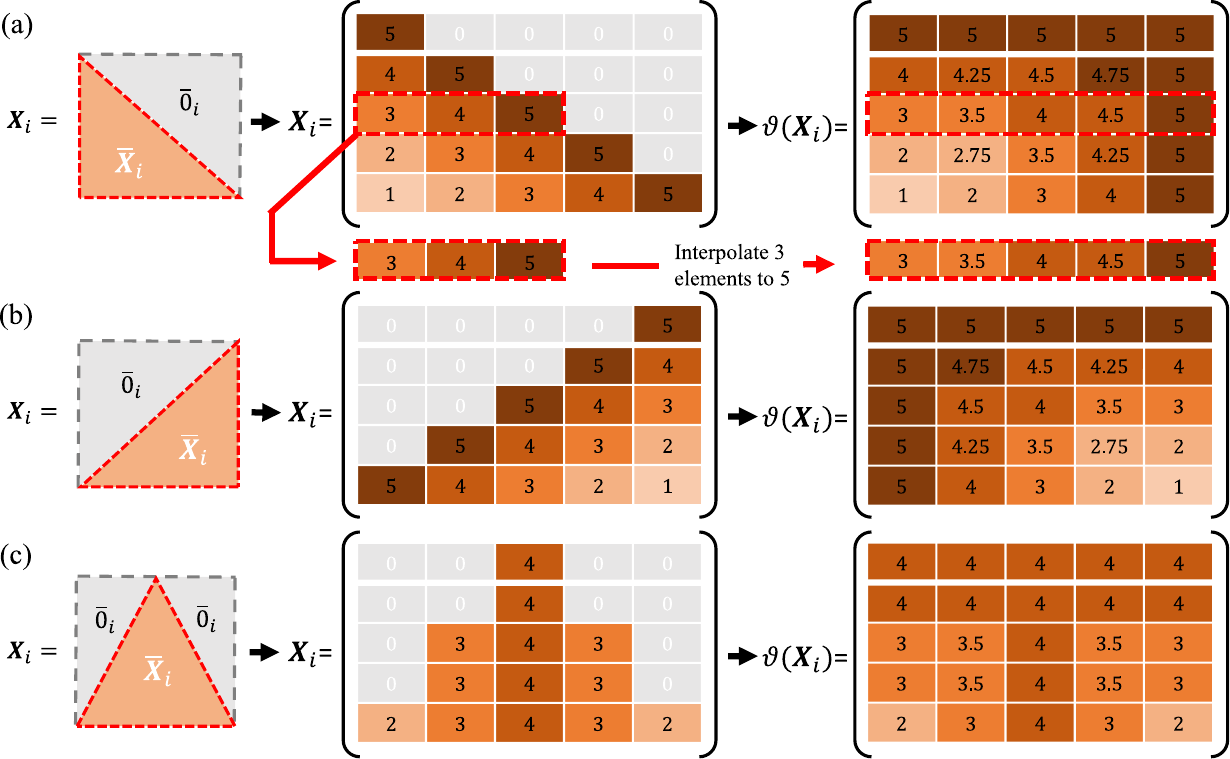}
\caption{: The topological transform \(\vartheta( \cdot )\) applied to
three typical true regions of interest examples given the use of
sigmoidal axis oblique splits}
\end{figure}

\begin{figure}[!ht]
\centering
\includegraphics[width=\linewidth]{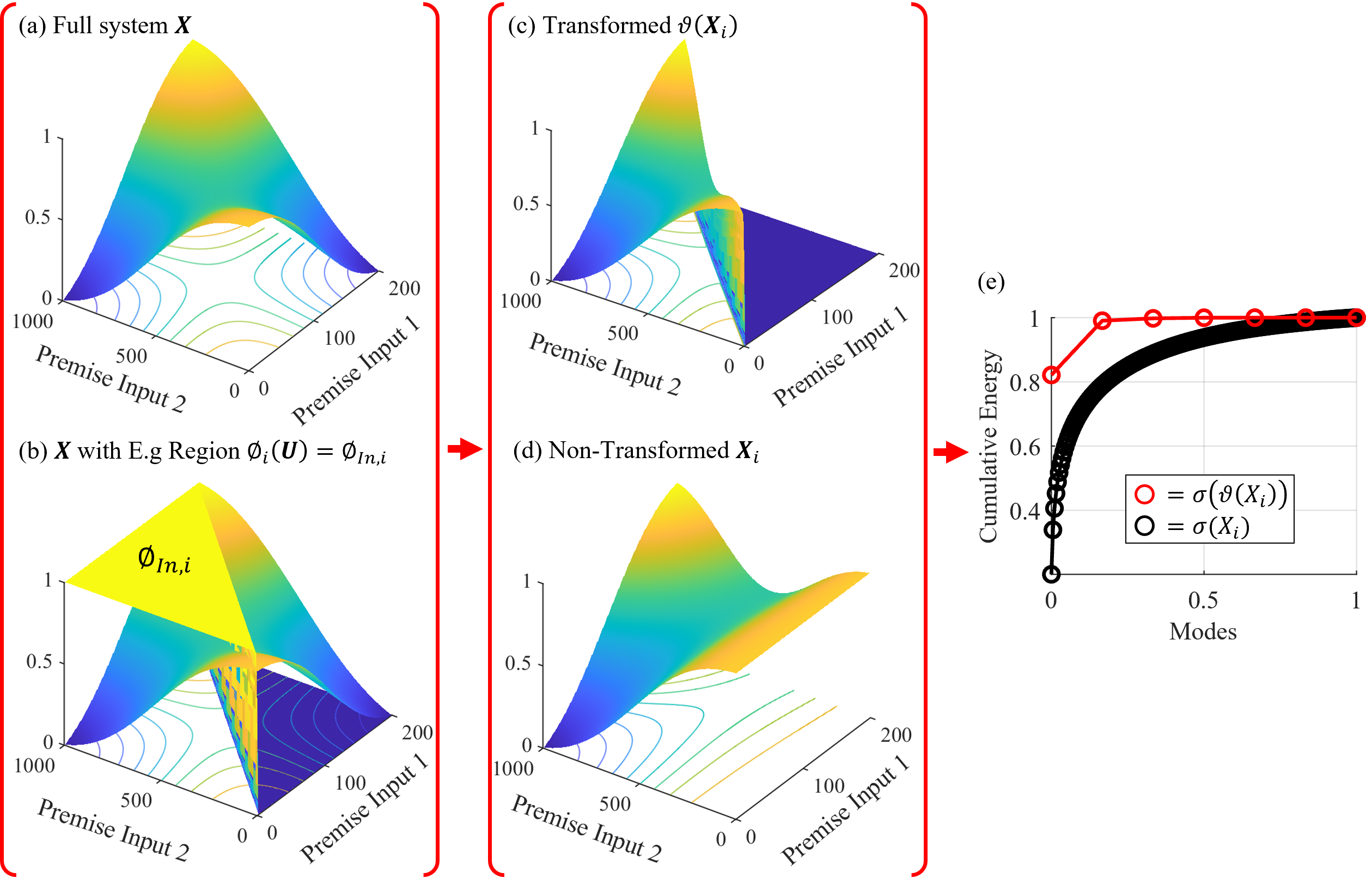}
\caption{: Comparison of resultant spectra (e), when applying SVD
with the topological transform (c), and without (d), on an example
oblique region \(v = \lbrack 0, - 0.5,5\rbrack\) (b), taken from a
bistable toggle quassia potential surface (a) \cite{Verd2014}.}
\end{figure}

As these local models seek to identify invariant regions, an ideal
\({\overline{\mathbf{X}}}_{i}\) undergoing \(\vartheta( \cdot )\) will
be continuous and spatially correlated. Given the preservation of
structure, one would expect the new region \(\mathbf{Y}_{i}\) would then also be
continuous and spatially correlated but in slightly different
co-ordinates. A continuous, well-posed, and forward invariant region is
also the ideal for assessment by linear decomposition's such as SVD
\cite{Tikhonov1977,Turco2017,Staffans2005}. In this way, locating a well-suited region
for performing SVD on \(\vartheta(\mathbf{X}_{i})\) also encourages in
part the ideal behaviour for the split optimization itself. This
approach does not attempt to fully characterize the underlying data
structure within \(\mathbf{X}_{i}\) before selecting a suitable
transform (as is often the case in manifold-learning). Rather,
\(\vartheta( \cdot )\) abuses the geometry ensured by an oblique split
to rid the influence of~\({\overline{0}}_{i}\) elements from
\(\mathbf{X}_{i}\) that were artificially introduced by the oblique
validity function
\(\mathbf{X}_{i} \approx \mathbf{X}_{In,i}\odot\mathbf{\varnothing}_{In,i}\) (see SM-9),
mitigating the subsequent SVD alignment issues. Consequently, any
naturally occurring discontinuities in the data would remain, being
resolved by allocated models (i.e., the designation of correct split
position), not \(\vartheta( \cdot )\).~

Practically, \(\vartheta( \mathbf{X}_{i} )\) represents an
approach which iteratively interpolates each row of \(\mathbf{X}_{i}\)
to match the length of the temporal dimension
\textquotesingle{}\(q\)\textquotesingle. To achieve this, the column
indexes in each row of the subset elements
\({\overline{\mathbf{X}}}_{i}\) w.r.t. \(\mathbf{X}_{i}\) must be known.
As described in Section~\ref{subsubsec:local-region-interpretation}, these are determined by the application
of \(\eta\) to the local validity function \(\varnothing_{In,i}\), and
thus available. SM-11 provides an example algorithm for
\(\vartheta( \cdot )\) utilizing linear interpolation \cite{Lepot2017} for
speed and simplicity. Further discussion on the choice of interpolation
method can be found in \cite{BokorThesisfSRD}, but the authors expect this to be
an area for future optimization. Figure-6 demonstrates the use of
\(\vartheta( \mathbf{X}_{i} )\) with the given algorithm via
the resultant TSVD spectra on an oblique region \(\mathbf{X}_{i}\). The
alignment issue appears to be circumvented, enabling SVD to provide a
low dimensional embedding for eigen decomposition.

\subsubsection{\texorpdfstring{Inverse Transform and Global Mapping
}{Inverse Transform and Global Mapping }}\label{subsubsec:inverse-transform-and-global-mapping}

With the exact row co-ordinates of
\(\vartheta( \cdot ):\mathcal{M}_{x} \rightarrow \mathcal{M}_{y}\)
defined upon creation, constructing an inverse interpolation
\(\vartheta^{- 1}( \cdot )\) to revert any resultant model
\(\vartheta( \mathbf{X}_{i} ) \approx \mathbf{\Phi}_{i}diag( \mathbf{b}_{i} )\mathbf{T}^{t}( \boldsymbol{\omega}_{i} )\)
to the original per row column indices of
\(\ {\overline{\mathbf{X}}}_{i}\) s.t.
\(\vartheta^{- 1}:\mathcal{M}_{y} \rightarrow \mathcal{M}_{x}\) for
global representation via
\({\widetilde{f}}_{K,i}( \mathbf{X}_{i} ) = \ {\widetilde{f}}_{K,i}( {\vartheta_{i}^{- 1}\{\mathbf{\Phi}}_{i}diag( \mathbf{b}_{i} )\mathbf{T}( \boldsymbol{\omega}_{i} )\} )\)
is readily achievable. For each row \(j\) of the
\(\vartheta( \mathbf{X}_{i} )\) model, the goal is to retain
the predicted signal form, sampling a reduced set of points to compress
back to the original length of the row \(l_{j}\), i.e.,
compression/down-sampling \cite{AlanVOppenheim1998}. This is again interpolating but via
a courser grid, so many approaches can be taken. In this study linear
interpolation is again used for simplicity, but future work may explore
other options. An example algorithm to assemble
\(\vartheta_{i}^{- 1}( \cdot )\) is provided in SM-12.

\begin{figure}[!ht]
\centering
\includegraphics[width=\linewidth]{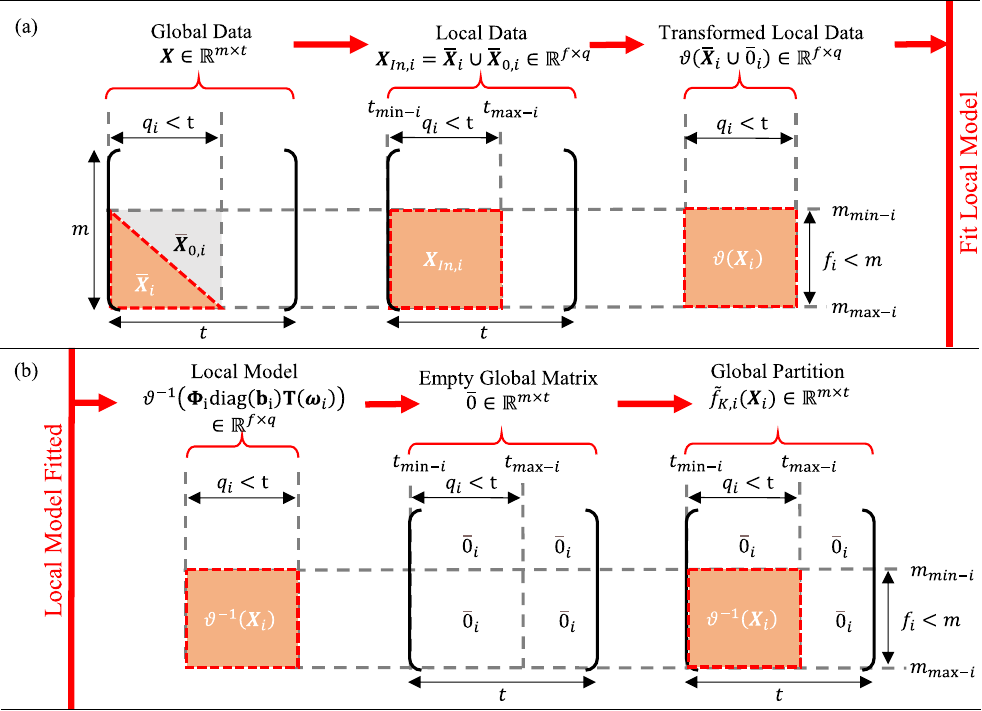}
\caption{: Full procedure for (a) assigning local co-ordinates and
applying a topological transform \(\vartheta( \cdot )\) to a candidate
local region
\(\mathbf{X}_{i} = \overline{\mathbf{X}}_{i} \cup {\overline{0}}_{i}\),
and (b) mapping the resultant local model \(\vartheta^{- 1}( \cdot )\)
back to global co-ordinates via \({\widetilde{f}}_{K,i}( \cdot )\)}
\end{figure}

From there the local model
\(\vartheta^{- 1}( \mathbf{\Phi}_{i}diag( \mathbf{b}_{i} )\mathbf{T}( \boldsymbol{\omega}_{i} ) ) \in \mathbb{R}^{f \times q}\)
must be mapped back to the global data co-ordinates
\(\mathbf{X} \in \mathbb{R}^{m \times t}\) in order to be aggregated via
the global model \eqref{eq:fSRD}. This is represented by the operator
\({\widetilde{f}}_{K,i}:\mathbb{R}^{f \times q} \rightarrow \mathbb{R}^{m \times t}\),
but simply entails placing each sub-matrix model \(\mathbf{X}_{i}\) in a
matrix of zeros matching the dimensions of
\(\mathbf{X} \in \mathbb{R}^{m \times t}\)~ via the same column index
extremities of \(\mathbf{\varnothing}_{i}(\mathbf{U})\) described in
Section~\ref{subsubsec:local-region-interpretation}, namely \(m_{min - i}\), \(m_{max - i\ }\),
\(t_{min - i}\) and \(t_{max - i}\). For a given
\({\overline{\mathbf{X}}}_{i}\), the full process from global data to
local transform, local model fit, and back to global mapping is provided
visually in figure-7. With this, fSRD's rule consequents are fully
specified, enabling a naïve implementation of fSRD's full invariant
decomposition for a given \(\mathbf{X}\). Next we will cover additions
for robustness.

\subsection{\texorpdfstring{Ergodic Partitioning of Continuous spectra
}{Ergodic Partitioning of Continuous spectra }}\label{subsec:ergodic-partitioning-of-continuous-spectra}

While \(\vartheta(\mathbf{X}_{i})\) solves the alignment issue, a high dimensionality may still occur if the new space
contains continuous spectra inherited from \(\mathbf{X}_{i}\)'s system characteristics. SM-13 
demonstrates this concept. In some systems, partitioning a continuous spectra (assuming correct split placement)
as locally restricted projections (i.e., ergodic partitions) with finite spectra is sufficient in off itself to 
approximate a parsimonious model both locally and globally. In fSRD, this corresponds to identifying invariant 
subspaces of the observable space \(G(\mathcal{M})\) that correspond to restricted integrals \([a,b]\subset[-\pi,\pi]\) within \eqref{eq:K_to_restricted_integral}
that sufficiently characterize the local dynamics while also producing a reduced local dimensionality. 
In other systems however, low dimensional localized ergodic partitions of the continuous spectra may be difficult
to define due to the limited dimensionality of the input space provided regardless of split position \cite{Mezic2017,Mezi2005}. 
In this case, fSRD as currently defined would be left only with further asymptotic decomposition to reduce 
local complexity, increasing global complexity as a trade-off to keep arbitrarily increasing accuracy.

This issue demonstrates the limits of using the spatial observable \(g(\mathbf{x}_{t})=\mathbf{x}_{t}\) to 
characterize certain nonlinearities, as discussed in Section~\ref{subsec:extended-dynamic-mode-decomposition-edmd}.
One solution is to similarly seek data-driven observables/transforms that collapse the continuous spectra
via alternative system co-ordinates. As was covered however, seeking universally generalisable observables
has its own issues. Alternatively, one can frame the presence of a continuous spectra as an ill-posed problem.
Consider again the exact DMD outlined in SM-1. This can be percieved as a two stage process. First a feature 
extraction of \(\mathbf{A}\) is implemented to solve \eqref{eq:DMD_cost} via a truncated/dimension-reducing
decomposition \(\widetilde{\mathbf{A}}\). Second, a projection back to the original dimensionality by 
reconstruction from the created spectrum \(\sigma(\widetilde{\mathbf{A}})\). This implies that any inaccuracies 
in the formation of \(\widetilde{\mathbf{A}}\) from the pseudo-inverse \(\mathbf{X}_{1}^{+}\) will affect \(\sigma(\widetilde{\mathbf{A}})\),
and thus the final model \(\mathbf{X} = f(\sigma(\widetilde{\mathbf{A}}))\).

Continuous spectra does not directly cause \(\mathbf{X}_{1}^{+}\)to be ill-posed, but its presence tends
to produce slowly decaying singular values within \(\mathbf{X}_{1}\), causing \(\mathbf{X}_{1}^{+}\) to become numerically 
ill-conditioned. Regulurization is a known tool for improving conditioning \cite{Tikhonov1977}, and can thus be used here at the decomposition stage to
minimize \eqref{eq:DMD_cost}, providing a best-fit minimal dimensionality estimate for the local-region 
data before fitting the spectral model (i.e., regularization to select a truncation point within SVD).
Regulizing each local partition would maximize the possible dimension reduction before considering 
further invariant decomposition, thus reducing the need for additional global partitions aswell (i.e., reduced global complexity).

To retain generalisable automatability, an automated method for regularization was implimented in fSRD. 
Assuming \((q-1)<<f\), consider for a given
\(\vartheta_{1}(\mathbf{X}_{i}) \in \mathbb{R}^{f \times (q-1)}\) that one seeks a mapping \(\mathbf{S} \in \mathbb{R}^{(q-1) \times (q-1)}\) for the linear basis 
\(\mathbf{Z} \in \mathbb{R}^{f \times (q-1)}\) that exists within a reduced dimensionality expectation plane
\(\varepsilon_{r}\) while capturing the maximum information content possible from the original system. The SVD of
\(\vartheta_{1}(\mathbf{X}_{i})\) provides an initial forward mapping \(G(\mathcal{M}) \rightarrow \varepsilon_{r}\):

\begin{equation}
\vartheta_{1}(\mathbf{X}_{i}) = \mathbf{Z}\mathbf{S} + \mathbf{E}
= \mathbf{U}\mathbf{\Sigma}\mathbf{V}^{\mathsf{H}} + \mathbf{E}
\text{, \ \ s.t., \ \ }
\mathbf{Z} = \mathbf{U}\mathbf{\Sigma}
\text{, \ \ and, \ \ }
\mathbf{S} \approx \mathbf{V}^{\mathsf{H}}
\label{eq:svd-linear-projection}
\end{equation}

where \(\mathbf{E} \in \mathbb{R}^{f \times (q-1)}\) is an explicit error term whose columns are 
independent and identically distributed (IID). The product \(\mathbf{U}\mathbf{\Sigma}\) provides a linear
orthonormal basis, and \(\mathbf{V}^{\mathsf{H}}\) an estimate for the inverse mapping \( \varepsilon_{r} \rightarrow G(\mathcal{M})\).
Once any singular values \(\approx 0\) are removed for rank deficiency, the mapping is then within \(r_{1}\) s.t.,
\(r_{1} \leq (q-1) \), providing a candidate orthonormal projection. The corresponding candidate inverse mapping, 
\(\mathbf{V}_{H}\), will at this stage still result in ill-conditioning, but can be optimized via 
ridge regulization, providing a data driven identification of unrequired dimensions through there reduced contribution (i.e., shrinkage) \cite{Tikhonov1977}.
More precisely, one can use a ridge formulation of \(\mathbf{S}\) to estimate an inverse mapping that provides
the smallest dimensionality within \(\varepsilon_{r_{1}}\) paramatized by \(\delta\):

\begin{equation}
\mathbf{V}^{\mathsf{H}} \approx
\mathbf{S} \approx
(\mathbf{Z}\mathbf{Z}^{\top} + \delta\mathbf{I})^{-1}\mathbf{Z}^{\top}\mathbf{X}_{1}
=(\mathbf{\Sigma}^{\top}\mathbf{\Sigma} + \delta\mathbf{I})^{-1}\mathbf{\Sigma}^{\top}\mathbf{U}^{\top}{X}_{1}
\in 
\mathbb{R}^{r_{1} \times (q-1)}
\label{eq:svd-ridge-reg}
\end{equation}

\(\delta\) may then be derived from data for each unique system to maximise dimensionality reduction while 
retaining system information. \(\mathbf{S}\) via \eqref{eq:svd-ridge-reg} will now contain several rows that contribute 
negligibly to the overall output, indicating truncatable dimensions from \(r_{1} \rightarrow r_{2}\) to reach
a data-driven \(\varepsilon_{r_{2}}\).

This forms the basis for a novel 
hueristic approuch for reguluriing SVD, aligned conceptually with elastic-net regularization \cite{Goutte1997}.
First, one truncates any singular values \(\approx 0\) to handle rank deficiency, reducing dimensionality from \((q-1) \rightarrow r_{1}\). Implementing
ridge next allocates weighting to the singular modes, maintaining dimensionality \(r_{1}\). A final truncation of 
any singular values \(\approx 0\) then removes dimensions with negligible contribution after weighting. This reduces dimensionality 
from \(r_{1} \rightarrow r_{2} \text{, s.t., } r_{2} \leq r_{1} \leq (q-1)\). In fSRD this is implemented via the algorithm outlined in SM-14 for 
each candidate local region. To impliment this, an automated method to identify unique regularization parameter’s \(\delta\) for each local region 
is required. A data-driven fixed point iteration scheme adapted from \cite{CARY2026} is used here. The algorithm for this scheme is also provided in SM-14
during stage-5. This notably includes the use of the following BIC derived multivariate fixed point iteration eqaution, \(h(\delta)\), and its analytical derivative,
\(dh(\delta)/d\delta\):

\begin{equation}
  \delta_{BIC} 
  =
  h(\delta) 
  =
  \frac{\log(f(q - 1))tr\left( \mathbf{B}^{- 1} - \delta\mathbf{B}^{- 2} \right)}{2f^{2}(q - 1)}
  \left( \frac{\sum_{j}^{q - 1}{{\widetilde{\mathbf{x}}}_{i,j}^{\top}\left( \mathbf{I} - \mathbf{S}(\delta) \right)^{2}{\widetilde{\mathbf{x}}}_{i,j}}}
  {\sum_{j}^{q - 1}{{\widetilde{\mathbf{x}}}_{i,j}^{\top}\mathbf{Z}\mathbf{B}^{- 3}\mathbf{Z}^{\top}{\widetilde{\mathbf{x}}}_{i,j}}} \right)
\label{eq:fixed-point-iteration-h}
\end{equation}

where the power superscript used for symmetric matrices refers to the multiplication of several when the
indication of their transpose is redundant, i.e., \(B^{\top} = B, s.t, B^{3} = BBB = B^{\top}B^{\top}B^{\top}\). 
\eqref{eq:fixed-point-iteration-h} is a novel development of the multivariate formulation found within \cite{CARYMultivariate} for this works matrix based application in 
SM-14, with the original univariate derivation found within \cite{CARY2026}. SM-15 contains the full derivation for \eqref{eq:fixed-point-iteration-h} and 
\(dh(\delta)/d\delta\) as used in this study, outlining all developments from the original method \cite{CARYMultivariate,CARY2026}.
SM-16 demonstrates results of the full regularization methodology via the same three spectra examples used in SM-13.

\subsection{Error Model}\label{subsec:error-model}

To demonstrate fSRD functionality, for all examples in this study 
error is assumed to be IID and have a normal distribution s.t. each
column of \( \mathbf{E} \) is \( e_{j} = \mathcal{N}(0,\sigma^{2}W)\), where \(j=1,\cdots,q-1\).
\(\mathbf{E}\) is an explicit error term defined at the local system level for each region.
An explicit error model is defind here for each \(\mathbf{E}\) via the SVDe method \cite{Epps1st2019,Epps2nd2019}, 
filtering local noise via a whitening process to eliminate \(\mathbf{E}\) before respective 
SVD, truncation, regularization, and spectral decomposition. This order is chosen to mitigate the influence of noise
on split position optimization, which could obscure the location of real invariant interfaces.
While not the focus of the study, this approach readily allows the incorporation of 
alternative error modelling methodologies to handle different types of noise such as
heteroscedastic or serially correlated error. Exploring these elements further is left to
future work.

\subsection{Summary}\label{subsec:summary}

The total operation of fSRD is summarised via the pseudocode provided in
SM-18, SM-20, and SM-21, representing the forward pass, local model fit,
and backwards elimination respectively. Standard operation involves a
forward pass training to construct the candidate tree architecture,
followed by a backwards elimination pass to remove redundancy, both via a BIC difference measure.
Flow charts for each are thus provided in SM-17 and SM-19 respectively.
Local model selection (i.e., split position) is dictated by WNRMSE, \eqref{eq:WNRMSE}. Hyper
parameter values and additional implementation specifics are covered in
SM-22 for ease of reproducibility, but further discussion for these choices 
can be found in \cite{BokorThesisfSRD}.

In terms of applications, in this paper we demonstrate that fSRD has a very broad scope. By
systematically automating the Koopman framework with multi-operator
spectral decomposition (i.e., invariant decomposition), finite
dimensional Koopman representation is achieved, targeting the proposed
challenges identified in Section~\ref{subsec:problem-statement} by prioritizing interpretable
invariant operators, but allowing ever increasing accuracy via subsequent
restricted integrals. This suggests that fSRD possesses universal
approximator capability in the representational sense for sequential
processes represented as finite data matrices. Rather than approximating
static input--output mappings as in NN's, the method constructs
finite-dimensional representations of the evolution of ordered data
through one or more linear operators acting on a sufficient observable
space induced by \(\mathbf{X}\). Importantly, unlike Koopman methods
assuming globally valid observables, this does not require the
underlying system to admit a single global linear Koopman operator.
Nonlinear, nonstationary, or chaotic dynamics may thus be represented
through compositions or collections of operators that are locally valid
in \(g(x)\), with their respective fuzzy boundaries enabling non-ideal
boundary conditions generated from either the system itself or data
quality. Sufficiency w.r.t the observable thus refers only to sufficient
localised non-trivial structure to exploit an estimate. Universality is
therefore applicable with respect to the operator-valued representation
of the induced dynamics on observables, not the original nonlinear
state-transition map.

As with NN universality results \cite{Hornik1991,Cybenko1989}, this expressivity
does not imply guarantees of identifiability or generalisation from
finite data. Provided that at least partial structure exists within the
data however, the resulting surrogate yields a constrained
representation that propagates sequential dependencies via linear
operators. Relative to unconstrained nonlinear models like NN's, this
may enhance identifiability and generalisation by restricting admissible
solutions to interpretable linearly evolving representations. This
constraint substantially reduces model ambiguity, promoting local
identifiability (inductive bias towards locally valid invariant
regions), while global identifiability remains dependant on data
coverage (i.e., non-unique tree construction). Via
local spectral models, fSRD yields both a prediction and an 
interpretable spectral decomposition of a given dataset.

With this, and by using generic vectorized inputs, fSRD's `structured'
data requirement is only that a sequential dependency between columns
(i.e., dynamical structure) exists in some sub regions of the data,
otherwise permitting flexibility in row representation and assembly. It
can thus be applied to a wealth of data formats and applications where a
tensor representation can be formed. Applicable major data formats
include but are not limited to; univariate and multivariate time series,
spatial (e.g., images), spatiotemporal (e.g., videos), functional
mappings (e.g., response surfaces), multimodal (e.g., sensor fusion),
and graphs/structured data (e.g., networks). A proof of this universal approximator 
capability was beyond the scope of this work, and we leave it to future 
work to explore this hypothesis.

%% file: sections/results_duffing.tex
\section{Results}\label{sec:results}

The potential generality described has broad implications. In this way,
fSRD could offer a pathway to unifying nonlinear sequential systems
under a tractable, interpretable, and computationally efficient
framework. Within the limited scope of this initial study, focus will be
placed on demonstrating fSRD's core principles and behaviour. While
elements of fSRD's generality will be explored, the full extents of this
will be left to future work.

Three case studies are presented that demonstrate the application of
fSRD to the analysis of nonlinear dynamic systems. Given fSRD's highly
generalised structure, interpretability will have differing meaning
between case studies, depending on the application, the chosen evolution
variable, and the structure/assembly of the input \(\mathbf{X}\). The use of a
tree architecture and Koopman spectral decompositions in fSRD provides
several advantages for algorithm interpretability that are well
documented in other ML algorithms \cite{Babuka2003,LiX2021,Hartmann2012,Gohel2021}
and individual Koopman operators \cite{Brunton2021,Mezi2005,Tu2014,Williams2015}
respectively. Two elements of fSRD exemplify this:
the placement of the linear invariant regions, and the modal breakdown
of their Koopman spectra in each region. Both will be brought up
routinely to assess the following results. Note that all datasets presented in this study 
are re-scaled to the range {[}0,1{]} to mitigate numerical conditioning issues \cite{Shahzad2020}.

\subsection{Duffing System}\label{subsec:duffing-system}

The Duffing oscillator \cite{GeorgeDuffing1918} has three equilibria, a saddle point at
\(\lbrack 0,0\rbrack\), and two fixed points at \(\lbrack 1,\ 0\rbrack\)
or \(\lbrack - 1,0\rbrack\). The dynamics are governed by the equations
\cite{Salova2019}:

\begin{equation}
{\dot{x}}_{1} = x_{2},\ \ {\dot{x}}_{2} = x_{1} - x_{1}^{3}
\label{eq:Duffing}
\end{equation}

Such systems can be interpreted experimentally in several ways in which
fSRD can readily adapt (e.g., duffing as a quasi-potential response
surface). This is demonstrated in \cite{BokorThesisfSRD}, but here sampled
trajectories will be assembled as time series rows. Similarly, these
trajectories can be assembled in several ways (e,g., separate matrices
for each co-ordinate), but for demonstration the temporal signals will
be weaved into a single matrix via:

\begin{equation}
\mathbf{X} = \begin{bmatrix}
\begin{matrix}
| \\
\mathbf{x}_{1,1} \\
|
\end{matrix} & \begin{matrix}
| \\
\mathbf{x}_{2,1} \\
|
\end{matrix} & \begin{matrix}
\cdots & \begin{matrix}
| \\
\mathbf{x}_{1,N} \\
|
\end{matrix} & \begin{matrix}
| \\
\mathbf{x}_{2,N} \\
|
\end{matrix}
\end{matrix}
\end{bmatrix}^{\top}
\label{eq:Duffing_matrix}
\end{equation}

where \(i = 1,\ldots,N\) is the number of trajectories recorded, and
\(\mathbf{x}_{1,i}\) \(\mathbf{x}_{2,i}\) are vectors of the axis values as trajectory
\(i\) evolves over time (for the respective axis in the phase portrait).
In \(g(x) = x\), the Duffing systems
global Koopman operator has no true eigenvalues except for
\(\lambda = 1\) with eigenspace corresponding to its conserved
Hamiltonian energy and positive invariant sets \cite{Colbrook2023}. This is because
multiple fixed points and/or attractors cannot be topologically
conjugate to a finite single fixed point linear representation as per
the previously discussed Hartman Grobman theory \cite{Hartman2002,Hartman1960,Grobman1959},
acting as a hard structural limit to invariance.

\subsubsection{Time Series - DMD}\label{subsubsec:time-series---dmd}

As demonstrated in \cite{Colbrook2023}, EDMD with an observable dictionary of
radial basis functions applied to random samples of the Duffing
oscillator (\(\mathrm{\Delta}t = 0.25)\) yields a Koopman spectral
approximation containing many spurious (non-physical) eigenvalues. This
spectral structure can be readily improved, but as shown in
\cite{Colbrook2023} only by exploiting system-specific dynamical
information (namely known invariant sets and basins of attraction)
rather than through a systematic, system-agnostic procedure. This is
typical, with the EDMD estimate not yielding a globally closed
finite-dimensional invariant subspace of observables, systematically or
otherwise.

\begin{figure}[!ht]
\centering
\includegraphics[
    width=14cm,
    height=7cm
]{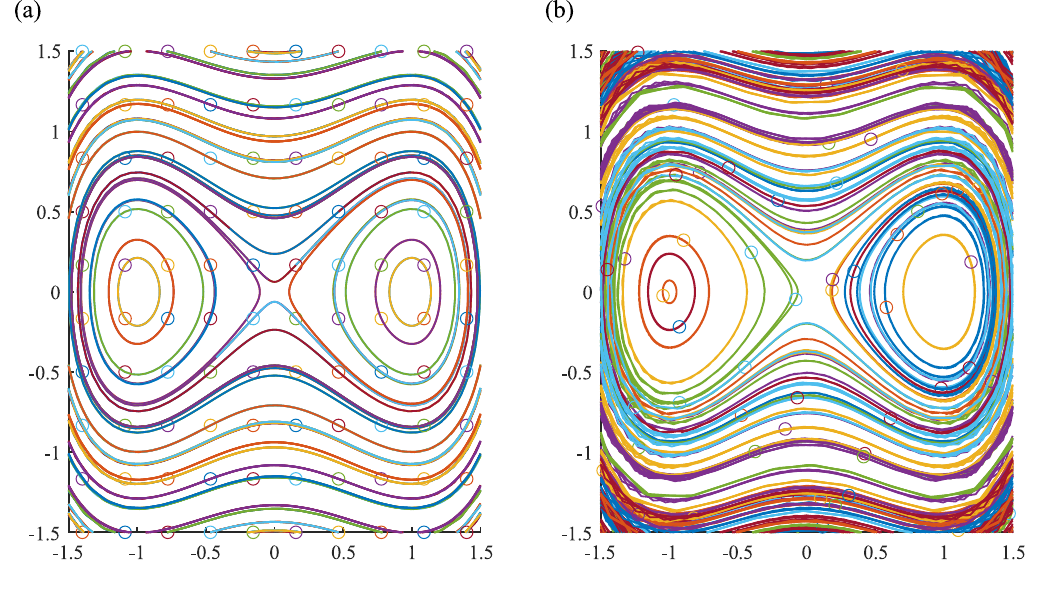}
\caption{: (a) Duffing attractor trajectories with equally spaced
sampling. (b) Duffing attractor trajectories with uniform random
sampling. \(X = \mathbb{R}^{200 \times 100}\)}
\end{figure}

Consider the illustrative Duffing sets assembled via \eqref{eq:Duffing} depicted in
figure-8. To enforce systematic application and no a-priori system
information, exact-DMD with SVDe is implemented with no observables
(i.e., \(g(x) = x\)), utilizing a fixed truncation of \(r = 1e - 4\) to
handle rank deficiency \cite{Turco2017}. While all experimental parameters are
otherwise identical (e.g., same number of samples), figure-8-(a) uses
equally spaced symmetric sampling across the designated parameter ranges
and figure-8-(b) uses a uniform pseudo-random sampling (Monte
Carlo). This comparison is made not from the perspective of optimizing
an experimental design, but to illustrate even simple exploratory
sampling strategies can produce misleading outcomes and mask a lack of
model robustness. Figure-8-(a) for example characterises each of the
main attractors in the Duffing set, assigning a proportional density of
the available trajectories. This produces several level orbitals around
each fixed point before merging beyond the saddle point into the outer
orbitals. Experimentally, this is the ideal, but almost complete prior
system knowledge is used to achieve this even without the use of
observables.

\begin{figure}[!ht]
\centering
\includegraphics[
    width=\linewidth,
    trim={0 0.9cm 0 0},
    clip
]{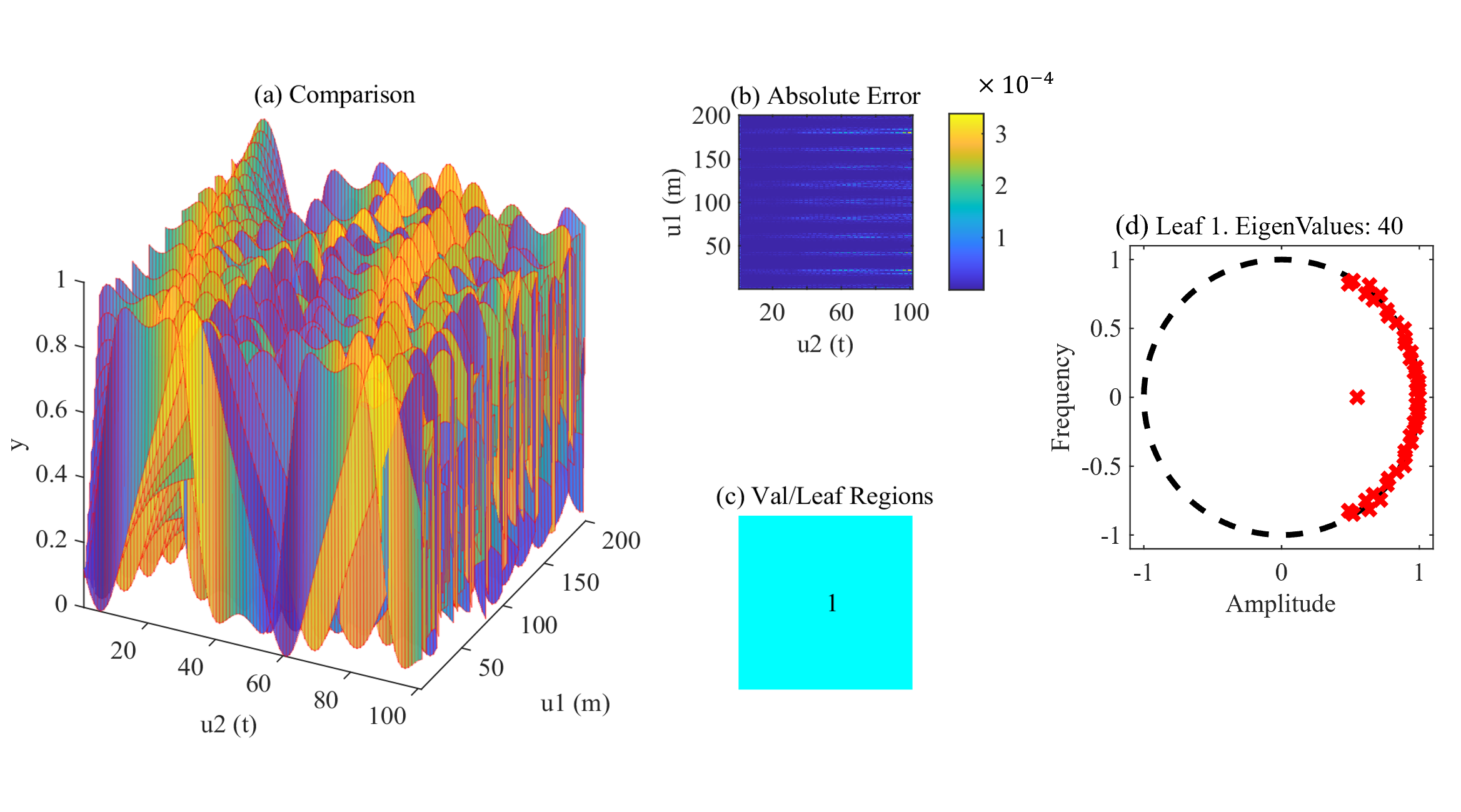}
\caption{: DMD with \(r = 1e - 4\) truncation fitted Duffing equally spaced sampling data}
\end{figure}

\begin{itemize}
\item
  Trajectory lengths (columns of X) are intentionally chosen to limit
  their individual chance of crossing multiple attractors, confining
  ideally to level sets of eigenfunctions that are invariant in time
  (e.g., level contours/orbits in phase space containing symmetries).
  This is evident in figure-9-(d), where the spectra is filled with
  Koopman eigenfunctions (i.e., eigenvalues on the unit circle)
  \cite{Brunton2021}.
\item
  Sample initial points are equally spaced across pre-determined ideal
  parameter ranges, both to confine level orbits and to include
  mathematically precise symmetries present in the Duffing system.
\item
  The number of trajectories (spatial rows of X) is chosen to ensure
  sufficient density across each attractor for the chosen trajectory
  lengths and sampling method/distribution.
\item
  No noise is present.
\end{itemize}

These correspond to several factors in experimental design for nonlinear
or chaotic systems that influence the performance of resulting models.
The three relevant here include:

\begin{itemize}
\item
  Sampling symmetry, regardless of sampling method, can systematically
  bias results if the sampling pattern aligns with inherent system
  structure (e.g., resonance, periodicity, orbital symmetry) \cite{Niederreiter1992}.
\item
  Sampling coverage, including both number of samples (i.e., rows of
  \(X\)) and positioning across the parameter space (sampling
  design/strategy), dictates the experiments characterization of the
  given system. In high sensitivity systems such as chaotic dynamics,
  sparse or uneven coverage may leave critical regions, such as specific
  orbitals, unsampled \cite{Ott2002,DouglasCMontgomery2019}.
\item
  Increasing trajectory lengths (i.e., columns of \(X\)) inherently
  increases the predictive horizon required by a resultant model, often
  leading to reduced accuracy \cite{Ott2002,Sprott2003},
\end{itemize}

All these aspects can in principle be improved for a given task, though
often requiring iterative incorporation of prior system knowledge
\cite{Niederreiter1992,Ott2002,DouglasCMontgomery2019}. Figure-9 demonstrates this as to suit
DMD's strengths (avoiding systematic eigenvalue bias due to noise,
mitigating multi-attractor effects, reducing matrix condition to enhance
amplitude estimates, etc.) \cite{Bagheri2014,Dawson2016,Jovanovi2014,Tu2014}, 
resulting in a high predictive accuracy (\(NRMSE\sim 6.58e-05\)).

\begin{figure}[!ht]
\centering
\includegraphics[
    width=\linewidth,
    trim={0 1cm 0 0},
    clip
]{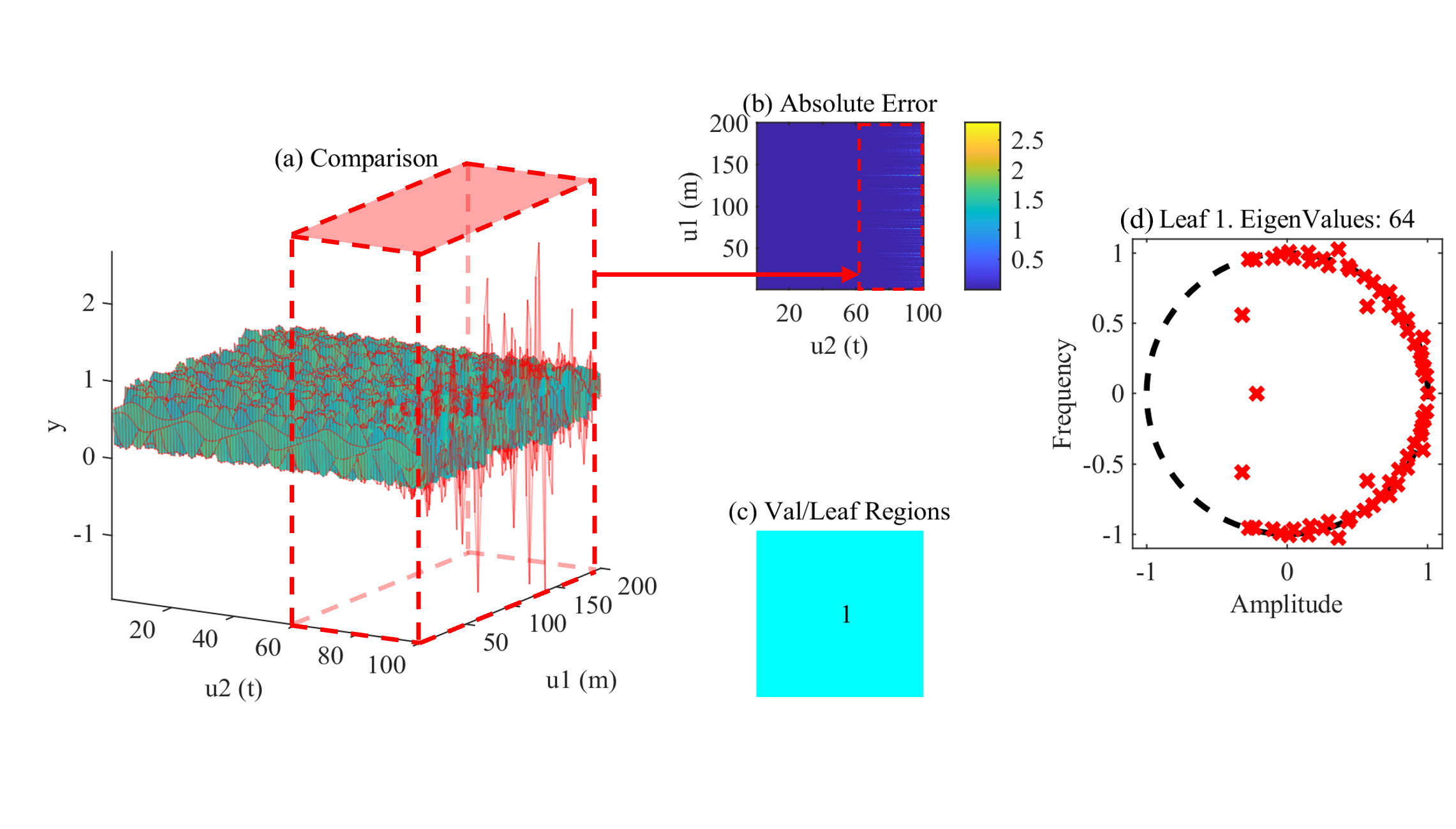}
\caption{: DMD with \(r = 1e - 4\) truncation on Duffing random uniform sampling data}
\end{figure}

By contrast, consider the Monte Carlo sampling in figure-8-(b), where
the listed factors that eased DMD's performance are absent. The
corresponding DMD model is shown in figure-10. The trajectories are now
insufficient to densely characterise all attractors (e.g., insufficient
number of rows) and several trajectories increasingly violate limits to
invariance (e.g., trajectories cross multiple-attractors, non-symmetric
starting positions and paths, escaping level orbits). The latter point
is apparent in figure-10 (a) and (b), where the identified mostly unit
circle eigenfunctions inaccurately predict over half of the trajectories
(\(\sim110\)) past \(t > 60\), with overall predictive
performance dropping to \(NRMSE\sim 0.31\) (i.e., \(69\%\) accuracy).
This is despite the corresponding spectra containing vastly more
eigenvalues/parameters as in figure-10-(d) (\(40 \rightarrow 64\)). The
demonstrated reliance on systems priors' limits EDMD's and DMD's one
shot accuracy, thus limiting their broader use as a modelling
methodology in live and offline applications.

\subsubsection{Time series -- fSRD}\label{subsubsec:time-series-fsrd}

\begin{figure}[!ht]
\centering
\includegraphics[
    width=\linewidth,
    trim={0 0.3cm 0 0},
    clip
]{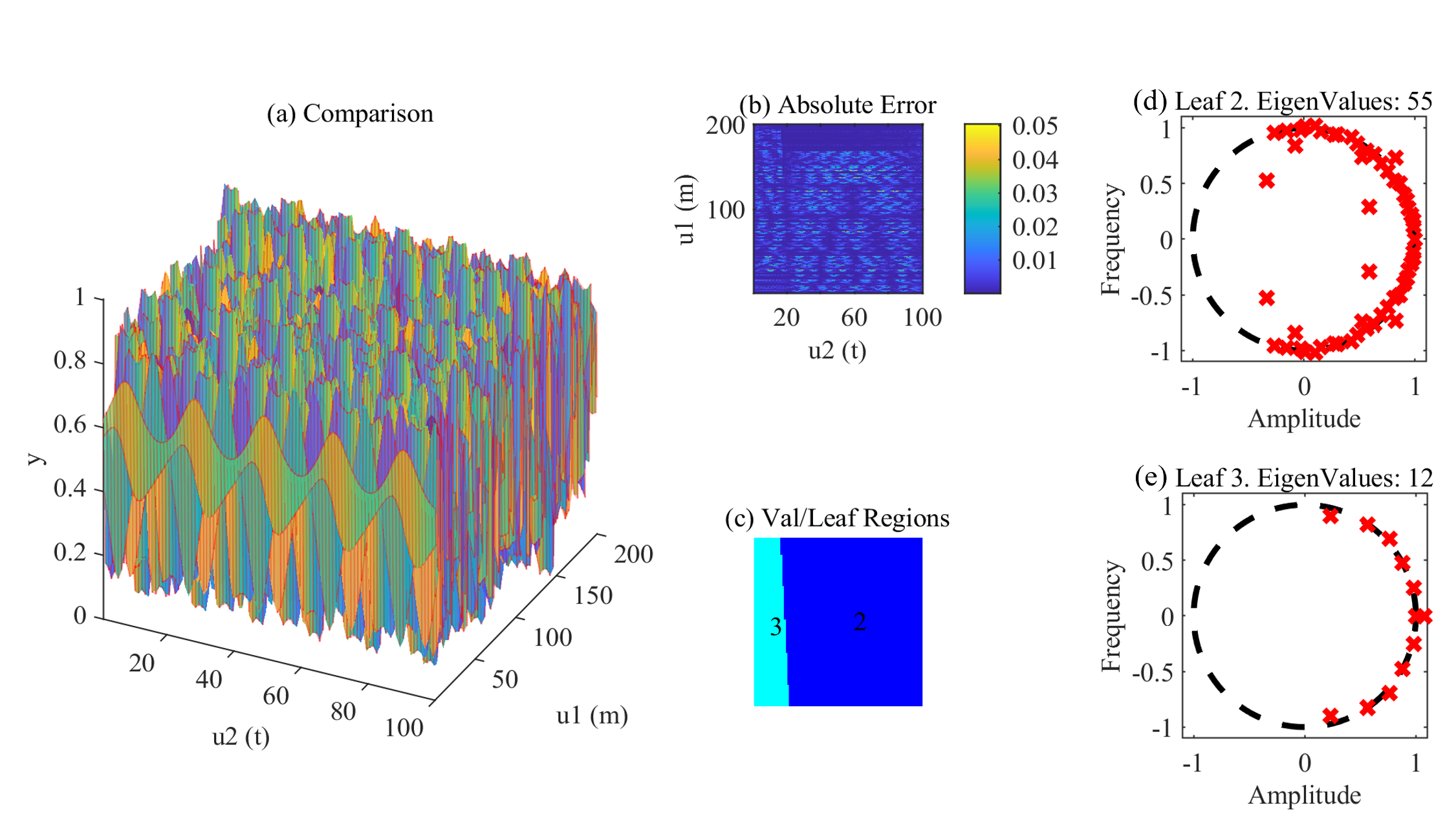}
\caption{: fSRD on Duffing random uniform sampling data. Global NRMSE = 0.026}
\end{figure}

Figure-11 illustrates the application of fSRD on the identical set of
Monte Carlo sampling (figure-10), improving performance with
\(NRMSE\sim 0.026\), i.e., \(97.4\%\) accuracy. fSRD removes the
predictive inaccuracy that occurred at \(t > 60\) via decomposition into
two invariant sets. System relevant reasoning for this can be readily
interpreted through the models' traits. First, with respect to region
placement, the split is perpendicular to the temporal axis \(u_{t}\),
suggesting that nonlinearities that break forward invariance occur
primarily as the system evolves with time. Second, consider these
results depicted in the original phase space co-ordinates as per
figure-12. Leaf-3 appears to isolate the non-symmetric starting
conditions of the trajectories in time, while leaf-2 formulates several
level orbitals that capture the remaining system, as per the mostly unit
circle eigenvalues in figure-11-(d). Together, this suggests that by
isolating non-ideal trajectory asymmetries, fSRD enables an ideal
invariant partition corresponding physically to a stable global limit
cycle. This demonstrates successful systematic representation of a
system both absent of global eigenfunctions and containing no global
topological conjugacy, all while extracting meaningful interpretability
without system priors.

\begin{figure}[!ht]
\centering
\includegraphics[
    width=14cm,
    height=10cm
]{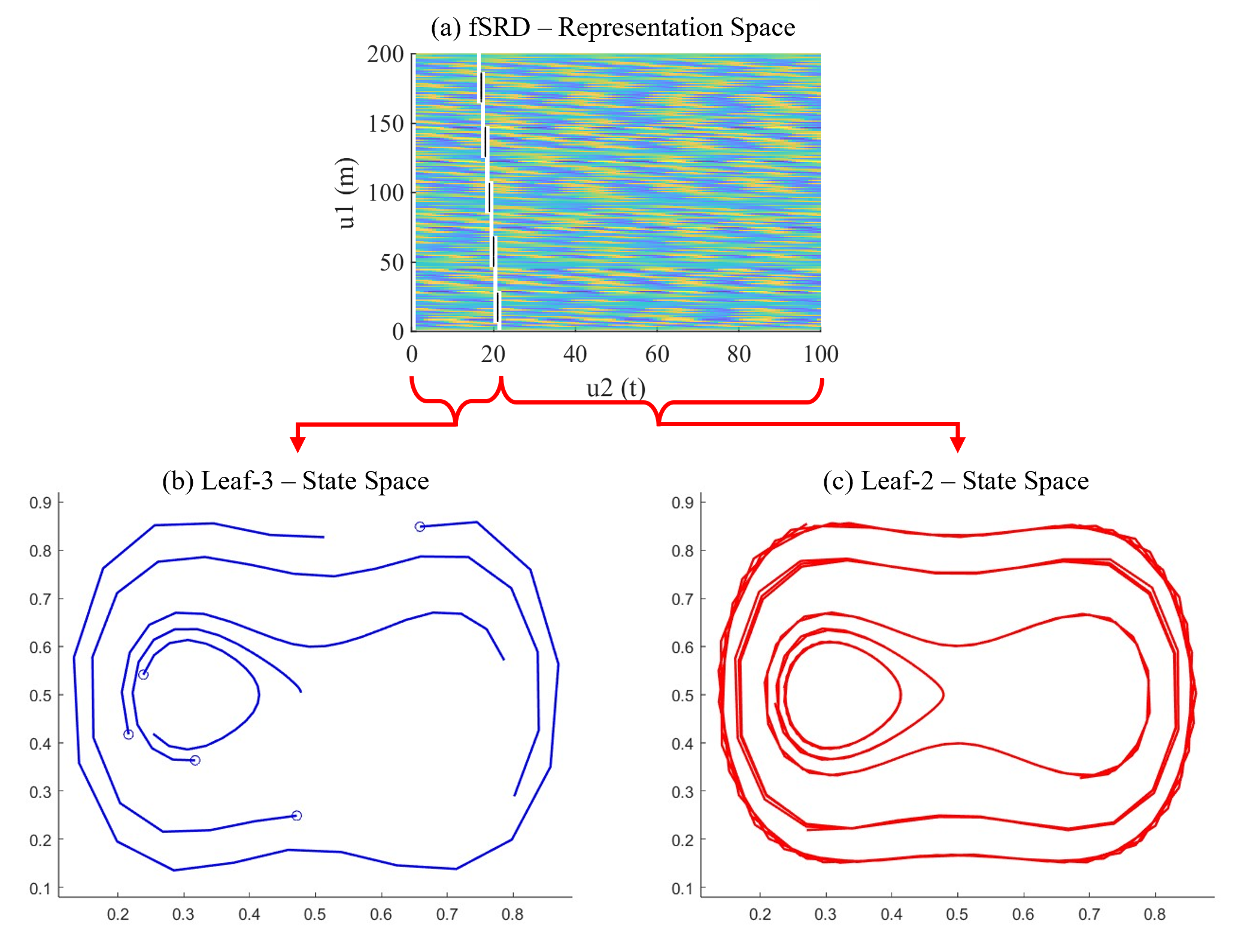}
\caption{: fSRD model in original state-space. Black lines = fuzzy interpolation.}
\end{figure}

This illustrative example demonstrates fSRD's interpretable adaptation
to full-orbit invariant partitions \('I'\) in the presence of partial
orbits \('pO'\), emphasising the unreliability of single operator
representation in non-ideal data sets. That said, a more robust
experimental design is required to explore overarching one shot model
robustness.

\subsubsection{Factorial Experiment}\label{subsubsec:factorial-experiment}

\begin{figure}[!ht]
\centering
\includegraphics[
    width=13cm,
    height=6cm
]{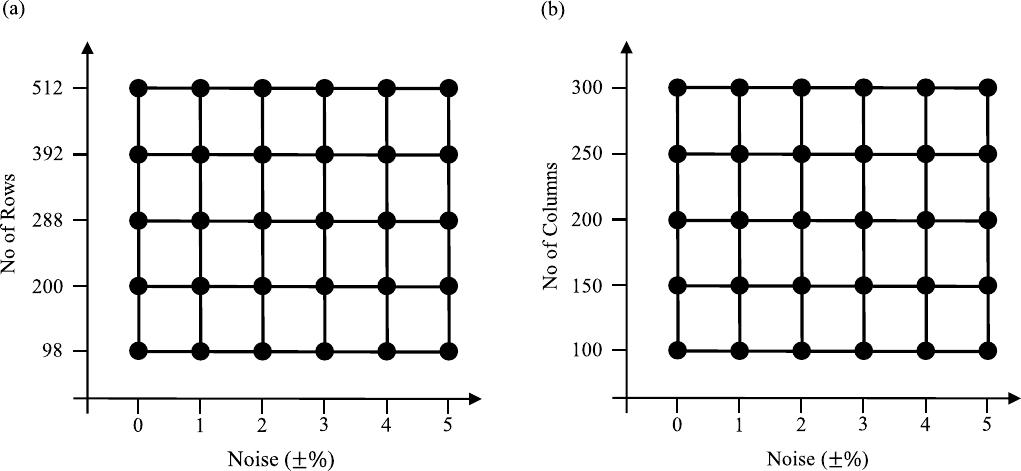}
\caption{: Bootstrap sampled \(2^{k}\) factorial DOE. (a) spatial design, (b) temporal design}
\end{figure}

Consider the two-factor factorial design of experiments (DOE) \cite{DouglasCMontgomery2019}
presented in figure-13. The Duffing system is sampled via Monte Carlo
with replacement (i.e., bootstrap sampled) 10 times for each of the
indicated factor combinations \cite{Davison1997}. Each design varies gaussian IID
noise of the form \(e_{j}\mathcal{= N}(0,\sigma^{2}W)\) between
\(0 \rightarrow \pm 5\%\), proceeding to vary the number of rows (number
of trajectories) and columns (trajectory lengths) in figure-13-(a) and
figure-13-(b) respectively. Each DOE has 50 samples per noise level with
350 samples total.

This noise's effect on the overarching system is demonstrated in SM-23.
At \(e_{j} = \pm 3\%\), broad attractor behaviour becomes partially
obscured as neighbouring trajectories occasionally overlap. At
\(e_{j} > \pm 3\%\), noise induced variation rises beyond the tolerance
between trajectories, i.e., average distance between trajectories on
stable orbits without noise,
\(\sqrt{x_{1}^{2} + x_{2}^{2}} \approx 0.05\); possible noise induced
trajectory variation at \(e_{j} = \pm 4\%\),
\(\sqrt{x_{1}^{2} + x_{2}^{2}} \approx 0.057\). This can fully obscure
or potentially change system/attractor behaviour, with trajectories both
overlapping each other and reverting behind their prior samples if the
variation becomes larger than the distance covered between time points.

SM-24 discusses DMD's performance across these factorial experiments.
Summarised, DMD is demonstrated to be broadly unreliable, being highly
sensitive to the data form and quality. Introducing a noise model
improves reliability, but emerging relationships with the number of rows
or columns limits useful performance. Consequently, as per the
illustrative example provided in Section~\ref{subsubsec:time-series---dmd}, while DMD can generate
practically useful models in some cases, its success applied to this
nonlinear/chaotic system heavily relies on the integration of prior
information during experimental design. fSRD's performance on these
factorial experiments is provided in figure-14, with both it and DMD's
exact values provided in SM-25 and SM-26. DMD with SVDe results from
SM-24 are transparently overlayed for easy comparison. fSRD
is implemented here with
\(\Theta = 3\) in BIC (see SM-8), discouraging arbitrary predictive accuracy by
prioritizing more parsimonious solutions (i.e., reducing additional
partitions that increase global BIC).

\begin{figure}[!ht]
\centering
\includegraphics[width=\linewidth]{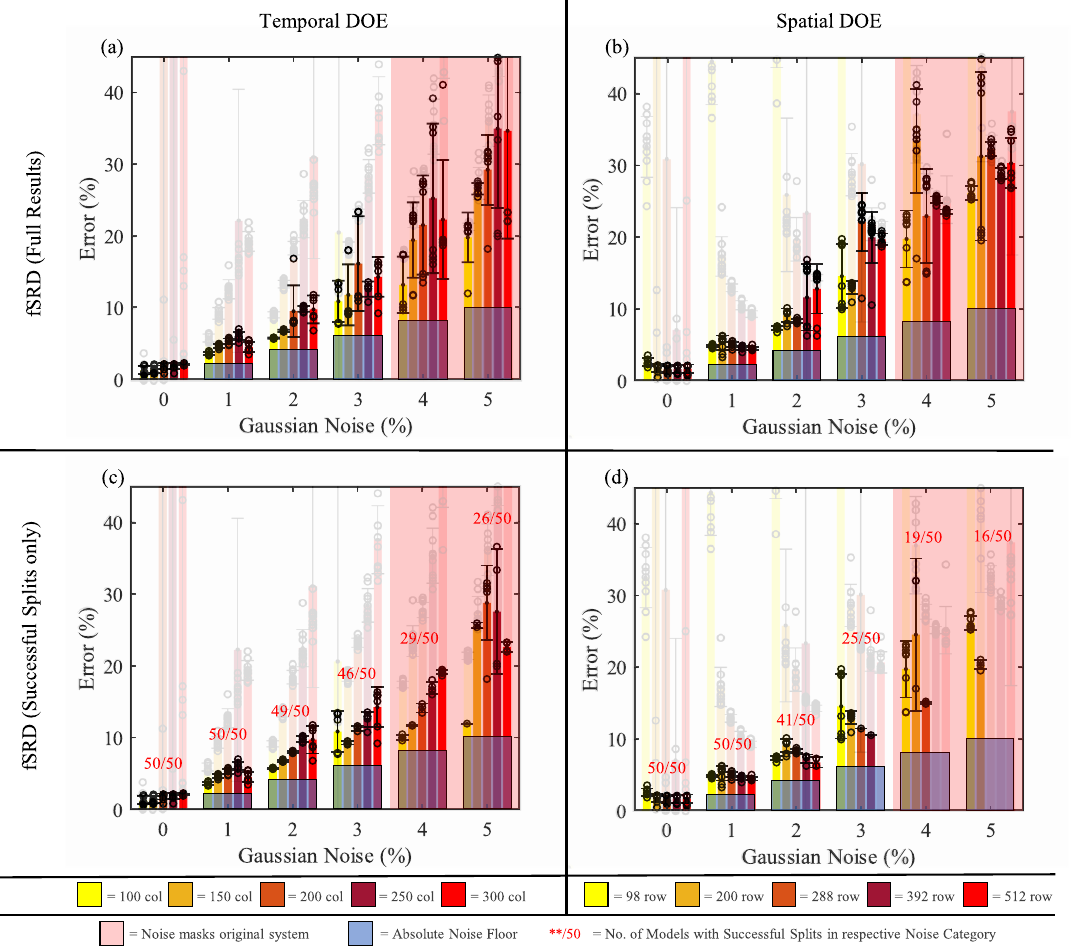}
\caption{: Factorial Experiment results (fSRD)}
\end{figure}

Several trends and comparisons are observed, detailed within SM-27.
Summarising, fSRD reliably characterises highly accurate, precise and
non-overfitting models of the Duffing system independent of noise,
trajectory position, experimental limitations, etc., so long as the
system's features/structure are not themselves obscured (i.e.,
insufficiently rich or low-quality data). In this case, where/when
dynamic obscuration occurs is partly dependant on the chosen noise
model, so gains may be achieved via a more bespoke selection.

%% file: sections/results_lorenz.tex
\subsection{Lorenz System}\label{subsec:lorenz-system}

\begin{figure}[!ht]
\centering
\includegraphics[width=\linewidth]{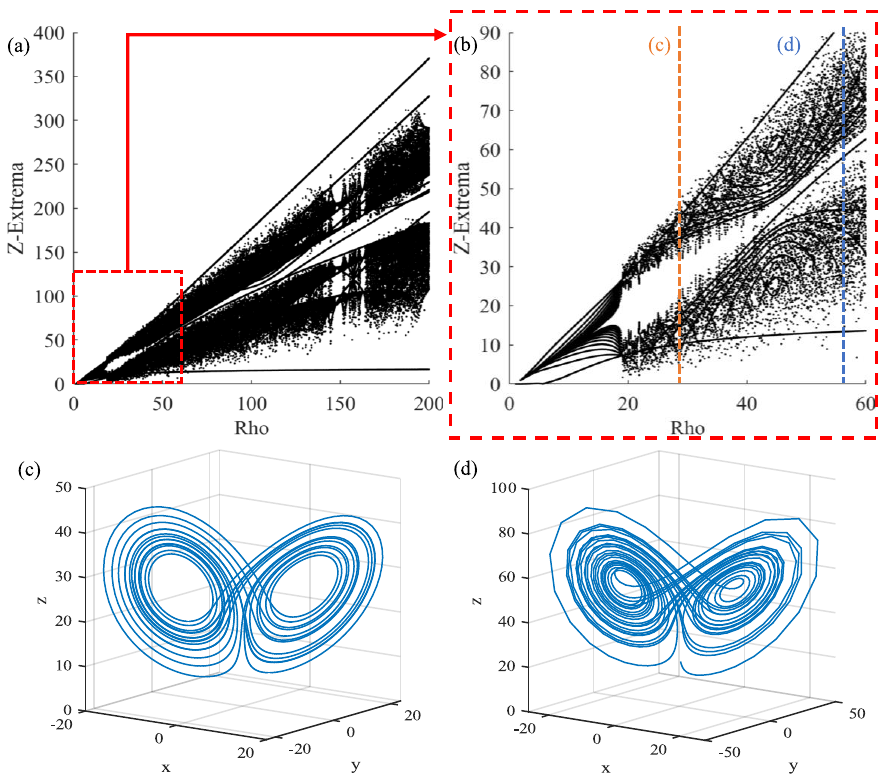}
\caption{: (a) Lorenz system bifurcation graph with \(\rho\), (b) zoom
in to region of interest, (c) 3d Lorenz ‘butterfly’ form, (d) altered ‘butterfly’ form}
\end{figure}

The Lorenz system \cite{EdwardNLorenz1963} is a dynamical system, characterized by three
coupled ordinary differential equations:

\begin{equation}
\frac{dx}{dt} = \sigma(y - x),\ \ 
\frac{dy}{dt} = x(\rho - z) - y,\ \ 
\frac{dz}{dt} = xy - \beta z
\label{eq:Lorenz}
\end{equation}

These give rise to a system that bifurcates along the variables
\(\sigma,\rho\ \&\ \beta\), generating strongly mixing and chaotic
trajectories for certain parameter values and initial conditions.
Figure-15 demonstrates this via bifurcation of the constant \(\rho\). A
popular three-dimensional form is shown in figure-15-(c) via
\(u_{m} = \lbrack x,y,z\rbrack\), \(\mathrm{\Delta}t = 0.001\), and the
constants \(\sigma = 10,\ \ \rho = 28,\ \ \beta = 8/3\). Two elements
are present however that ease its modelling complexity.

\begin{itemize}
\item
  The chosen \(\mathrm{\Delta}t\) is based on prior system knowledge to
  ensure a smooth/continuous function beyond the minimum required to
  capture its form (e.g., equivalent to increasing the number of rows in
  \(\mathbf{X}\)).
\item
  \(\sigma,\ \rho,\ \beta\) are chosen with prior knowledge to provide
  an almost complete orbital with several mathematically precise
  symmetries.
\end{itemize}

As shown in the factorial experiment, DMD performance is greatly
diminished without these biases. Consequently, both elements are altered
here to better represent real-world applications/ experiments, producing
figure-15-(d):

\begin{itemize}
\item
  \(\mathrm{\Delta}t = 0.02\) is chosen to reduce resolution close to
  the minimum required before the form/features of the attractor become
  obscured.
\item
  \(\rho = 54.5\) is used to disturb said symmetry, off-setting the
  initial condition and orbital trajectories while preserving the
  two-lobe form.
\end{itemize}

Typical of many real-world problems, if the resultant co-ordinate
signals were stacked as per the weaving technique in \eqref{eq:Duffing_matrix}, a matrix
where \(t \gg m\) is produced:

\begin{equation}
\mathbf{X} = \begin{bmatrix}
x_{1} & x_{2} & \begin{matrix}
\cdots & x_{t}
\end{matrix} \\
y_{1} & y_{2} & \begin{matrix}
\cdots & y_{t}
\end{matrix} \\
z_{1} & z_{2} & \begin{matrix}
\cdots & z_{t}
\end{matrix}
\end{bmatrix} = \begin{bmatrix}
| & \cdots & | \\
\mathbf{x}_{1} & \cdots & \mathbf{x}_{t} \\
| & \cdots & |
\end{bmatrix} \in \mathbb{R}^{3 \times 1000}
\label{eq:Lorenz_matrix}
\end{equation}

This is un-wieldy for many operator-theoretic based techniques, and so
time-delay embeddings forming a Kyrlov subspace are commonly implemented
\cite{Kamb2020,BruntonChaos2016,Arbabi2017}. For the initial example, \(\mathbf{X}\)
in \eqref{eq:Lorenz_matrix} is augmented \(h = 300\) times, producing a time delay of
\(\mathrm{\Delta}t = 6s\) s.t. \(t - h = 700\) snapshots per row form
the Hankel matrix
\(\mathbf{H} \in \mathbb{R}^{m(h + 1) \times (t - h)} = \mathbb{R}^{903 \times 700} \):

\begin{equation}
\mathbf{H} = \begin{bmatrix}
\mathbf{x}_{1} & \mathbf{x}_{2} & \cdots & \mathbf{x}_{t - h} \\
\mathbf{x}_{2} & \mathbf{x}_{3} & \cdots & \mathbf{x}_{t - h + 1} \\
 \vdots & \vdots & \ddots & \vdots \\
\mathbf{x}_{1 + h} & \mathbf{x}_{2 + h} & \cdots & \mathbf{x}_{t}
\end{bmatrix} 
=
\begin{bmatrix}
\begin{bmatrix}
x_{1} \\
y_{1} \\
z_{1}
\end{bmatrix} & \begin{bmatrix}
x_{2} \\
y_{2} \\
z_{2}
\end{bmatrix} & \cdots & \begin{bmatrix}
x_{t - h} \\
y_{t - h} \\
z_{t - h}
\end{bmatrix} \\
 \vdots & \vdots & \ddots & \vdots \\
\begin{bmatrix}
x_{1 + h} \\
y_{1 + h} \\
z_{1 + h}
\end{bmatrix} & \begin{bmatrix}
x_{2 + h} \\
y_{2 + h} \\
z_{2 + h}
\end{bmatrix} & \cdots & \begin{bmatrix}
x_{t} \\
y_{t} \\
z_{t}
\end{bmatrix}
\end{bmatrix}
\label{eq:Lorenz_hankel}
\end{equation}

The use of time-delay embeddings is justified via Takens Embedding
theorem, stating that under certain conditions delay
embedding a system signal can reconstruct the original attractor up to a
diffeomorphism \cite{Colbrook2023,Arbabi2017}. The main challenge of this two lobed
Lorenz is that the Koopman operator in \(g(x) = x\) possesses a purely
continuous spectrum with no eigenvalues except a single constant
eigenfunction (\(\lambda = 1\)) \cite{Colbrook2023}. This suggests no global finite
invariant transform of the dynamics exists within \(g(x) = x\), limiting
a singular linear representation to approximating the `atomic'
projections without a solution for the `continuous' components \cite{Mezi2005,Mezi2004}.
W.r.t. \eqref{eq:invariant_decomposition_pO}, this introduces continuous spectra \('cS'\)
into the non-discernible partitions.

\subsubsection{Hankel Matrix -- DMD}\label{subsubsec:hankel-matrix-dmd}

\begin{figure}[!ht]
\centering
\includegraphics[
    width=\linewidth,
    trim={0 0.9cm 0 0},
    clip
]{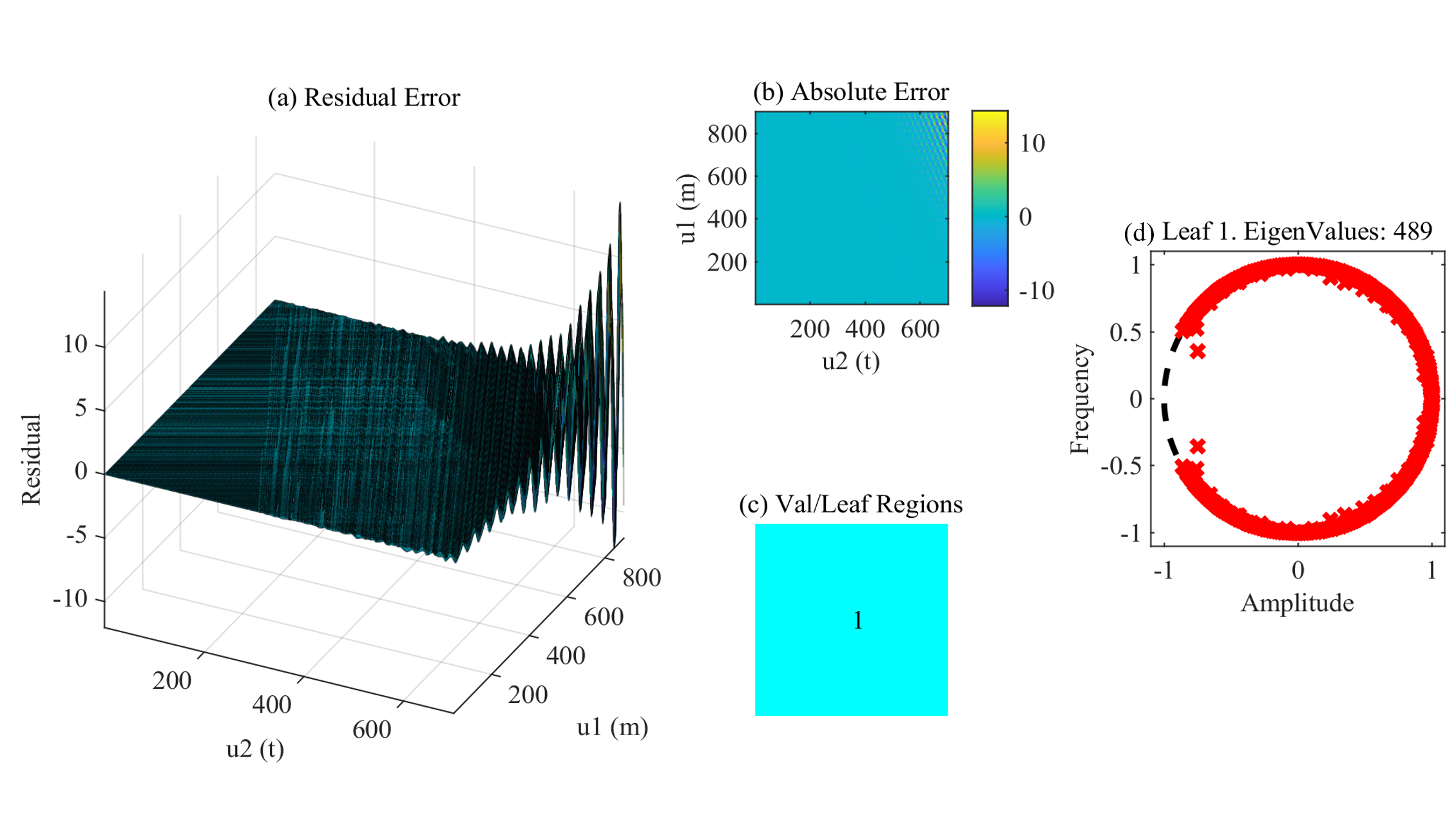}
\caption{: DMD (\(r = 1e - 4\)) on Lorenz hankel
matrix. Note \(Residual = 1 \Rightarrow Err = 100\%\)}
\end{figure}

Figure-16 depicts exact-DMD's performance with economy SVD on this
Hankel Lorenz example, with corresponding SVD spectra in SM-28. The
stated lack of eigenvalues is reflected by the continuous SVD spectra in
with only a singular prominent mode. The mostly unit circle eigen
decomposition from DMD, figure-16-(d), becomes unstable, oscillating
residuals exponentially as \('t'\) increases (NRMSE = 3.373). With
oscillations beginning at \(t = 20\), a distinct mismatch with the two
lobed attractor is immediately present. Unlike the Duffing example, the
much longer signal length here may reduce the ability for a singular set
of eigenfunctions to permeate the entire signal on two fronts, both by
limited single operator representation of a continuous spectra, and
solver noise aggravating DMD amplitude estimates. The latter is a known
issue when numerically estimating extended temporal trajectories in
chaotic systems \cite{Colbrook2023}.

\subsubsection{Hankel Matrix -- fSRD}\label{subsubsec:hankel-matrix-fsrd}

\begin{figure}[!ht]
\centering
\includegraphics[
    width=\linewidth,
    trim={0 0.2cm 0 0},
    clip
]{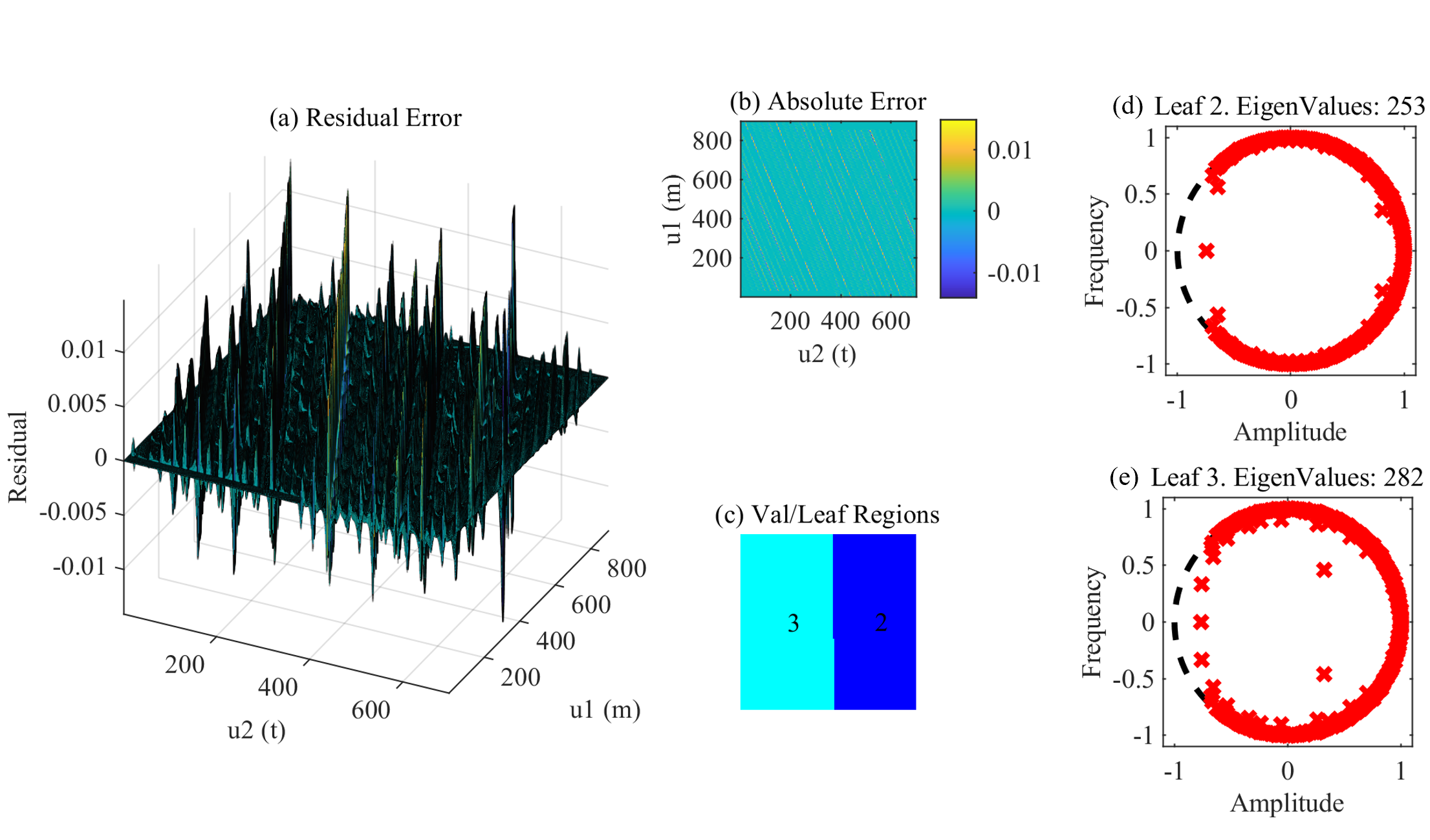}
\caption{: fSRD applied to Lorenz hankel matrix}
\end{figure}

fSRD applied to this Hankel Lorenz example is presented in figure-17. A
high accuracy model is produced (global NRMSE = 0.0068, Acc = 99.32\%)
via a purely temporal split (i.e., perpendicular to the temporal axis
\(u_{t}\)) located around \(t \approx 350\). Convergence occurs at
iteration-2, pruning back to the first split leaves. Compared to the DMD
results in figure-16, the use of two ergodic partitions split in time
successfully removes the long trajectory instability. As in figure-17
(d) and (e), given the split position leaf-3 uses 282 dimensions out
of a possible 383, with leaf-2 using 253 out of 342.

\begin{figure}[!ht]
\centering
\includegraphics[width=\linewidth]{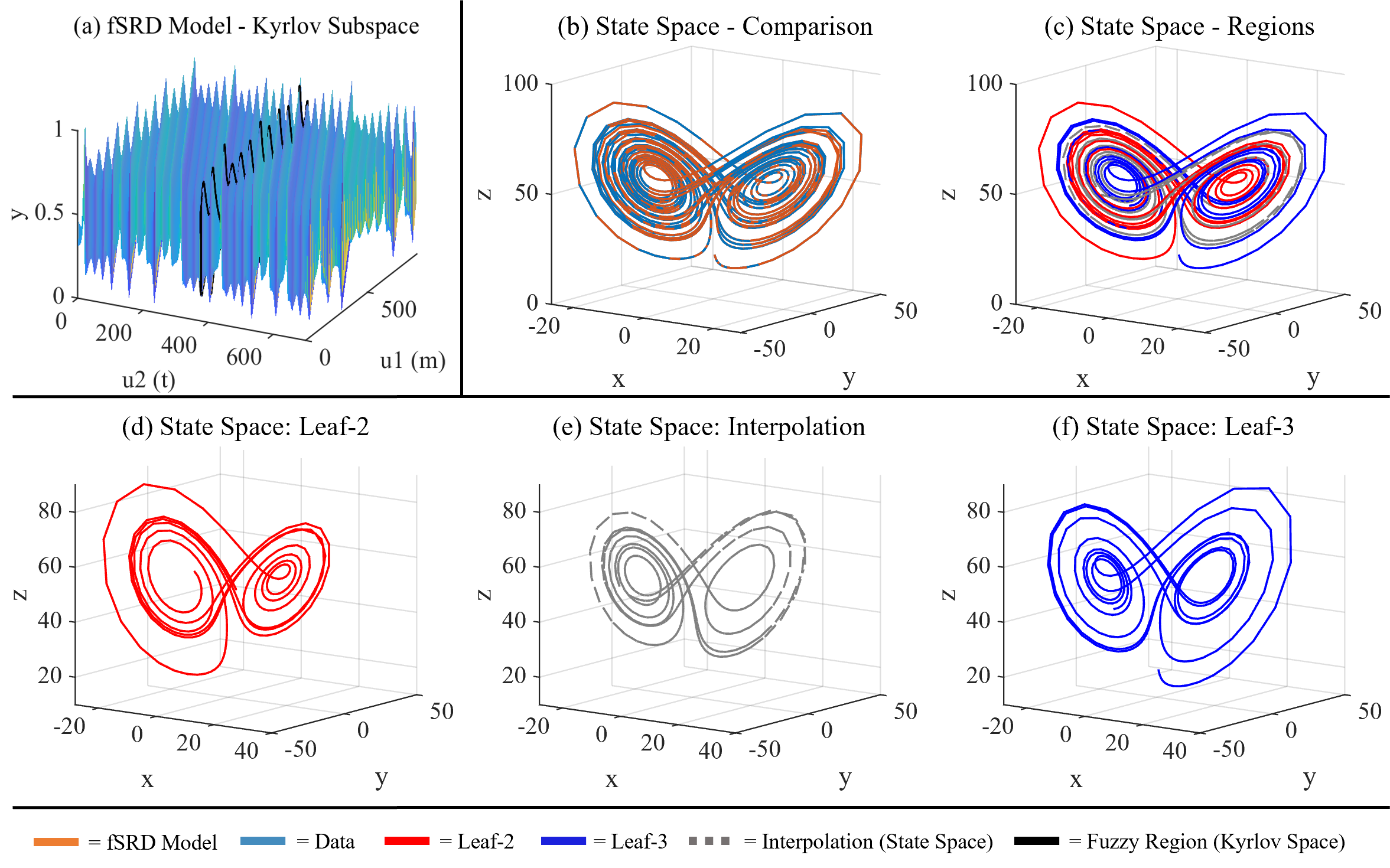}
\caption{: fSRD model of hankel Lorenz in state space with per region breakdown}
\end{figure}

Consider figure-18 where fSRD's state-space model is reconstructed from
the Kyrlov space. As shown, one can isolate the contributions of the
individual regions despite the use of time-delay embeddings. Each
ergodic partition appears to share a rough symmetry, respectively
capturing the outer orbits of one lobe and the inner orbits of the
other. However, unlike the Duffing examples, this does not correspond to
a clear segmentation of lobes. Such segmentation is demonstrated in
\cite{Sinha2020} for the \(\rho = 28\) attractor, utilizing two equivariant sets
but via a prior informed rotational symmetry. Results like this
highlight fSRD's limits w.r.t. interpretability due to the data's
co-ordinates. This manifests on two fronts, due to fuzzy interpolation
mixed with time-delay embeddings, and the limits of \(g(x) = x\).

\vspace{0.5em}
\textbf{Fuzzy Rules and Time-Delay Embeddings}

In fSRD each region's boundaries overlap its neighbours by design, with
fuzzy rules serving to smoothly transition between them. This fuzzy
region represents a numerical necessity, providing little insight in
isolation on the system. Within \(\mathbf{H}\) (the Kyrlov space),
fuzzy coverage is minimised by design to maximise local region validity.
Consequently, as in figure-6.18-(a), this covers a very narrow set of
columns relative to the total size of \(\mathbf{H}\)
(\(\sim 3/700\ cols \approx 0.4\%\) of Kyrlov signal). Re-assembling the
time-delay signals back to state space however expands the fuzzy
region's influence to a much larger portion of the total signal
(\(290/1000\ cols = 29\%\) of state-space signal). This is due to the
repeated elements in time-delayed embeddings, creating the interpolated
signal in figure-18-(e) that holds contributions from all three Kyrlov
sub-space fitted regions as per the breakdown in figure-19.

Such interpolation can limit the interpretability of each leaf
(linearity, placement, eigenmodes, etc) across the affected columns. In
this case, given the fuzzy-regions minor contribution in \(\mathbf{H}\),
vast portions of this state-space interpolation are dominated by one or
the other mode, with leaf-3 dictating \(\sim99.9\%\) of the
signal across the affected columns (Kyrlov fuzzy region contribution
\(< 0.01\%\)). Leaf-3 is thus valid for physical interpretation across
columns \(375 \rightarrow 650\). Columns \(650 \rightarrow 665\)
demonstrate the issue however, containing large contributions from the
Kyrlov fuzzy region with no dominant leaf, losing most interpretability.
While only minor here, the contamination will differ for each data set,
scaling with several factors (split angle, Kyrlov fuzzy region width,
number of splits, etc). fSRD's overall interpretability can thus be
reduced when used with time-delay embeddings.

\begin{figure}[!ht]
\centering
\includegraphics[
  width=12cm,
  height=7cm]{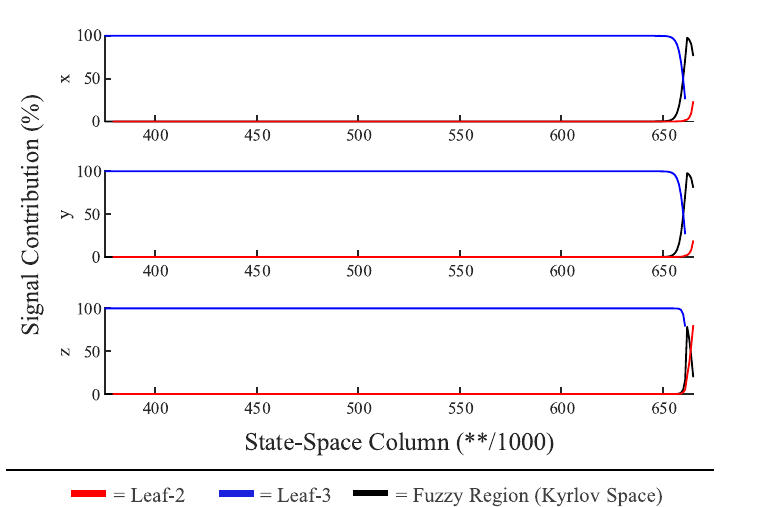}
\caption{: Region contributions to state
space interpolated signal, col's 375-650 of \eqref{eq:Lorenz_matrix}}
\end{figure}

\newpage
\textbf{Pure Continuous Spectra with g(x) = x}

Consider figure-20 where fSRD discretizes the Lorenz attractors
continuous spectra dynamics to combine viable restricted spectral
integral estimates. As evident by figure-20-(b), the continuous spectra
form is passed into each partition's SVD spectra. While each spectra
progressively declines, most modes hold little majority contribution
over the others. This is because the Lorenz signal's dynamics remain
unchanged from the time-delayed state-space observable \(g(x) = x\). 
Via truncation, 700 initial columns produces two regions with 282 and 253 
modes respectively. With regularization, no further modes are removed, given each
regions regularization coefficient converged close to zero, as shown in figure-20-(c). 
Regularization is thus unable to collapse the spectra. Many DMD extensions have been
shown to collapse such continuous spectra for the \(\rho = 28\) Lorenz
example with vastly reduced dimensionality compared to these results.
This is achieved however via an increased reliance on the data
quantity/quality beyond simple state space measures (rigged Hilbert
space, increasing \(m\), etc) \cite{Colbrook2023,Madrid2005}, and the utilization
of prior information (prior library of observables, imposing
pre-identified global properties as observables, etc) \cite{Sinha2020,ColbrookEDMDmp2023}.

\begin{figure}[!ht]
\centering
\includegraphics[scale=0.85]{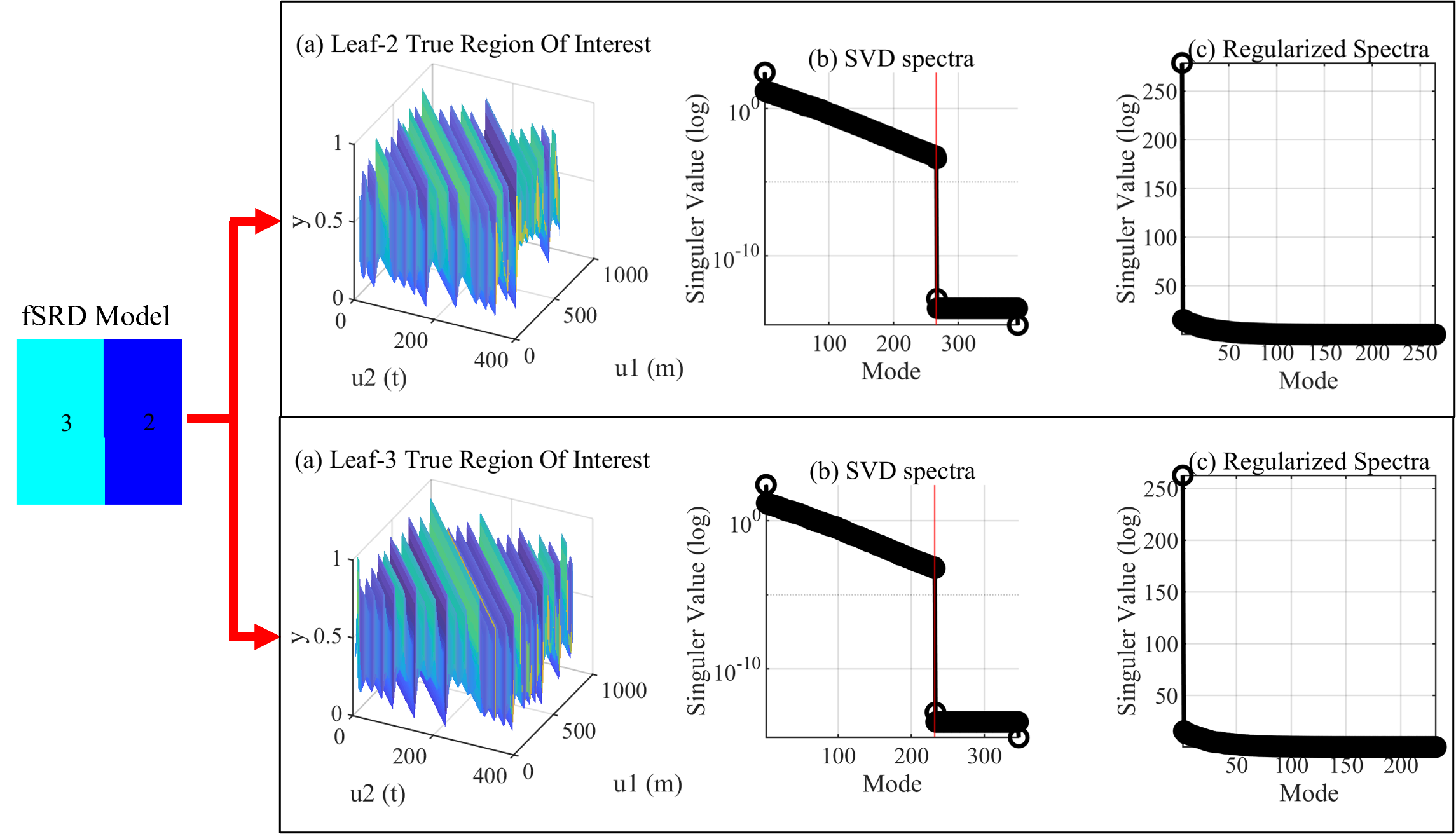}
\caption{: SVD Spectra for each region of hankel Lorenz fSRD model}
\end{figure}

Again, these methods sacrifice automatability unlike fSRD with its broad
multi-attractor adaptability independent of priors. Even though
accurately estimated by the localised restricted spectral integral
framework, continuous spectrum applications like this may not contain
split positions s.t. fSRD can collapse the spectra in \(g(x) = x\)
(i.e., reduce dimensionality) regardless of partition number. This
inability corresponds to a lack of `real' finite embeddings even in
localized regions, highlighting concerns in Section~\ref{subsec:ergodic-partitioning-of-continuous-spectra} with even the
automated truncation and regularization hitting limits to improvement.
In some cases with large continuous spectra, fSRD applied with
\(g(x) = x\) thus retains automatability at the cost of model
complexity.

\newpage

\subsection{Bifurcations, Discontinuity, and Data
Co-ordinates}\label{subsec: bifurcations-discontinuity-and-data-co-ordinates}

\FloatBarrier

\begin{figure}[!ht]
\centering
\includegraphics[width=\linewidth]{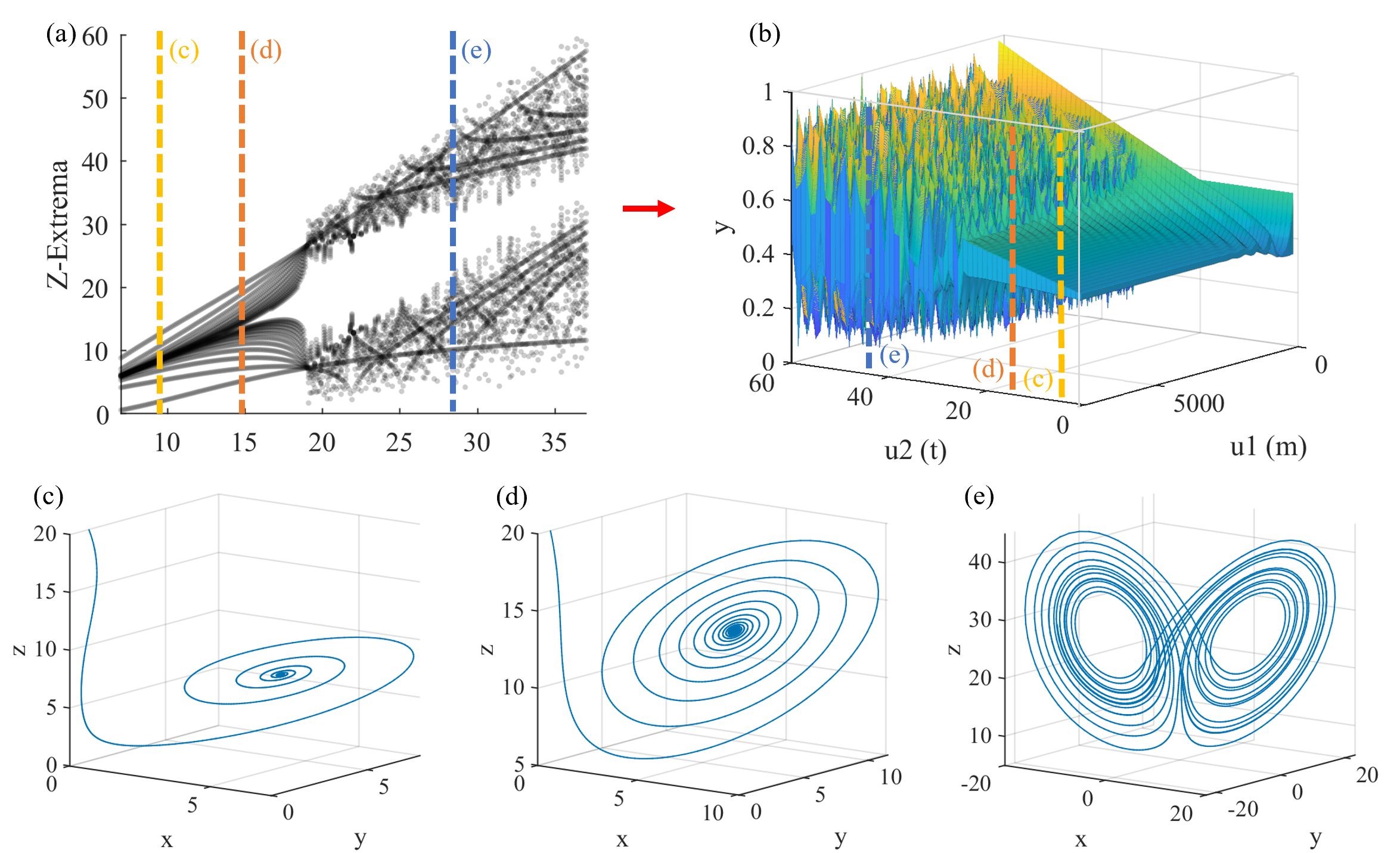}
\caption{: Bifurcation graph of Lorenz along \(\rho\). (c)
corresponds to \eqref{eq:Lorenz_bifurcation_matrix}, while (d)-(f) highlight
shifts in global behaviour}
\end{figure}

\FloatBarrier

The limitation in local dimensionality shown for systems with pure
continuous spectral dynamics can be readily tackled via the use of
co-ordinate transformations, e.g., ideal observables \(g(x)\) or
alternative data co-ordinates. While future work may attempt to retain
fSRD's advantages while utilizing the former, it will be shown here that
simple alterations to the inputted data's co-ordinates can vastly
improve interpretability.

Consider again Lorenz's bifurcation graph along the variable \(\rho\) in
figure-21. At \(\rho \approx 18\), a bifurcation occurs, drastically
changing the behaviour of the Lorenz system in the
\(\lbrack x,y,z\rbrack\) co-ordinates from a singular fixed-point
attractor, figure-21-(c), into the well-known 2-lobed attractor,
figure-21-(e) \cite{Sparrow1982}. The former also makes a transition,
figure-21-(d), where the trajectory `unravels', increasing trajectory
time in outer orbits as \(\rho\) increases until the fixed point
disappears just before bifurcation. Bifurcations in dynamical systems
exemplify abrupt qualitative changes in system behaviour. Although the
underlying governing equations remain continuous, such transitions also
serve to illustrate apparent discontinuities manifesting in observed
data, non-stationary signals, etc, as regime shifts dependant on the
sampling strategy employed. fSRD can be applied to the full region
depicted in figure-21-(a), modelling the chaotic bifurcations of the
Lorenz attractor with \(\rho\) as the evolution variable rather than
time.

Consider an individual Lorenz attractor:

\begin{equation}
\mathbf{X}(\rho)_{lorenz} = \begin{bmatrix}
x_{1} & x_{2} & \begin{matrix}
\cdots & x_{t}
\end{matrix} \\
y_{1} & y_{2} & \begin{matrix}
\cdots & y_{t}
\end{matrix} \\
z_{1} & z_{2} & \begin{matrix}
\cdots & z_{t}
\end{matrix}
\end{bmatrix}_{\rho} = \begin{bmatrix}
 - & \mathbf{x}_{\rho} & - \\
 - & \mathbf{y}_{\rho} & - \\
 - & \mathbf{z}_{\rho} & - 
\end{bmatrix}_{\rho} \in \mathbb{R}^{3 \times 2500}
\label{eq:Lorenz_single_attractor}
\end{equation}

Sampling is increased to \(\mathrm{\Delta}t = 0.008\) s.t. the dynamics of
\(\rho\) can be sufficiently captured. Each column of
\(X(\rho)_{lorenz}\) may be stacked to contain the full description of
the attractor \(\lbrack x,y,z\rbrack_{\rho}\) for a given value of
\(\rho\) in a single column of \(\mathbf{X}\), creating a weaving effect much
like \eqref{eq:Duffing_matrix}. Taking increments of \(\mathrm{\Delta}\rho = 0.5\) within
the chosen window \(\rho_{start} = 7 \rightarrow \rho_{end} = 36.5\)
produces \(\mathbf{X} \in \mathbb{R}^{7500 \times 60}\):

\begin{equation}
\mathbf{X} = \begin{bmatrix}
x_{1,\rho_{start}} & x_{1,\rho + \mathrm{\Delta}\rho} & \cdots & x_{1,\rho_{end}} \\
y_{1,\rho_{start}} & y_{1,\rho + \mathrm{\Delta}\rho} & \cdots & y_{1,\rho_{end}} \\
z_{1,\rho_{start}} & z_{1,\rho + \mathrm{\Delta}\rho} & \cdots & z_{1,\rho_{end}} \\
 \vdots \  & \vdots & \ddots & \vdots \\
x_{t,\rho_{start}} & x_{t,\rho + \mathrm{\Delta}\rho} & \cdots & x_{t,\rho_{end}} \\
y_{t,\rho_{start}} & y_{t,\rho + \mathrm{\Delta}\rho} & \cdots & y_{t,\rho_{end}} \\
z_{t,\rho_{start}} & z_{t,\rho + \mathrm{\Delta}\rho} & \cdots & z_{t,\rho_{end}}
\end{bmatrix}
\label{eq:Lorenz_bifurcation_matrix}
\end{equation}

This corresponds to the matrix depicted in figure-21-(b). Compared to
the time-delay Lorenz system, the number of columns total for the
evolution variable is vastly reduced from 700, limiting the dimensions
for fitted models to \(\leq 60\) (e.g., in figure-20 fSRD's continuous
spectra partitions were minimum 253). Further, chaotic dynamics
demonstrated up to this point involved minor changes in input parameters
drastically changing system trajectories, for example, transition into
new attractors. Bifurcations face the same challenges (i.e., no global
topological conjugate) but represent an extreme s.t. practical sampling
often interprets them as discontinuities \cite{Sprott2003}. This parallels the
sudden changes over time due to external forcing typical in control
applications \cite{Ott1990}. While these changes to system data would often
require completely different modelling approaches to tackle, fSRD adapts
without algorithm alteration, benefiting from the co-ordinate change of
the evolution variable by shifting the continuous spectra dynamics
within the spatial parameters (i.e., rows). Within this proof of concept however,
the novel fixed-point regularization is not designed to deal with matrices of this size,
racking considerable computation time due to the scaling with rows (SM-15.2, \(O(f^{2})\)).
As such the following results will include only the truncation stage in SM-21 (stage IV),
otherwise assuming \(\delta_{BIC}=0\) (in stage V). Extending this procedure further to 
high-dimensional settings is left for future work.

\subsubsection{Lorenz Bifercation -- DMD}\label{subsubsec:lorenz-bifercation-dmd}

\begin{figure}
\centering
\includegraphics[width=\linewidth]{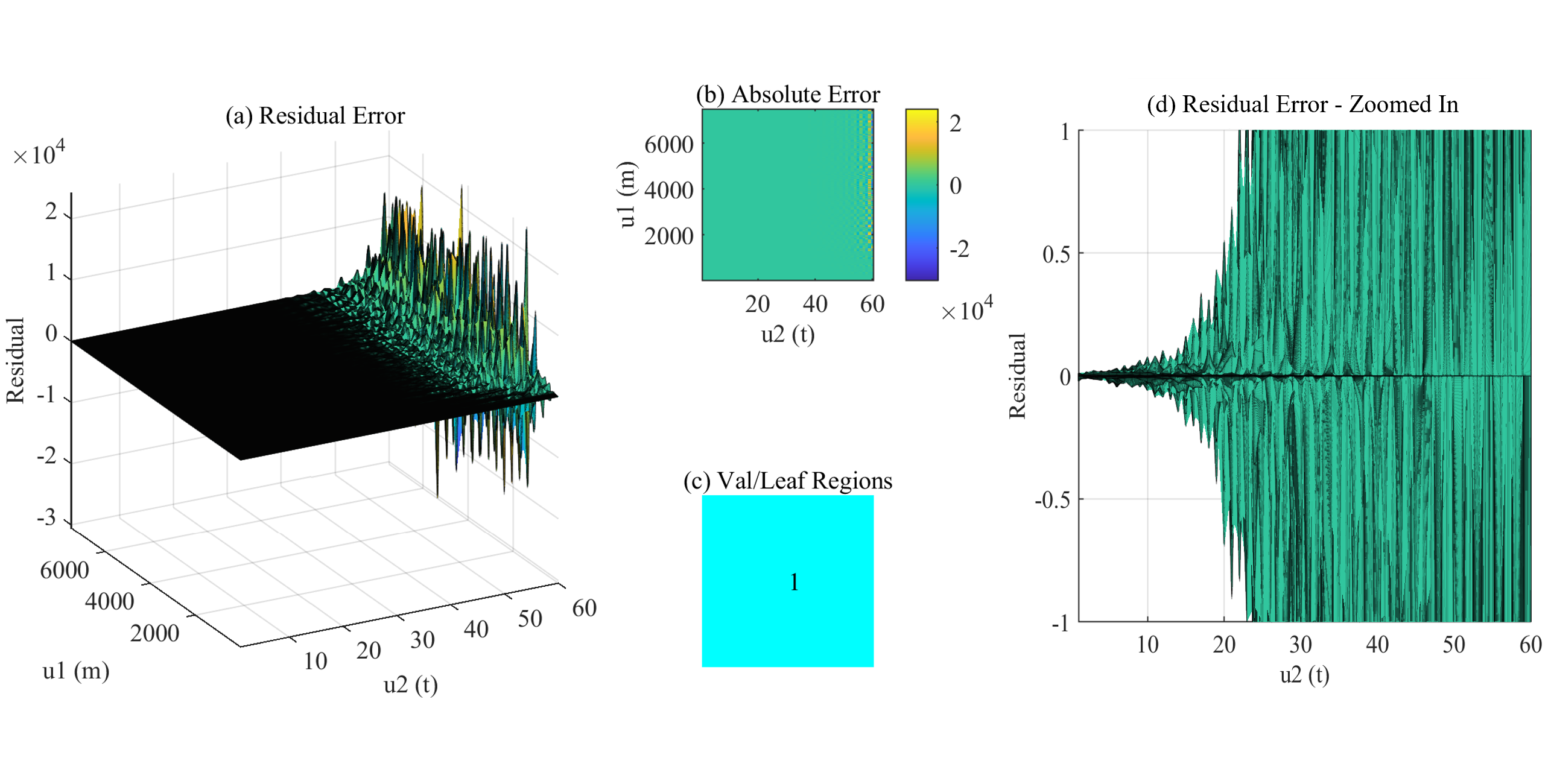}
\caption{: DMD (\(r = 1e - 4\)) on Lorenz
bifurcation region \eqref{eq:Lorenz_bifurcation_matrix}}
\end{figure}

Figure-22 depicts exact-DMD with economy SVD on \eqref{eq:Lorenz_bifurcation_matrix}. The single
linear operator becomes unstable the further into the trajectories it
extends (\(NRMSE = 13686\)). As shown in figure-22-(d), this is because
the operator predominantly fits the stable portion of the single
fixed-point attractor (\(u_{2}(t) = 1 \rightarrow 9\)), with residuals
dramatically rising as the initial attractor transitions
(\(u_{2}(t) \approx 10)\), and exponentially beyond 100\% error
(\(Residual > 1\)) upon bifurcation (\(u_{2}(t) \approx 21\)).

\subsubsection{Lorenz Bifercation --
fSRD}\label{subsubsec:lorenz-bifercation-fsrd}

\begin{figure}
\centering
\includegraphics[width=\linewidth]{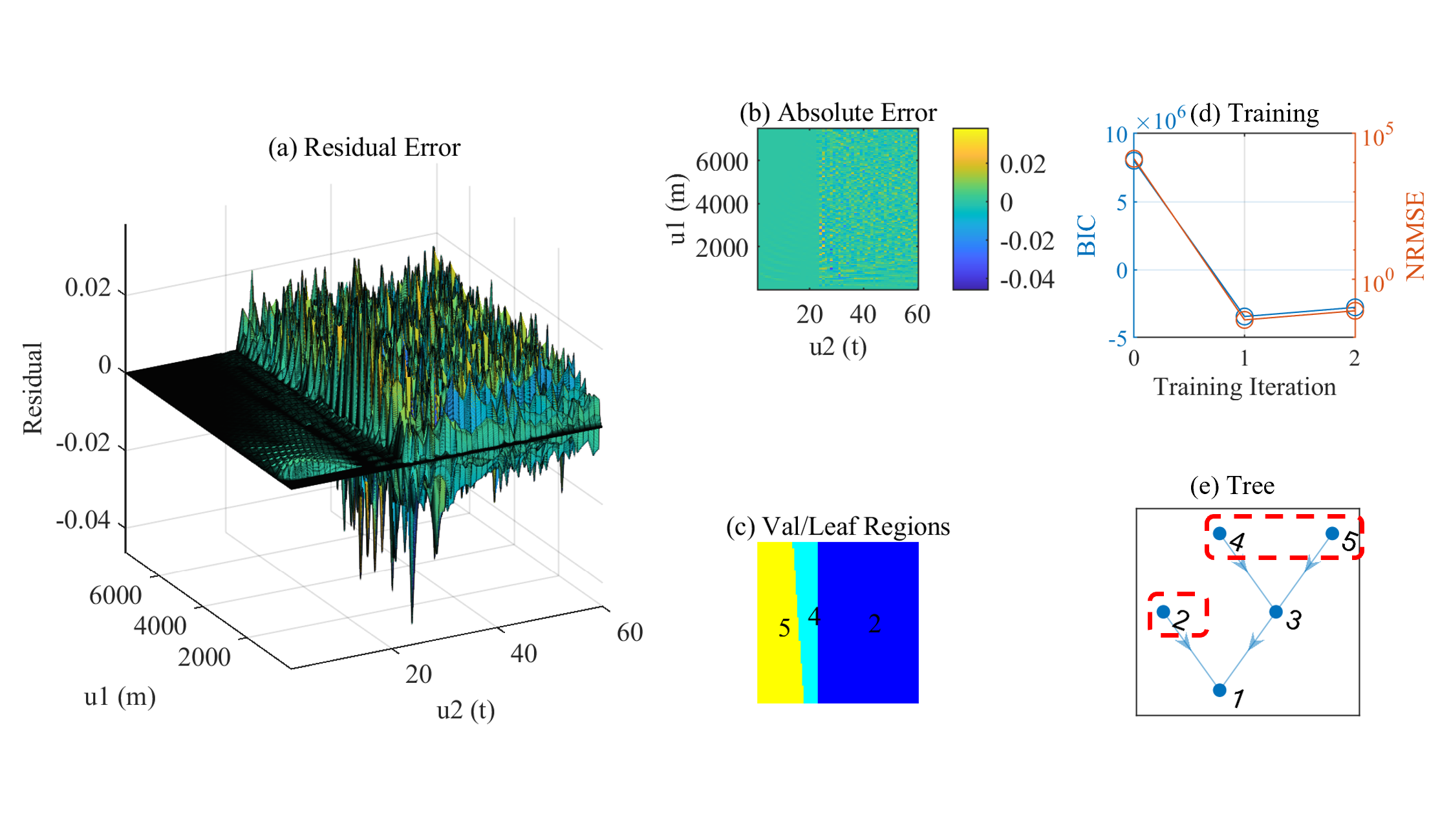}
\caption{: fSRD (\(r = 1e - 4\)) on Lorenz bifurcation region \eqref{eq:Lorenz_bifurcation_matrix}}
\end{figure}

Now consider fSRD's results depicted in figure-23. Training concludes at
the 2nd iteration, pruning back to include three local regions as in
figure-23 (d) and (e). Global predictive performance vastly improves
compared to exact-DMD, increasing to 96.23\% accuracy (NRMSE=0.0377).
W.r.t. to split location, figure-23-(c), the invariant interface between
leaf-2 and leaf-4 occurs at the bifurcation, angled purely in \(\rho\)
(vertical) s.t. the fuzzy region aligns directly with the bifurcation
columns (\(u_{2}(t) \approx 22\)). This isolates the two-lobe attractor to a
single linear partition, but as per the residuals in figure-23-(a)
contains some low amplitude residuals. Leaf-4 and leaf-5's interface
starts from the transitionary region where DMD's residuals began to
rise\((u_{2}(t) \approx 10\)), with a slight spatial bias towards the
early stages of the attractor trajectories. This segments the single
fixed-point attractor into a stable region (leaf-5) and a transitionary
region (leaf-4).

\vspace{0.5em}
\textbf{Region Placement}

\begin{figure}[!ht]
\centering
\includegraphics[width=\linewidth]{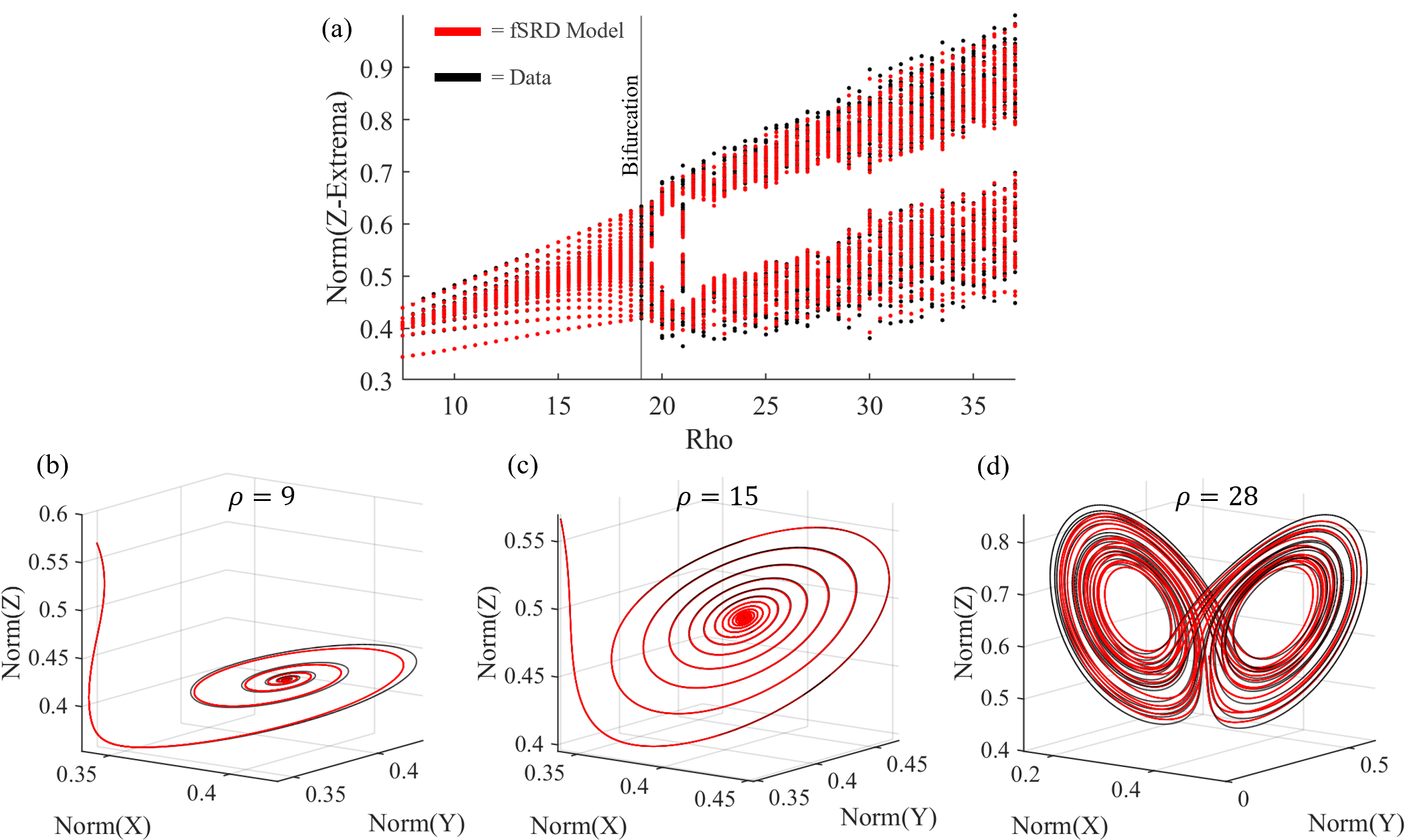}
\caption{: fSRD's Lorenz bifurcation model compared with example
state-space attractors of interest}
\end{figure}

Figure-24 compares the fSRD model against data within the original state
space. The fully automated optimization assigns partitions that
correspond well to one of the three described physical transitions,
accurately locating and modelling each as per (b), (c) and (d) with
leaf-5, leaf-4, and leaf-2 respectively. While both amplitude and form
are mostly captured, the increased residuals in the third two lobed
attractor is now evident from (a) and (d), with small shifts in outer
orbitals slightly missing z-extrema values.

\begin{figure}[!ht]
\centering
\includegraphics[width=\linewidth]{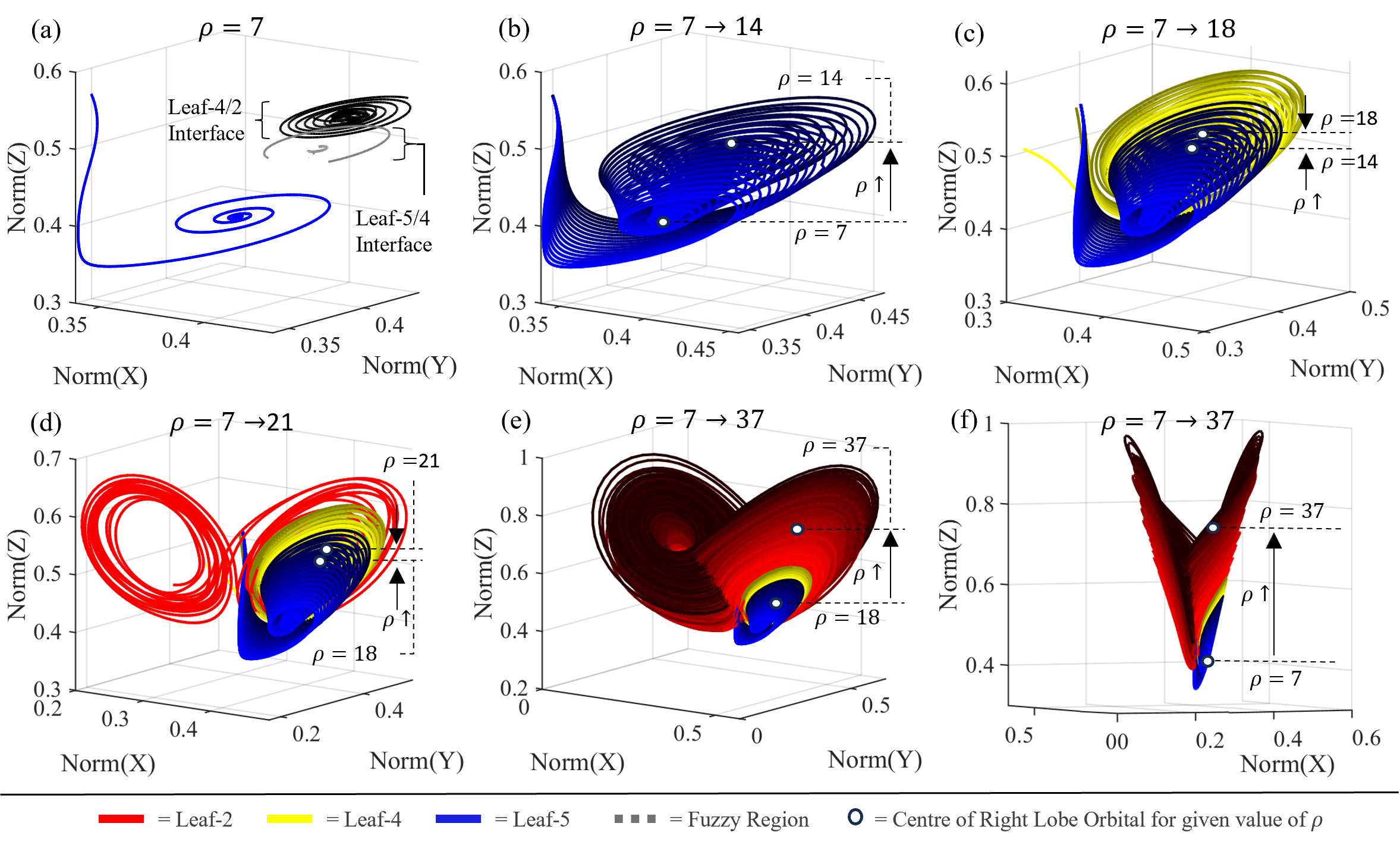}
\caption{: fSRD model regions as \(\rho\)
increases, highlighting interpretability. Light to darker colour
transition corresponds to \(\rho\)'s increase in each region}
\end{figure}

Figure-25 demonstrates via visual graphics how these local regions can
be used for system interpretation. Each column of the fSRD model is
plotted in the original state space such as to evaluate the evolution of
the system with \(\rho\). As depicted, leaf-5 begins at \(\rho = 7\) in
(a), comprised of an orbit about a fixed point and an offshoot
trajectory or `tail'. Proceeding in (b), the attractor translates
position and increases outer orbital density with minor change to its
overall size/behaviour. When leaf-4 takes over in (c), minor translation
occurs, but the outer orbital diameter increases rapidly as the inner
orbitals become non-circular/irregular, no longer converging to a fixed
point. In (d), when leaf-2 begins, a 2\textsuperscript{nd} lobe suddenly
emerges from the `tail' of the original attractor. This creates a new
orbital to the left while the original remains on the right, both
translating position and steadily increasing in size as \(\rho\)
increases in (e).

Alternatively, interpreting the eigenvalue distribution of each leaf,
figure-26, provides a direct interpretable metric to reach similar
conclusions:

\begin{itemize}
\item
  Leaf-5: low dimensional with slightly damped unit circle spectra such
  that all \(|\lambda| \leq 1\), i.e., stable attractor where
  trajectories decay over time to a fixed point \cite{Mezic2017}.
\item
  Leaf-4: retains low complexity but large under-damped spectra
  (\(|\lambda| > 1\)) suggesting an unstable attractor undergoing rapid
  growth in several directions \cite{Mezic2017}.
\item
  Leaf-2: greatly increases relative complexity (i.e.,
  number-of-eigenvalues) to capture the complex attracter. With one
  large real-valued eigenvalue and a slight underdamping in the
  remaining spectra, the attractor is unstable, undergoing mild growth
  in several directions (size increase) and strong exponential growth in
  one direction (position translation) \cite{Mezic2017}.
\end{itemize}

\FloatBarrier

\begin{figure}[!ht]
\centering
\includegraphics[width=\linewidth]{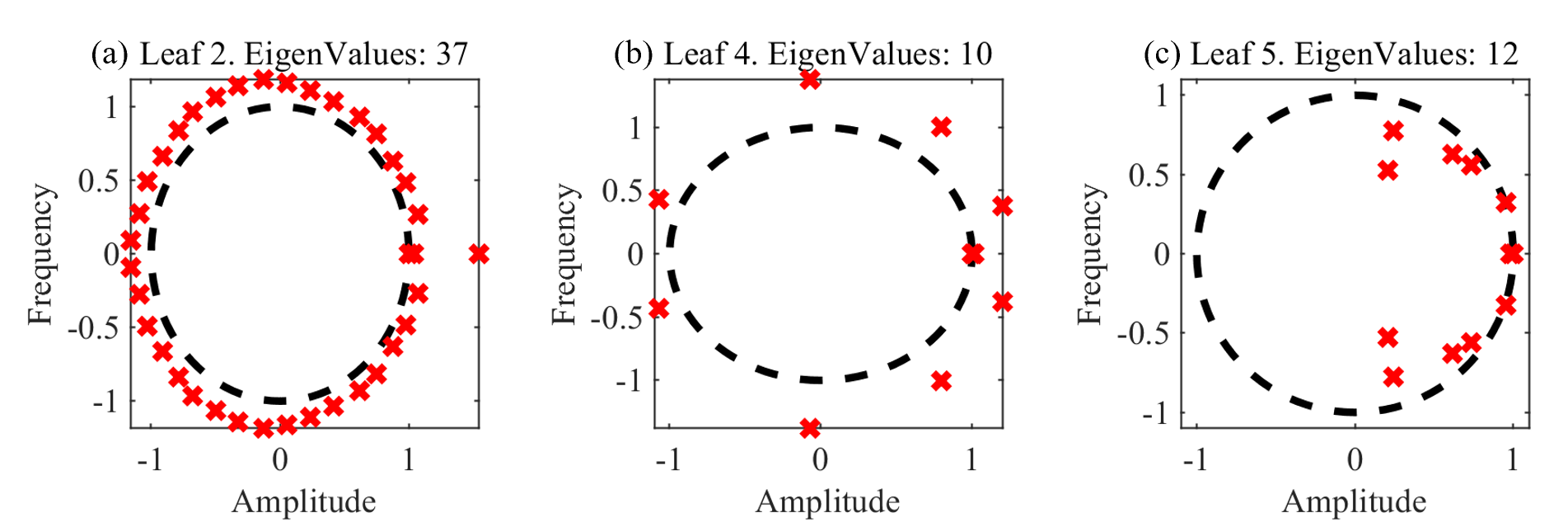}
\caption{: Eigenvalue spectra of fSRD model per region}
\end{figure}

\FloatBarrier

\vspace{0.5em}
\textbf{Modal Breakdown}

\FloatBarrier

\begin{figure}[!ht]
\centering
\includegraphics[width=\linewidth]{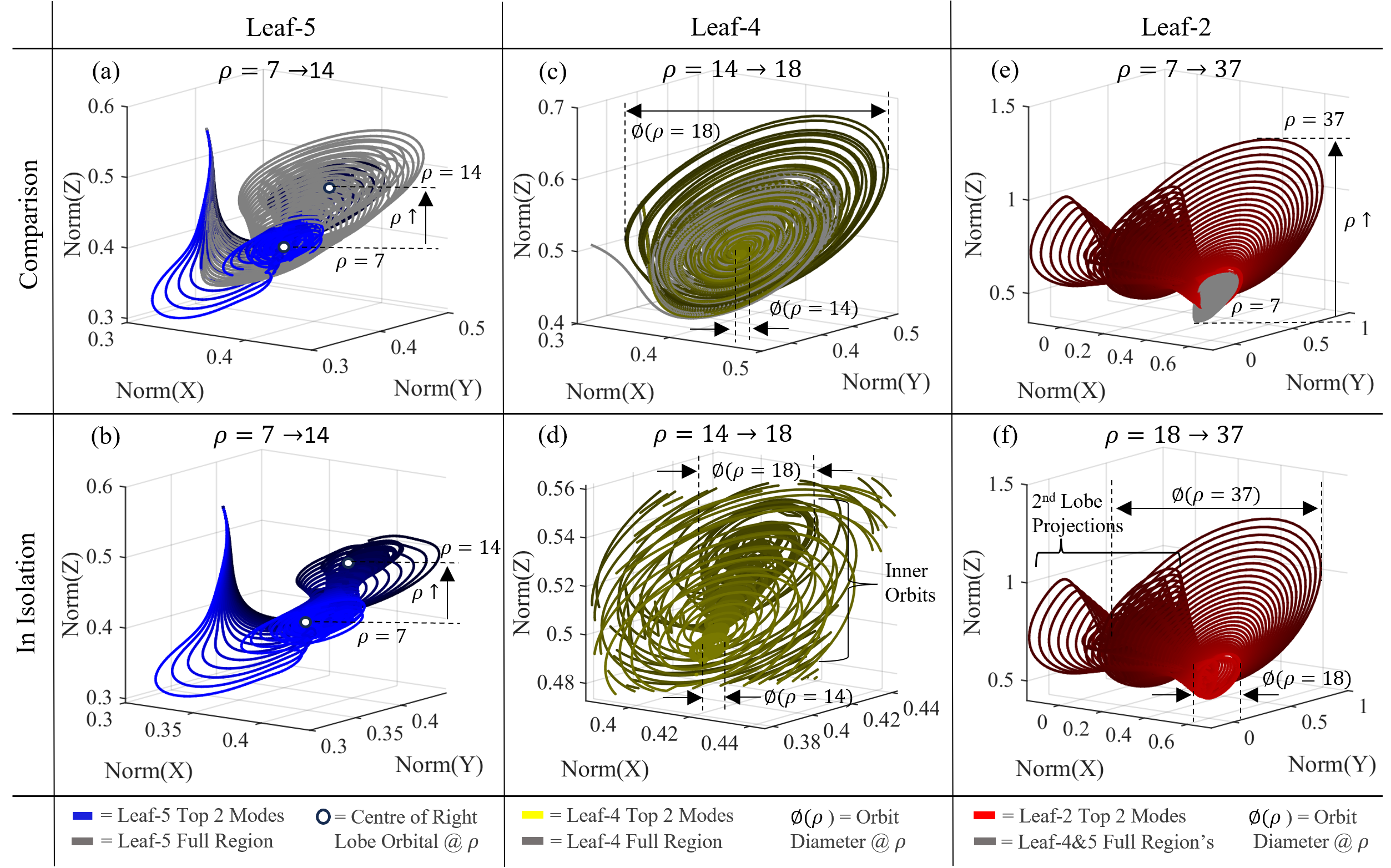}
\caption{: Comparison with first two modes of each region (colours darken as \(\rho\) increases)}
\end{figure}

\FloatBarrier

Koopman spectral decomposition allows for modal assessment, decomposing
and evaluating resultant operators into spatial temporal coherent modes
(i.e., eigenmodes) \cite{Brunton2021}. In fSRD, this can be implemented for each
identified region, allowing assessment of both their individual
contribution and characteristics without prior knowledge as the
eigenvalue assessment in figure-26 demonstrates. These orthogonal linear
modes can be further interpreted in several ways. To what extent each
interpretation provides insight on the system, which by extension
dictates their practical use, depends largely on the specific
application (e.g., system co-ordinates, how spatial parameters were
compiled into \(X\), chosen evolution variable, etc.).

As modal assessment via Koopman is well documented \cite{Brunton2021,Koopman1931,Mezi2005,Tu2014,Proctor2016}
and largely case specific, only a brief
visual example will be provided here. Consider figure-27, where the
contribution to the predicted system via a restricted subset of the
eigen spectra is depicted. Note that isolated modal projections
primarily indicate trends, with exact magnitudes potentially mismatching
the full prediction without the contribution of other modes in the
spectra, e.g., a complex conjugate pair may dampen and/or translate a
leading transient mode. With a chosen subset of modes
\(j = 1,\ldots,r_{i}\), each region \(i = 1,\ldots,p\) has a general
output of the form:

\begin{equation}
\begin{split}
{\widetilde{\mathbf{X}}}_{i} 
=
\varnothing_{i}( \mathbf{U} )\ 
\odot
\ {\vartheta_{i}^{- 1}\{\mathbf{\Phi}}_{i}diag( \mathbf{b}_{i} )\mathbf{T}^{t}( \mathbf{\omega}_{i} )\},
\ \ \text{s.t.,}\ \
\\
{\widetilde{\mathbf{X}}}_{i,j} = \varnothing_{i}( \mathbf{U} )\ 
\odot
\ \vartheta_{i}^{- 1}\left\{ \phi_{i,j}diag( b_{i,j} )\mathbf{T}^{t}( \omega_{i,DMD,j} ) \right\}
\label{eq:modal_assessment}
\end{split}
\end{equation}

Figure-27 depicts the leading \(r_{i} = 2\) for each region aggregated
s.t., \(b_{i} = diag\lbrack b_{i,1},b_{i,2}\rbrack\). As the highest
amplitude modes with no imaginary componant, these are expected to
correspond to phenomena such as dominant transient dynamics and stable
or unstable long-term behaviour \cite{Brunton2021}. The graphic can be interpreted
as such:

\begin{itemize}
\item
  In (a) and (b) leaf-5's leading modes isolate the initial attractors
  main translational vector through the 3-dimensional space. Further,
  they appear to dictate the form of the inner orbits, unravelling
  slightly as \(\rho \uparrow\).
\item
  Leaf-4's leading modes isolate the mentioned orbit expansion as
  \(\rho \uparrow\), both for the outer and inner orbitals as in (c) and
  (d) respectively. This notably excludes both the tail and any
  irregular orbits leading into the bifurcation which are managed by the
  remaining underdamped spectra.
\end{itemize}

\begin{itemize}
\item
  In (c) leaf-2's leading modes dictate the rapid expansion in diameter
  of the initial right lobe/attractor, also including part of its
  transient dynamics across the system co-ordinates. Trajectories
  emerging from the tail and orbit centroid of the right attractor also
  initiate projection onto the region in which the new lobe is expected
  to emerge. Notably the remaining eigenvalues in leaf-2 form a
  continuous spectra as per
  figure-6.33-(a), resulting in modes that are uninterpretable in
  isolation (i.e., require aggregation into large groups).
\end{itemize}

%% file: sections/results_solar.tex
\newpage

\subsection{Solar Imagery}\label{subsec:solar_imagery}

\FloatBarrier

\begin{figure}[!ht]
\centering
\includegraphics[
    width=15cm,
    height=7.5cm,
    trim={0 0.8cm 0 0},
    clip
]{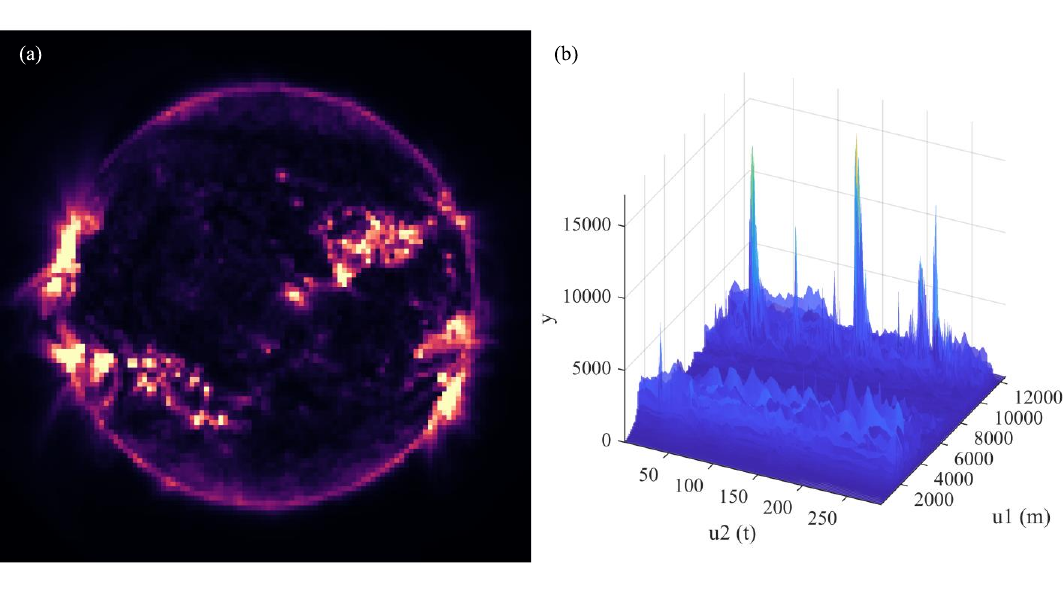}
\caption{Solar data "Mother's Day Storm", (a) example and (b) snapshot matrix}
\label{solar-data-example}
\end{figure}

\FloatBarrier

Solar imagery provides a compelling testbed for complex dynamics. Observations
across multiple extreme ultraviolet wavelengths capture different plasma temperatures
and solar layers \cite{Lemen2012}, producing signals with quasi-periodic behaviour, evolving
baselines, and strongly nonlinear, intermittently chaotic temporal dynamics.
The data also exhibit pronounced spatial heterogeneity, with coexisting active 
regions and quiet areas, as well as a wide range of transient phenomena from 
small-scale fluctuations to large eruptive events \cite{Benz2017}. 

These properties are representative of a broad class of real-world spatiotemporal 
systems, with similar challenges observed in domains such as computer vision, 
remote sensing, and geophysical fluid dynamics (satellite-based Earth observation, 
weather and climate systems, environmental monitoring, etc.). As a result, 
solar imagery provides a challenging real-world application in this study to 
demonstrate fSRD's generality, expressive power, and ability to extract interpretable
structure on a high dimensional and noisy real-world dataset, complimenting the 
results on the earlier controlled benchmark systems. In the present work the data 
is used solely for these purposes, though it was previously used by the first author 
in an exploratory anomaly detection project \cite{Bokor2026solar}.

The dataset used here consists of extreme ultraviolet observations at \(171\)~\AA \ from the
Atmospheric Imaging Assembly (AIA) \cite{Lemen2012} aboard the Solar Dynamics Observatory. 
The images were retrieved using the SunPy Python library \cite{Barnes2020} and the Virtual Solar
Observatory. The dataset covers the period 00:00 UTC May 10 to 00:00 UTC May 12, 2024.
Images were sampled at 10-minute intervals, producing 288 images of size \(4096\times4096\) 
pixel. 

Due to computational constraints, each image was downsampled to \(112\times112\) pixels.
Each image, \(\textit{I}_{t} \in \mathbb{R}_{112\times112}\), was then vectorized by stacking column-wise 
raster ordering (column-major indexing), yielding the snapshot vector \(\mathbf{x}_{t}\).
Specifically, pixels were ordered s.t.,:

\begin{equation}
\mathbf{x}_{t}
=
\left[
I_t(1,1), I_t(2,1), \ldots, I_t(112,1),
I_t(1,2), I_t(2,2),
\ldots,
I_t(1,112), \ldots, I_t(112,112)
\right]^{\top}
\label{eq:solar-vectorized-image}
\end{equation}

the snapshot matrix is then given by:

\begin{equation}
\mathbf{X}
=
\left[
\mathbf{x}_{1},\mathbf{x}_{2},\cdots,\mathbf{x}_{288}
\right]
\in
\mathbb{R}^{12544\times288}
\label{eq:solar-snapshot-matrix}
\end{equation}

An example \(\textit{I}_{t}\) and the snapshot matrix \eqref{eq:solar-snapshot-matrix} are demonstrated
in figure-\ref{solar-data-example}. This interval was chosen as it includes the 'Mother's 
Day Solar Storm' event reported by NASA \cite{MaraJohnson-Groh2024}, containing several large solar flares and
coronal mass ejections that produced one of the largest solar storms to reach Earth in two 
decades \cite{Schmlter2025}. Unlike quiet Sun periods, which often exhibit low-amplitude, slowly evolving
dynamics, this event emphasises the large-amplitude, transient, and spatially heterogeneous
activity discussed above.

\subsubsection{Mother's Day Storm -- fSRD }\label{mothers-day-storm-fsrd}

\FloatBarrier

\begin{figure}[!ht]
\centering
\includegraphics[width=\linewidth]{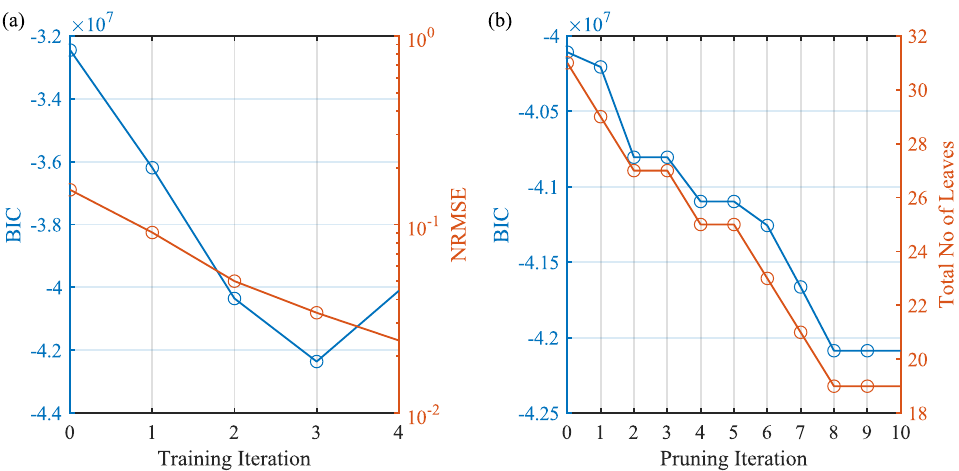}
\caption{fSRD on Solar data, (a) convergance and (b) pruning results}
\label{solar-data-fsrd-model-convergance}
\end{figure}

\FloatBarrier

Figure-\ref{solar-data-fsrd-model-convergance} depicts the convergence results of fSRD when applied to Solar
data from \eqref{eq:solar-snapshot-matrix}. Note total leaves refers 
to all nodes in the tree (i.e., active regions and parents). Starting at \(~85\%\) 
reconstructive accuracy with one region, fSRD's forward pass concluded at the 4th 
iteration, peaking at \(~98\%\) (NRMSE = 0.023, 16 active 
regions), before pruning back for a more parsimonious \(~97\%\) (NRMSE = 0.03, 10 active regions).
This final model included 837 total eigenmodes across regions.
Again, \(\delta_{BIC}=0\) for all local regions in this example, as the current fixed-point iteration  
implementation is not optimized for datasets of this scale (SM-14).
This is thus expected to be inflated given the lack of regularization. 

\newpage

\begin{figure}[!ht]
\centering
\includegraphics[
    width=\linewidth,
    trim={0 0.9cm 0 0},
    clip
]{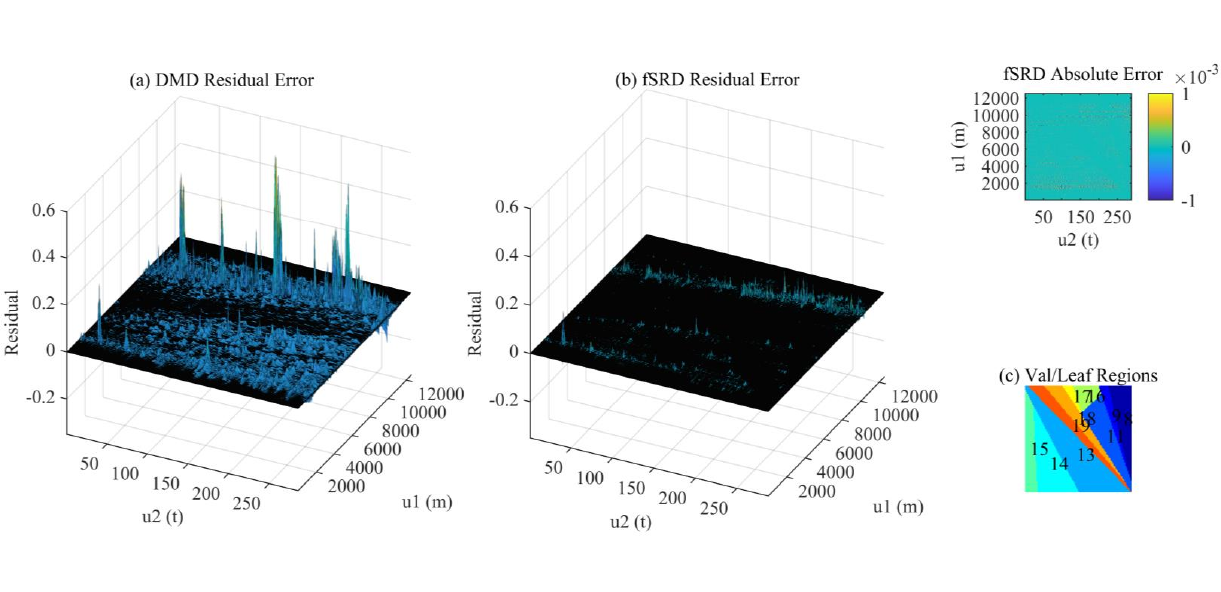}
\caption{(a) DMD and (b) fSRD residuals on Solar data}
\label{solar-data-fsrd-residuals}
\end{figure}

Consider the residual plots provided in figure-\ref{solar-data-fsrd-residuals} across the representational space.
Figure-\ref{solar-data-fsrd-residuals}-(a) corresponds to DMD with a noise model (SVDe) and 
truncation of singular values \(r\le0\) (to handle rank deficiency), while figure-\ref{solar-data-fsrd-residuals}-(b)
corresponds to the final pruned fSRD output. In the DMD model, the systems low-dimensional coherent spatiotemporal dynamics
are relatively well captured, with local residuals \(\le0.01\) (\(\le 1\%\) error). This corresponds physically to background solar evolution, 
such as the differential rotation, long-lived active regions, and gradually evolving coronal loops. In terms of per element reconstruction, this encompasses 
the majority of the data, leading to the global NRMSE value of \(85\%\) given the normalization.
Non-stationary coherent events, such as the shifting solar storm observed between columns \(0\rightarrow6000\) across all rows
are reconstructed less accurately, raising local residuals between \(0.01\leftrightarrow0.15\). Within columns \(8000\rightarrow12000\), several
major solar flare events occur, corresponding to spatiotemporal  anomalies  or rare transient events. These events are largely missed by DMD
because the fitted linear dynamics are dominated by persistent coherent modes, resulting in localised residuals between 
\(0.2\leftrightarrow0.6\) (i.e., 20 to 60\(\%\) error). As these residuals correspond to the events' amplitudes relative to the normalized data range, 
they indicate that a substantial fraction of the flare signal remains unreconstructed. Although DMD achieves a global NRMSE of 85\%, this performance is 
dominated by its accurate reconstruction of the persistent background evolution. Consequently, while DMD provides an accurate representation of the 
nominal solar dynamics, the model fails to capture the sparse transient events and rare  anomalies  that distinguish the Mother's Day storm 
from a comparatively quiet solar day.

\FloatBarrier

\FloatBarrier

By contrast, consider fSRD's model in figure-\ref{solar-data-fsrd-residuals}-(b). Residuals across the full data range are reduced within the order of
\(\le1e-3\) as seen in the 'Absolute Error' figure, though some individual or small groups of elements within the transient and anomalous regions stretch between \(0.05\leftrightarrow0.1\).
These correspond predominantly to missed peaks in a handful of anomaly amplitudes, otherwise representing some minor oscillatory misalignments 
with non-stationary coherent events. Given this, fSRD successfully captures linear dynamical representations for the low-dimensional coherent dynamics, the non-stationary events, 
and the rare anomalies. Consistent with the bifurcating Lorenz results, this improvement arises because invariant decomposition assigns locally valid invariant partitions across 
changes in system invariance, allowing distinct dynamical regimes to be segmented and modelled independently. 
For example, the DMD model contains 107 eigenvalues total, while in fSRD regions 13, 18, and 19, which isolate one of the major solar flares each, truncated to 148, 141, and 203 eigenvalues
respectively, greatly expanding local expressivity. While future work would need to explore wether this dimensionality is necessary (\(\delta_{BIC}=0\)),
this illustrative example serves to demonstrate fSRD's highly expressive representational capacity via linear representations, 
even when applied to this representative case of real-world noisy data.

%% file: sections/descussion.tex
\section{Discussion and Conclusions}\label{discussion-and-conclusions}

In summary, this work introduces a powerful new machine learning
framework, fSRD, for the parsimonious modelling of high dimensional,
nonlinear and/or chaotic dynamics with minimal prior knowledge via a
non-parametric fuzzy hierarchical tree network. Through the novel concept
of invariant decomposition, fSRD is a systematic and automated method
for generating a multi-operator finite Koopman representation for a
given matrix, which to the author's knowledge has not been demonstrated.
W.r.t. Koopman literature, this greatly reduces Koopmans historic
dependence on restrictive system priors, avoiding hard structural
limits to traditional globally defined observable techniques such as a
lack of global topological conjugacy or lack of globally defined
eigenvectors. Further, fSRD's one shot reconstructive accuracy was
demonstrated to be highly robust and adaptable by contrast across tested conditions, despite
experimental limitations (gaussian noise, trajectory position, non-ideal
sampling, etc.).

More generally, developing upon Neural Fuzzy LMN architectures, fSRD
automates the search, allocation, and fuzzy union of locally valid
disjoint Koopman invariant partitions (i.e., resolving region placement
and boundary integration). Strong reasoning for model structure is
provided, both via highly interpretable orthogonally based linear
representation and clear statistical justification in both the forward
and backwards passes (i.e., information theoretic criteria). Through
this systematic Koopman representation without the use of system priors,
fSRD has demonstrated highly expressive representation, suggesting potentially universal approximation capability for sequential
processes represented as finite data matrices. Consequently, fSRD provides a general
alternative to a broad class of non-parametric estimators. This
expressivity was demonstrated across both a representative real-world
application and a range of benchmark systems. These included systems
exhibiting highly nonlinear and chaotic dynamics, as well as cases with
quasi-periodic behaviour, continuous spectra, and delay-embedded
representations, alongside heterogeneous and spatiotemporal signals,
transient dynamics, and evolving baselines. The evaluation further
encompassed systems requiring multiple local or regime-dependent
operators, alongside both small and large data regimes, varying
experimental quality (e.g., sampling strategies and noise levels), and
non-stationary settings such as bifurcations.

\newpage

This capability is supported by the following:

\begin{itemize}
\item
  It admits highly accurate reconstruction of structured input
  representations.
\item
  It demonstrates highly expressive approximation of the induced
  evolution operator, such that observed transitions exhibiting
  nontrivial structure can be represented in a finite-dimensional (or
  asymptotically growing) space in which their evolution is approximated
  by linear operators. This does not require the
  underlying system to admit a single global linear operator.
\item
  It operates on generic vectorized inputs, requiring only that
  sequential dependencies exist between columns or sub regions of the
  data matrix. This permits flexible representation and assembly,
  enabling application across several data formats beyond the dynamic's
  inspiration (time series, images, videos, response surfaces,
  multi-modal, graphs/networks, etc).
\end{itemize}

Although we provide no proof of universality, the case studies presented demonstrate that fSRD 
possesses highly expressive and adaptive approximation capability for sequential structure as 
realised in finite data (i.e, not over arbitrary infinite-horizon dynamical 
or stochastic systems). In these cases,
the ability to achieve highly accurate approximation appeared to be primarily limited 
by the presence of stochastic noise superimposed on the signal. In the Duffing factorial experiments, 
expressivity remained high until the noise level became sufficiently large that the underlying
local structure was obscured or no longer recoverable, consequently preventing structured representation.
Future studies will need to explore fSRD's potential in this respect.

fSRD fills a gap between deep learning's neural architectures that
approximate sequence-to-sequence mappings, and analytical
system-theoretic models that use explicit priors to describe state
evolution, bridging the two paradigms. As a novel ML architecture, many
elements were chosen in this work to demonstrate proof of concept, with
the scale of generality making it impractical to explore all
trajectories in a single study. It is the author's expectation that many
elements are thus ripe for algorithmic improvement (e.g., split position
optimization, choice of local interpolant, etc), though several
contributions were provided to reach the overarching algorithm:

\begin{itemize}
\item
  A framework for assembling locally invariant Koopman regions
  (geometric perspective of representation space and invariant
  decomposition).
\end{itemize}

\begin{itemize}
\item
  A core ML/AI architecture to unite said disjoint regions (adaptation
  of LMN principles to produce a non-parametric fuzzy hierarchal tree
  network).
\item
  A methodology for selecting and optimizing candidate region ensembles
  (forward pass and bespoke invariance metrics).
\item
  A local region definition considering architectural limitations (local
  DMD framework and topological transform).
\item
  Frameworks for enforcing local and global parsimony (automated
  heuristic for SVD regularization and restricted integral pruning
  respectively).
\end{itemize}

Future work will need to explore these optimizations, but also the
useful limits of fSRD's interpretable generalisation via asymptotic
invariant decomposition. For example, to maintain simplicity,
generalisability, and \emph{a-priori} independence, this proof of
concept did not separate the premise and consequent space, further
confining the total input vector to two axes'
\(\mathbf{u = \lbrack}u_{m},u_{t}\mathbf{\rbrack}\) (i.e., vectorized
state and evolution). This corresponds to the assumed use of
\(g( \mathbf{x} ) = \mathbf{x}\). As was demonstrated across
the Lorenz system case studies, fSRD systematically broke down the
continuous spectra system into locally valid finite ergodic partitions,
producing high reconstructive accuracy. Dependent on the
systems data co-ordinates however (e.g., data matrix assembly), the
level of interpretability for resulting representations, while still
present, varied while confined to \(g(x) = x\). Koopman literature has
readily demonstrated that enriching the observable space of any
individual operator (e.g., nonlinear transforms) can dramatically
improve this performance on specific applications. Alternatively,
current LMN literature demonstrates the ability to explicitly augment
the input vector with additional axes (e.g. delayed states or scheduling
variables), which can then be selectively separated into premise and
consequent variables generically or per application. Further exploration
of these input space design choices while retaining fSRD's strengths
(e.g., automatability via a-priory independence) holds potential to
increase performance and extend its applicability, either via richer
structured input spaces, stronger expressivity, or more flexible
partitioning.

Even with relatively limited adaptation, fSRD demonstrates clear potential for
real-world commercial applications. Given the generality
described, the possible applications are vast, but the following
non-exhaustive broad categories are provided:

\begin{itemize}
\item
  Computer Vision: Parameterization via Koopman decomposition along with
  locating invariant interfaces (i.e., fuzzy split locations) enables
  identification and segmentation of linear and nonlinear dynamics.
  Combined with localised modal feature extraction, integration of fSRD
  could thus enable more efficient detection, robust object tracking,
  and reduced data requirement compared to current deep convolutional
  NN's, recurrent NN's and transformer-based pipelines. This is relevant
  in anomaly detection, object detection, medical imaging, surveillance,
  etc.
\item
  Control \& Signal Processing: Both fields fundamentally rely on
  expressing systems in linear bases where powerful tools such as
  filtering, estimation, optimisation, and decomposition become
  tractable. Techniques ranging from state estimation and transfer
  analysis to predictive control all depend on the ability to work with
  structured linear operators and their matrix factorizations. Expanding
  the class of nonlinear systems that admit useful linear
  representations therefore has broad practical and methodological
  implications. In practice, this could enable longer prediction
  horizons, more reliable model predictive or linear quadratic regulator
  style controllers, and more effective sensing or reconstruction
  strategies. The resulting improvements in robustness, safety, and
  computational efficiency are directly relevant to autonomous vehicles,
  drones, aerospace systems, industrial robotics, and communications
  infrastructure.
  \newpage
\item
  AI/ML: Koopman-based finite embeddings could provide a structured,
  linear alternative to sequence models (transformers, RNNs, etc.),
  offering more parsimonious, interpretable, and stable representations.
  This approach could improve efficiency at scale, enable longer-term
  sequence modeling, and be applied across a wide range of tasks,
  including natural language processing, speech, video generation,
  time-series forecasting (e.g., weather, finance), and reinforcement
  learning. By providing a principled, dynamical perspective on
  sequential data, a sufficiently accurate finite approximation of
  Koopman embeddings may offer a complementary or alternative approach
  to current nonparametric sequence estimators, especially in the
  context of edge devices.
\end{itemize}